\documentclass[pdflatex,sn-vancouver,Numbered]{sn-jnl}% Vancouver Reference Style
\usepackage{graphicx}%
\usepackage{multirow}%
\usepackage{amsmath,amssymb,amsfonts}%
\usepackage{amsthm}%
\usepackage{mathrsfs}%
\usepackage[title]{appendix}%
\usepackage[table,xcdraw]{xcolor}%
\usepackage{textcomp}%
\usepackage{manyfoot}%
\usepackage{booktabs}%
\usepackage{algorithm}%
\usepackage{algorithmicx}%
\usepackage{algpseudocode}%
\usepackage{listings}%
\usepackage{booktabs}
\usepackage{multirow}
\usepackage{array}
\usepackage[font=footnotesize,labelfont=bf]{caption}
\usepackage{pdflscape}
\usepackage{sourcecodepro} % Use Source Code Pro as Monospace Code Font
\usepackage[section]{placeins}
\usepackage{tabularx}
\usepackage{hyperref}
\usepackage{newunicodechar}
\newunicodechar{≤}{\ensuremath{\leq}}

\lstdefinestyle{prompt}{
    basicstyle=\ttfamily\footnotesize, % Monospace font, small size
    backgroundcolor=\color{gray!10},   % Light gray background
    frame=single,                      % Box around the code
    rulecolor=\color{black},           % Box color
    breaklines=true,                   % Break lines if too long
    breakindent=0pt,                   % No additional indentation for wrapped lines
    prebreak={},                       % No extra characters at the start of wrapped lines
    postbreak={},                      % No extra characters at the end of wrapped lines
    tabsize=4,                         % Tab size
    showstringspaces=false,            % Don't show spaces in strings
    numbers=none,                      % No line numbers
}

\lstdefinestyle{prompt_linenum}{
    basicstyle=\ttfamily\footnotesize, % Monospace font, small size
    backgroundcolor=\color{gray!10},   % Light gray background
    frame=single,                      % Box around the code
    rulecolor=\color{black},           % Box color
    breaklines=true,                   % Break lines if too long
    breakindent=0pt,                   % No additional indentation for wrapped lines
    prebreak={},                       % No extra characters at the start of wrapped lines
    postbreak={},                      % No extra characters at the end of wrapped lines
    tabsize=4,                         % Tab size
    showstringspaces=false,            % Don't show spaces in strings
    numbers=left,                      % Show line numbers
    numberstyle=\ttfamily,
}

\begin{document}

\title[Article Title]{\textit{VeriFact}: Verifying Facts in LLM-Generated Clinical Text with Electronic Health Records}

%%====================%%
%% Author Information %%
%%====================%%

\author*[1]{\fnm{Philip} \sur{Chung}}\email{chungp@stanford.com}
\author[2]{\fnm{Akshay} \sur{Swaminathan}}
\author[1]{\fnm{Alex J.} \sur{Goodell}}
\author[1]{\fnm{Yeasul} \sur{Kim}}
\author[1,2,3]{\fnm{S. Momsen} \sur{Reincke}}
\author[1]{\fnm{Lichy} \sur{Han}}
\author[1]{\fnm{Ben} \sur{Deverett}}
\author[1]{\fnm{Mohammad Amin} \sur{Sadeghi}}
\author[1,2,3]{\fnm{Abdel-Badih} \sur{Ariss}}
\author[1]{\fnm{Marc} \sur{Ghanem}}
\author[1,6,7]{\fnm{David} \sur{Seong}}
\author[1]{\fnm{Andrew A.} \sur{Lee}}
\author[1]{\fnm{Caitlin E.} \sur{Coombes}}
\author[1]{\fnm{Brad} \sur{Bradshaw}}
\author[1]{\fnm{Mahir A.} \sur{Sufian}}
\author[1]{\fnm{Hyo Jung} \sur{Hong}}
\author[1]{\fnm{Teresa P.} \sur{Nguyen}}
\author[1]{\fnm{Mohammad R.} \sur{Rasouli}}
\author[1]{\fnm{Komal} \sur{Kamra}}
\author[1]{\fnm{Mark A.} \sur{Burbridge}}
\author[1]{\fnm{James C.} \sur{McAvoy}}
\author[1]{\fnm{Roya} \sur{Saffary}}
\author[4]{\fnm{Stephen P.} \sur{Ma}}
\author[5]{\fnm{Dev} \sur{Dash}}
\author[1]{\fnm{James} \sur{Xie}}
\author[1]{\fnm{Ellen Y.} \sur{Wang}}
\author[1]{\fnm{Clifford A.} \sur{Schmiesing}}
\author[4,8,9]{\fnm{Nigam} \sur{Shah}}
\author[1,2,3]{\fnm{Nima} \sur{Aghaeepour}}

\affil*[1]{\orgdiv{Department of Anesthesiology, Perioperative and Pain Medicine}, \orgname{Stanford Medicine}}
\affil[2]{\orgdiv{Department of Biomedical Data Science}, \orgname{Stanford Medicine}}
\affil[3]{\orgdiv{Department of Pediatrics}, \orgname{Stanford Medicine}}
\affil[4]{\orgdiv{Department of Medicine}, \orgname{Stanford Medicine}}
\affil[5]{\orgdiv{Department of Emergency Medicine}, \orgname{Stanford Medicine}}
\affil[6]{\orgdiv{Immunology Program}, \orgname{Stanford Medicine}}
\affil[7]{\orgdiv{Medical Scientist Training Program}, \orgname{Stanford Medicine}}
\affil[8]{\orgdiv{Clinical Excellence Research Center}, \orgname{Stanford Medicine}}
\affil[9]{\orgdiv{Technology and Digital Solutions}, \orgname{Stanford Health Care}}

%%==========%%
%% Abstract %%
%%==========%%

\abstract{Methods to ensure factual accuracy of text generated by large language models (LLM) in clinical medicine are lacking. \textit{VeriFact} is an artificial intelligence system that combines retrieval-augmented generation and LLM-as-a-Judge to verify whether LLM-generated text is factually supported by a patient’s medical history based on their electronic health record (EHR). To evaluate this system, we introduce \textit{VeriFact-BHC}, a new dataset that decomposes Brief Hospital Course narratives from discharge summaries into a set of simple statements with clinician annotations for whether each statement is supported by the patient’s EHR clinical notes. Whereas highest agreement between clinicians was 88.5\%, \textit{VeriFact} achieves up to 92.7\% agreement when compared to a denoised and adjudicated average human clinician ground truth, suggesting that \textit{VeriFact} exceeds the average clinician’s ability to fact-check text against a patient's medical record. \textit{VeriFact} may accelerate the development of LLM-based EHR applications by removing current evaluation bottlenecks.}


%%\pacs[JEL Classification]{D8, H51}

%%\pacs[MSC Classification]{35A01, 65L10, 65L12, 65L20, 65L70}

\maketitle

%%==============%%
%% Introduction %%
%%==============%%

\begin{figure}[htb]
    \centering
    \includegraphics[width=1\linewidth]{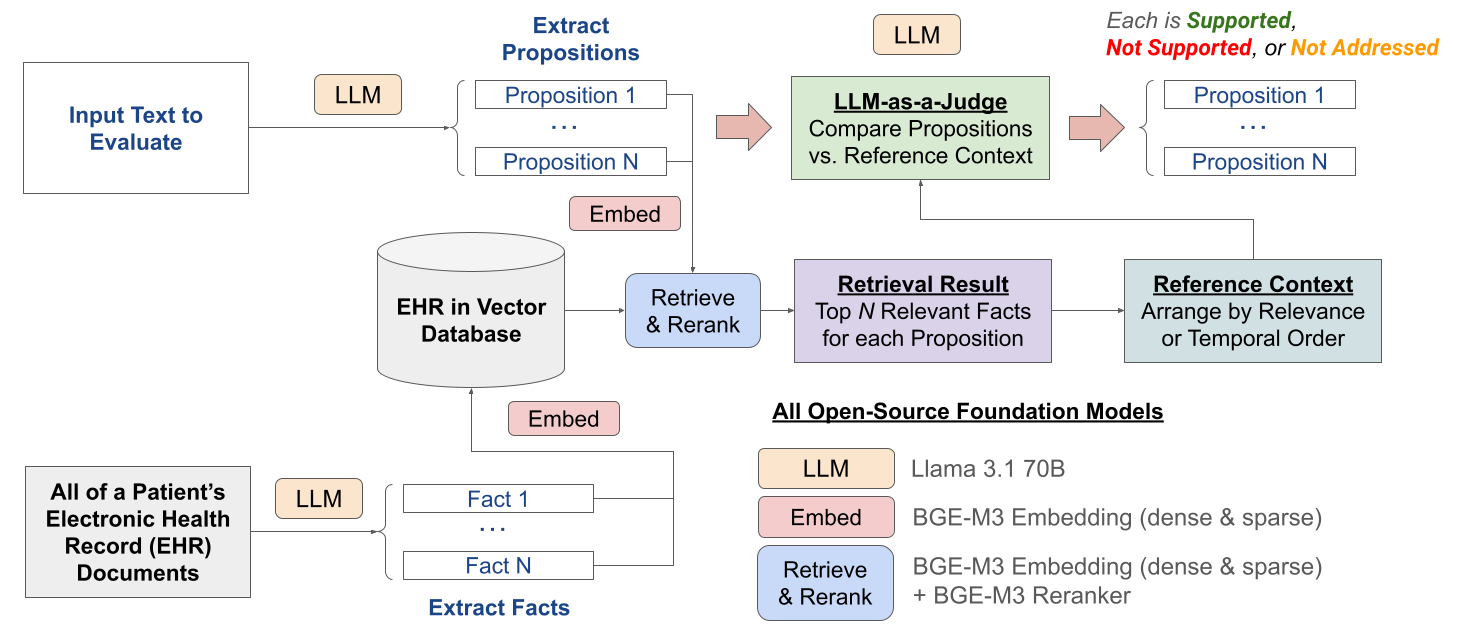}
    \caption{Schematic of the \textit{VeriFact} system. \textit{VeriFact} decomposes long-form input text such as a Brief Hospital Course narrative into a set of propositions for more detailed evaluation. Each patient’s EHR is also decomposed into a set of facts. For each proposition, \textit{VeriFact} dynamically retrieves only the most relevant facts in the EHR to form a reference context specific to that proposition. Subsequently, each proposition and its corresponding reference context of facts are presented to an LLM-as-a-Judge tasked with determining whether each proposition is \textit{Supported}, \textit{Not Supported}, or \textit{Not Addressed} by the facts in the EHR. Multiple foundation models are used in \textit{VeriFact}: (1) “LLM” refers to a Llama 3.1 70B large language model that is used to decompose text into propositions and facts. It is also applied as a judge to read and compare each proposition with its reference context to perform a classification task. (2) “Embed” refers to the BGE-M3 bi-encoder language model that produces vector representations of propositions and EHR facts for vector search. Dense vector search utilizes semantic similarity whereas sparse vectors search utilizes token \& lexical similarity to perform information retrieval. (3) “Rerank” refers to the BGE-M3 Reranker cross-encoder language model that can jointly attend to both proposition and fact and assign a more nuanced ranking score than bi-encoder language models. Experiments are conducted using only open-source foundation models to maximize transparency and reproducibility, but similar foundation models can be substituted.}
    \label{fig:figure1}
\end{figure}

\section{Introduction}\label{intro}

Large language models (LLMs) can summarize patient electronic health records (EHR)\cite{Van_Veen2024-tm, Tang2023-jn}, perform clinical risk prediction\cite{Chung2024-dk} and diagnostic reasoning\cite{Savage2024-yh, Goh2024-cc}, and translate patient instructions or discharge summaries to patient-friendly language\cite{Zaretsky2024-oj, Hong2024-cd}. Yet LLM deployment in clinical practice is limited in part because LLM evaluation to date has largely focused on medical question-answer examination benchmarks with a minority of studies utilizing real patient care data or long-form text generation\cite{Bedi2024-rb, Bannur2024-qn, Shah2023-ai, Xie2024-lo, Munnangi2024-wl}. Long-form text such as clinical notes are pervasive in medical practice, but it is impossible for clinician vigilance alone to ensure factuality of LLM-generated text given the needle-in-a-haystack challenge of identifying subtle errors and hallucinations\cite{Adler-Milstein2024-qf, Augenstein2024-tv}. \textit{VeriFact} is an artificial intelligence (AI) evaluation system that addresses this gap by automatically verifying whether any text, whether LLM-written or human-authored, is supported by a patient’s EHR (\hyperref[fig:figure1]{Figure 1}). Text written about a patient should be internally consistent with information already known about the patient, such as that stored in their medical records. \textit{VeriFact} decomposes the text into a set of statements that it will verify using retrieval-augmented generation (RAG)\cite{Lewis2020-fk} with an LLM-as-a-Judge evaluator\cite{Zheng2023-kj, Guan2024-oq} to mimic the manual process of chart review and interpretation of facts that clinicians employ when evaluating patient documentation for veracity. \textit{VeriFact} focuses on fact-checking against patient-specific EHRs as opposed to general domain fact-checking\cite{Wei2024-ur, Min2023-zk, Kamoi2023-tm, Chen2024-rd, Es2024-qc}.

We introduce \textit{VeriFact-BHC} as a new dataset to evaluate \textit{VeriFact’s} fact-checking performance against human clinicians. \textit{VeriFact-BHC} contains 13,290 statements from the Brief Hospital Course (BHC) narratives of 100 patients drawn from the publicly-available MIMIC-III dataset\cite{Johnson2016-bt, Johnson2016-ra}. Each statement is annotated by multiple human clinicians as \textit{Supported}, \textit{Not Supported}, or \textit{Not Addressed} by the patient’s EHR. A final denoised human clinician ground truth set of labels is obtained via majority voting and adjudication. We publicly release \textit{VeriFact-BHC} to enable researchers to develop better patient-specific EHR fact-checking systems.

\textit{VeriFact} adopts concepts from Bertrand Russell’s logical atomism\cite{Russell2009-tm, Wanner2024-ts} and first-order predicate logic by breaking down candidate input text for evaluation into a set of logical proposition statements that can be individually verified (\hyperref[fig:figure1]{Figure 1}). A logical proposition is a declarative statement that asserts something about the world and must have a truth value, i.e. the statement is either true or false\cite{Smith2012-us}. We explore two types of text decompositions into propositions: complete sentences and atomic claim propositions (\hyperref[supfigs_sub1]{Appendix A.1})\cite{Min2023-zk, Chen2024-rd, Wang2024-gb, Wanner2024-ts,Chen2024-lm}. Sentences are extracted using a sentence splitter whereas atomic claims are extracted by instructing an LLM to re-express text as simple sentences that encapsulate Subject-Object-Predicate relationships. The resultant atomic claims tend to satisfy the definition of formal logic propositions and bear resemblance to the semantic triple data structure used in knowledge graphs and search engines\cite{Miller1998-ea, Bhattacharyya2015-zx}. All documents in a patient’s EHR also undergo the same extraction, but we refer to sentences and atomic claims derived from the EHR as facts because they represent known truths about the patient. Facts are stored in a vector database and collectively form the basis of reference knowledge used to verify any proposition asserted about the patient.

\begin{figure}[hbt]
    \centering
    \includegraphics[width=1\linewidth]{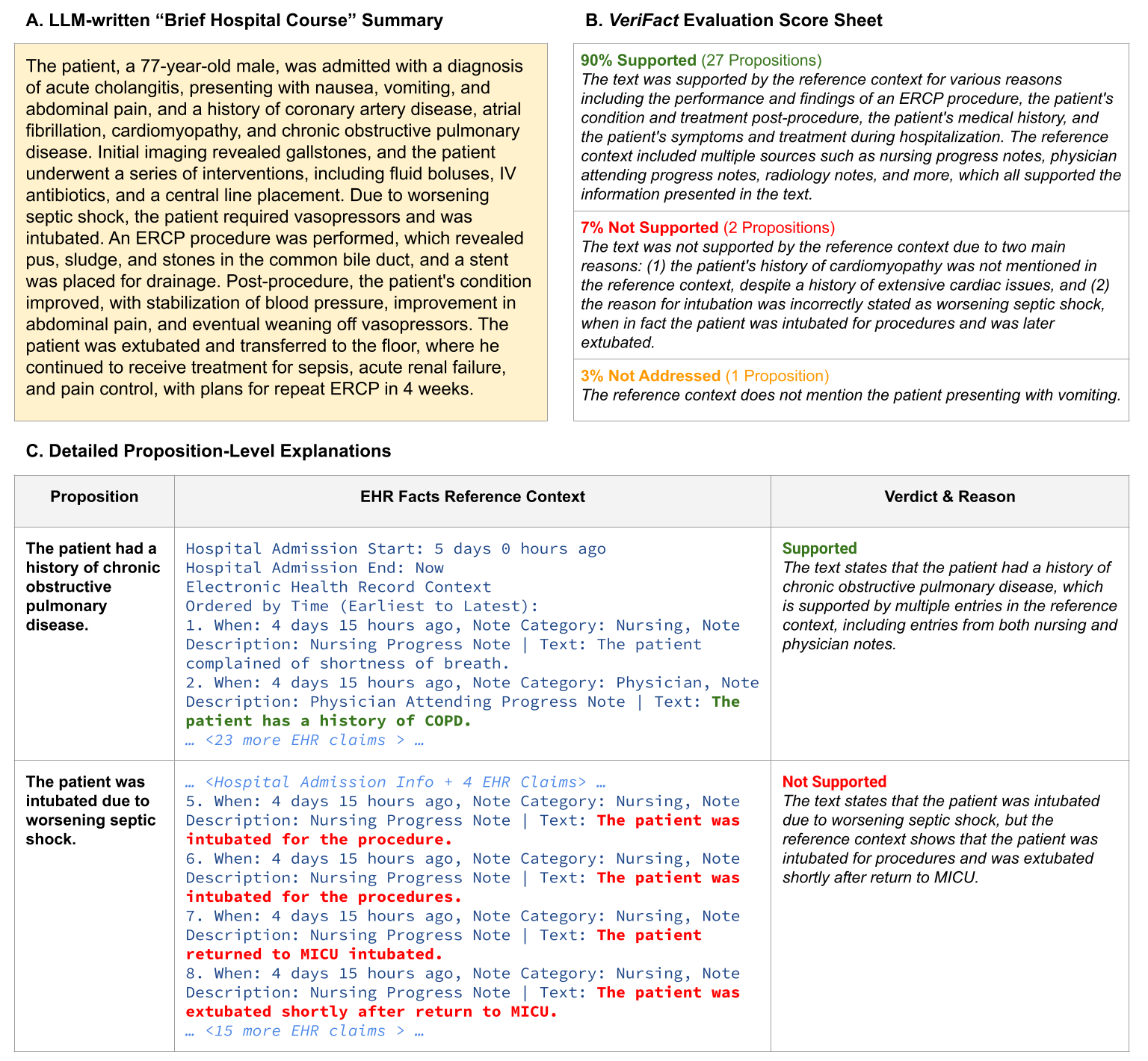}
    \caption{\textbf{(A)} An LLM-written Brief Hospital Course narrative that summarizes the hospital admission, created by applying iterative rolling summarization across all of the patient’s EHR clinical notes from the admission. \textbf{(B)} \textit{VeriFact} Evaluation Score Sheet output when verifying the LLM-written Brief Hospital Course with respect to the patient’s EHR clinical notes. The score sheet provides an overview of how much of the text is \textit{Supported}, \textit{Not Supported}, and \textit{Not Addressed}, and a summary explanation for the scores. The summary explanations are generated by summarizing verdict explanations for each proposition in the label category. \textbf{(C)} Illustration of detailed proposition-level information presented to the LLM-as-a-Judge along with the assigned verdict and explanation. \textit{VeriFact} jointly considers each proposition with the corresponding EHR Facts Reference Context to assign a verdict and generate a reason for the verdict. The Proposition column shows atomic claim propositions extracted from the LLM-written Brief Hospital Course summary in 2A. The EHR Facts Reference Context illustrates “Relative Time” formatting where the current time is set to the time of discharge (when the Brief Hospital Course narrative input text would be composed) and all other timestamps are converted into days and hours relative to the current time.}
    \label{fig:figure2}
\end{figure}

Each proposition being evaluated is paired with a dynamically constructed reference context, created by retrieving a subset of relevant facts from the EHR vector database. The reference context includes each fact along with source clinical note’s metadata such as note type, author type, date \& time. An LLM-as-a-Judge is then used to evaluate the proposition with respect to the reference context to determine a verdict–whether each proposition is \textit{Supported}, \textit{Not Supported}, or \textit{Not Addressed} by the patient’s EHR\cite{Zheng2023-kj}. \hyperref[fig:figure2]{Figure 2A} depicts an LLM-written discharge summary narrative with \textit{VeriFact’s} evaluation output shown in \hyperref[fig:figure2]{Figure 2B}. The evaluation score sheet indicates the percentage of the input text assigned to each verdict label accompanied by LLM-generated explanations for why these scores were assigned. Additionally, \textit{VeriFact} produces detailed proposition-level explanations for why each verdict was assigned (\hyperref[fig:figure2]{Figure 2C}). Even though the same LLM is used for text generation, decomposition, and evaluation, the differing prompts instruct LLMs to role-play as agents with different behaviors and task objectives (\hyperref[supfigs_sub2]{Appendix A.2}, \hyperref[supfigs_sub3]{Appendix A.3}, \hyperref[supfigs_sub4]{Appendix A.4}). 

Our experiments study two types of long-form clinical text: (1) the original human-written BHC from the MIMIC-III discharge summary (“human-written summary”), and (2) an LLM-written BHC (“LLM-written summary”) which was generated \textit{de novo} using all clinical notes from the patient’s hospital admission. Human-written narratives are the current status quo in clinical care, but LLM-written summaries represent an envisioned scenario where LLMs are deployed in clinical applications to generate text about patients. For each input text type, we study both sentence and atomic claim propositions as units of evaluation and fact representation from the EHR. Our experiments reveal that \textit{VeriFact} achieves human-level performance on long-form fact checking against a patient’s own EHR, paving the path toward using AI for patient-specific evaluation of LLM-generated text.

%%=========%%
%% Results %%
%%=========%%

\section{Results}\label{results}

\subsection{The \textit{VeriFact-BHC} Dataset}\label{results_sub1}

The \textit{VeriFact-BHC} dataset contains 13,290 proposition statements extracted from human-written and LLM-written Brief Hospital Course (BHC) narratives for 100 patients from the MIMIC-III Clinical Database. Paired with each patient’s BHC narratives is a longitudinal EHR containing all other clinical notes entered prior to the discharge summary for the hospital admission. \hyperref[suptables_sub1]{Appendix B.1} describes the patient demographics and resulting dataset.

Information content and length of the two types of BHC narratives differ based on information available to the author. The MIMIC-III dataset contains clinical notes from the intensive care unit (ICU) as well as radiology and cardiology reports, but lacks inpatient notes from other hospital units, services or outpatient clinic visits. The LLM-writer could generate a BHC based only on information in the MIMIC-III dataset, but the original clinician authoring the human-written BHC had access to other information sources such as notes from other hospital units, outpatient clinics, and bedside discussions with the patient and care team. This information asymmetry reflects a real-world difference between LLM-written and human-written clinical text. LLM-written summaries were shorter than human summaries with a median length of 164 versus 334 words, a consequence of the specific LLM-writer pipeline used in this study which aggressively compresses the BHC. In contrast, the original human-written summaries do not have any length or content limitation.

BHC narratives are decomposed into sentence and atomic claim propositions, representing two types of evaluation units. Each proposition undergoes a binary assessment of its validity based on formal logic criteria, followed by clinician annotation regarding whether each proposition is \textit{Supported}, \textit{Not Supported} or \textit{Not Addressed} by the clinical notes from the patient’s electronic health record. Each proposition is annotated by three physicians and a human clinician ground truth representing average clinician judgment is derived via majority vote and adjudication. All BHC narratives, corresponding patient EHRs, human proposition annotations and \textit{VeriFact} proposition verdicts are made publicly available as part of the \textit{VeriFact-BHC} dataset.

\begin{figure}[hbt]
    \centering
    \includegraphics[width=1\linewidth]{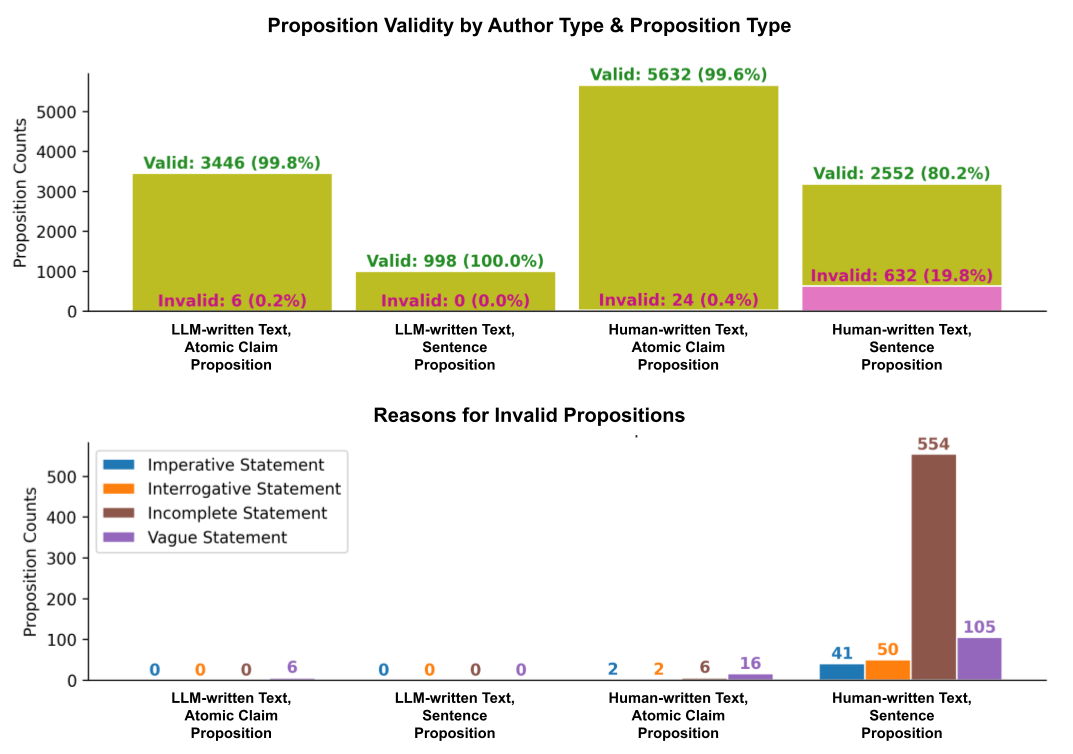}
    \caption{\textbf{Top:} Count and percent distribution of valid and invalid propositions for each of the author types (LLM-written vs. human-written Brief Hospital Course text) and proposition types (Atomic Claim vs. Sentence Proposition). \textbf{Bottom:} Breakdown of reasons for why propositions were invalid. Each proposition can have multiple reasons for being invalid. Sentence propositions from human-written text in particular exhibit many invalid propositions predominantly due to incomplete and vague statements. LLM-written text or atomic claim propositions which require an LLM to perform extraction rarely result in invalid propositions.}
    \label{fig:figure3}
\end{figure}

\subsection{Atomic Claim and Sentence Proposition Validity}\label{results_sub2}

Atomic claim extraction reliably produces valid propositions whereas sentence propositions result in 19.8\% invalid propositions when extracted from human-written text (\hyperref[fig:figure3]{Figure 3}). Human-written clinical notes often contained partial or vague statements with special formatting such as lists and bullets which may be poorly parsed by a sentence splitter. Human-written notes also contained imperative statements directing a clinical plan of action or interrogative statements expressing a clinician’s uncertainty, neither of which are propositions in first-order formal logic. These invalid propositions fail to assert a statement about the state of the world, resulting in an inability for \textit{VeriFact} or human annotators to evaluate the proposition as \textit{Supported}, \textit{Not Supported}, or \textit{Not Addressed}. These challenges are largely overcome by using LLM for atomic claims extraction which tends to rephrase information as valid propositions and reduces their incidence to 0.4\% in human-written summaries. LLM-written summaries rarely exhibit invalid propositions since LLMs tend to write fluent and grammatically correct sentences from which either proposition types can be easily extracted. Atomic claim propositions also enable more fine-grained evaluation with 9,108 propositions compared to sentence propositions with 4,182 propositions in \textit{VeriFact-BHC}, yielding roughly 2.2 times the number propositions from the same input text.

\subsection{Agreement Between Human Clinicians}\label{results_sub3}

\begin{table}[ht]
    \centering
    \label{tab:table1}
    \footnotesize % Reduce font size
    \renewcommand{\arraystretch}{1.2} % Adjust row height for readability
    \begin{tabularx}{\textwidth}{@{}l>{\raggedright\arraybackslash}ccccccX@{}}
        \toprule
        \multicolumn{6}{c}{\textbf{\shortstack{Table 1: Inter-Clinician Agreement on Whether Propositions are \\ \textit{Supported}, \textit{Not Supported}, or \textit{Not Addressed} by the Patient's Electronic Health Record}}} \\ 
        \midrule
        \textbf{Propositions} &
        &
        \textbf{\shortstack{LLM-written \\ Atomic Claims}} &
        \textbf{\shortstack{LLM-written \\ Sentences}} &
        \textbf{\shortstack{Human-written \\ Atomic Claims}} &
        \textbf{\shortstack{Human-written \\ Sentences}} \\ 
        \midrule
        \multirow{3}{*}{\textbf{All (N=13,290)}} &
        \textbf{\shortstack{\% Agreement \\ (95\% CI)}} &
        \shortstack{88.5\% \\(87.9-89.4\%)} &
        \shortstack{84.7\% \\(83.4-87.0\%)} &
        \shortstack{73.3\% \\(72.7-73.8\%)} &
        \shortstack{66.6\% \\(65.7-67.8\%)} \\
        & \\
        &
        \textbf{\shortstack{Gwet's AC1 \\(95\% CI)}} &
        \shortstack{0.88 \\(0.87-0.89)} &
        \shortstack{0.83 \\(0.82-0.86)} &
        \shortstack{0.64 \\(0.63-0.65)} &
        \shortstack{0.52 \\(0.50-0.54)} \\ 
        \midrule
        \multirow{3}{*}{\textbf{\shortstack{At Least One \\ \textit{Supported} \\ (N=10,835)}}} &
        \textbf{\shortstack{\% Agreement \\ (95\% CI)}} &
        \shortstack{88.7\% \\(87.9-89.6\%)} &
        \shortstack{84.9\% \\(82.9-86.7\%)} &
        \shortstack{75.5\% \\(74.5-76.5\%)} &
        \shortstack{65.8\% \\(64.3-67.4\%)} \\
        & \\
        &
        \textbf{\shortstack{Gwet's AC1 \\(95\% CI)}} &
        \shortstack{0.88 \\(0.87-0.89)} &
        \shortstack{0.84 \\(0.81-0.86)} &
        \shortstack{0.71 \\(0.70-0.73)} &
        \shortstack{0.58 \\(0.55-0.60)} \\ 
        \midrule
        \multirow{3}{*}{\textbf{\shortstack{At Least One \\ \textit{Not Supported} \\ (N=3,402)}}} &
        \textbf{\shortstack{\% Agreement \\ (95\% CI)}} &
        \shortstack{34.0\% \\(32.0-36.0\%)} &
        \shortstack{31.4\% \\(29.4-33.5\%)} &
        \shortstack{33.2\% \\(32.2-34.2\%)} &
        \shortstack{32.1\% \\(31.0-33.3\%)} \\
        & \\
        &
        \textbf{\shortstack{Gwet's AC1 \\(95\% CI)}} &
        \shortstack{0.09 \\(0.06-0.12)} &
        \shortstack{0.06 \\(0.03-0.10)} &
        \shortstack{0.02 \\(-0.00-0.03)} &
        \shortstack{0.00 \\(-0.02-0.02)} \\ 
        \midrule
        \multirow{3}{*}{\textbf{\shortstack{At Least One \\ \textit{Not Addressed} \\ (N=3,979)}}} &
        \textbf{\shortstack{\% Agreement \\ (95\% CI)}} &
        \shortstack{31.1\% \\(28.7-33.6\%)} &
        \shortstack{30.8\% \\(25.6-36.3\%)} &
        \shortstack{45.7\% \\(44.4-47.2\%)} &
        \shortstack{50.7\% \\(49.1-52.5\%)} \\
        & \\
        &
        \textbf{\shortstack{Gwet's AC1 \\(95\% CI)}} &
        \shortstack{0.03 \\(-0.01-0.07)} &
        \shortstack{0.02 \\(-0.07-0.10)} &
        \shortstack{0.25 \\(0.23-0.28)} &
        \shortstack{0.34 \\(0.31-0.37)} \\ 
        \bottomrule
    \end{tabularx}
    \caption{Clinician inter-annotator agreement for whether each proposition is \textit{Supported}, \textit{Not Supported}, or \textit{Not Addressed} by a patient’s electronic health record clinical notes. Agreement is measured with Percent Agreement and Gwet’s AC1 with 95\% confidence intervals. Each proposition was labeled by three clinician annotators. In addition to showing agreement across all propositions, the table also shows agreement computed over a subset of propositions where at least one clinician assigned a \textit{Supported}, \textit{Not Supported}, and \textit{Not Addressed} verdict. These subset analyses reveal that a majority of inter-clinician disagreement occur in propositions where at least one clinician assigns a \textit{Not Supported} or \textit{Not Addressed} verdict.}
\end{table}

Each of the 13,290 propositions was annotated by three clinicians from a cohort of 25 clinician annotators over a combined annotation time of 1618 hours (67.4 days). Clinicians had a mean and standard deviation of 5±9.2 years of practice and 8±4.5 years of experience using EHRs. Clinical specialties included anesthesiology, pediatrics, internal medicine, general practice, neurology, emergency medicine, and critical care medicine. Inter-annotator agreement along with author type and proposition type are shown in \hyperref[tab:table1]{Table 1}. Pairwise inter-annotator agreement is shown in \hyperref[supfigs_sub6]{Appendix A.6}.

Percent agreement on propositions between clinicians varied by author type and proposition type. There was higher agreement on atomic claim versus sentence propositions with 88.5\% versus 84.7\% agreement for LLM-written summary and 73.3\% versus 66.6\% agreement for human-written summary. Analysis of proposition subsets with at least one of three clinician annotators assigning a verdict of \textit{Supported}, \textit{Not Supported}, and \textit{Not Addressed} revealed that a majority of the disagreement among human clinicians arose from propositions where \textit{Not Supported} and \textit{Not Addressed} labels are assigned. In these categories, agreement was 30.8-50.7\%. Conversely, clinicians were in high agreement when assigning the \textit{Supported} label with 65.8-88.7\% agreement across the different author type and proposition types (\hyperref[tab:table1]{Table 1}).

\subsection{Human Clinician Ground Truth}\label{results_sub4}
Human clinician ground truth labels for each proposition were established by taking the majority vote of three clinician labels. The 478 propositions with disagreement between the three clinicians were labeled by two additional physicians; if the two additional clinicians disagreed, the propositions were adjudicated through manual discussion between the two physicians and a third physician to help break ties until a consensus ground truth label was established.

The human ground truth labels represent a “average” physician judgment across a variety of clinical specialties and years of experience. The variance in physician judgment is indicated by the inter-clinician agreement shown in \hyperref[tab:table1]{Table 1} with lower agreement indicating greater variance in clinician judgement for propositions in that category. Human clinicians had the greatest difficulty agreeing upon a single ground truth for human-written text and sentence propositions with 66.6\% agreement. Human clinician judgement was significantly more aligned with LLM-written text and atomic claim propositions with 88.5\% agreement. This indicates that the most consistent ground truth was derived from atomic claim propositions extracted from LLM-written text.

\subsection{\textit{VeriFact} Agreement With Human Clinician Ground Truth}\label{results_sub5}
\begin{figure}[!ht]
    \centering
    \includegraphics[width=1\linewidth]{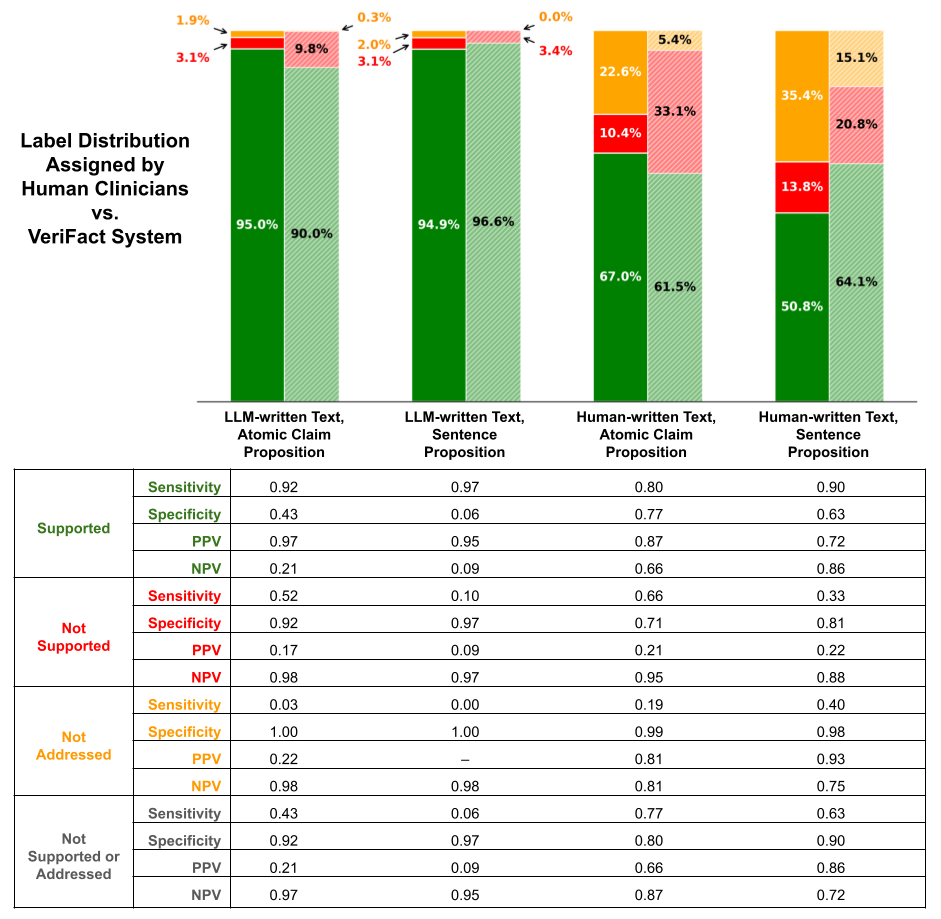}
    \caption{\textbf{Top:} Percent distribution of \textit{Supported}, \textit{Not Supported}, and \textit{Not Addressed} propositions as assigned by human clinician ground truth (solid bars) and \textit{VeriFact} (hatched bars). The \textit{VeriFact} system whose label assignments are depicted utilize the following hyperparameters: Retrieval Method = “Rerank”, Top N = 50, Reference Context Format = “Absolute Time”, Retrieve Facts Only From Current Admission = “No”. Other \textit{VeriFact} systems with different hyperparameters will have different label distributions. Human-written text used in this study contains information that does not appear in the reference EHR notes, leading to a high number of propositions that cannot be \textit{Supported} due to intrinsic information asymmetry between the text generation and evaluation process. In contrast, LLM-written text is generated using the same reference EHR notes as a knowledge source, leading to a high fraction of \textit{Supported} propositions due to the symmetric information utilization in text generation and evaluation processes. \textbf{Bottom:} Sensitivity, specificity, positive predictive value (PPV) and negative predictive value (NPV) for each verdict assignment of the \textit{VeriFact} system using the human clinician ground truth as the gold standard. “\textit{Not Supported or Addressed}” refers to the negative verdict label that arises when the task is binarized by combining \textit{Not Supported} and \textit{Not Addressed}. PPV for \textit{Not Addressed} for LLM-written Text, Sentence Propositions is undefined because zero sentence propositions from LLM-written text are assigned \textit{Not Addressed} verdicts by \textit{VeriFact}.}
    \label{fig:figure4}
\end{figure}

\textit{VeriFact} can now be quantified relative to clinicians by using the ground truth labels as a gold standard (\hyperref[suptables_sub2]{Appendix B.2}, \hyperref[suptables_sub3]{Appendix B.3}, \hyperref[suptables_sub4]{Appendix B.4}, \hyperref[suptables_sub5]{Appendix B.5}). For LLM-written summaries, the best \textit{VeriFact} agreement with ground truth labels achieves 88.8\% and 92.7\% for atomic claim and sentence propositions, respectively (\hyperref[suptables_sub2]{Appendix B.2}). This performance is similar to agreement between random individual clinicians at 88.5\% and 84.7\% for atomic claim and sentence propositions (\hyperref[tab:table1]{Table 1}), suggesting that \textit{VeriFact} performs as well as individual clinicians. We also find that sentence propositions yield slightly higher agreement than atomic claim propositions in our experiments.

One of the top performing \textit{VeriFact} hyperparameter combinations is compared against the human clinician ground truth in \hyperref[fig:figure4]{Figure 4}, which shows the difference in label distribution for information symmetric (LLM-written text) and asymmetric (human-written text) scenarios. \textit{VeriFact’s} performance reflects its ability to discriminate \textit{Supported} propositions well. When a proposition is not \textit{Supported}, \textit{VeriFact} is biased in assigning \textit{Not Supported} rather than \textit{Not Addressed} verdicts when compared against humans. This is especially apparent in the human-written summaries due to the label distribution in the information asymmetric scenario. Binarizing the label space by combining the negative labels as \textit{Not Supported or Addressed} results in a 10-15\% increase in percent agreement for human-written text and a corresponding increase in sensitivity and positive predictive value (PPV). Thus \textit{VeriFact} has poor calibration in assigning the two different negative labels compared to human clinicians, but can perform well when there is only a single negative label (\hyperref[suptables_sub3]{Appendix B.3}).

\subsection{\textit{VeriFact} System Configuration}\label{results_sub6}

\begin{figure}[hbt]
    \centering
    \includegraphics[width=1\linewidth]{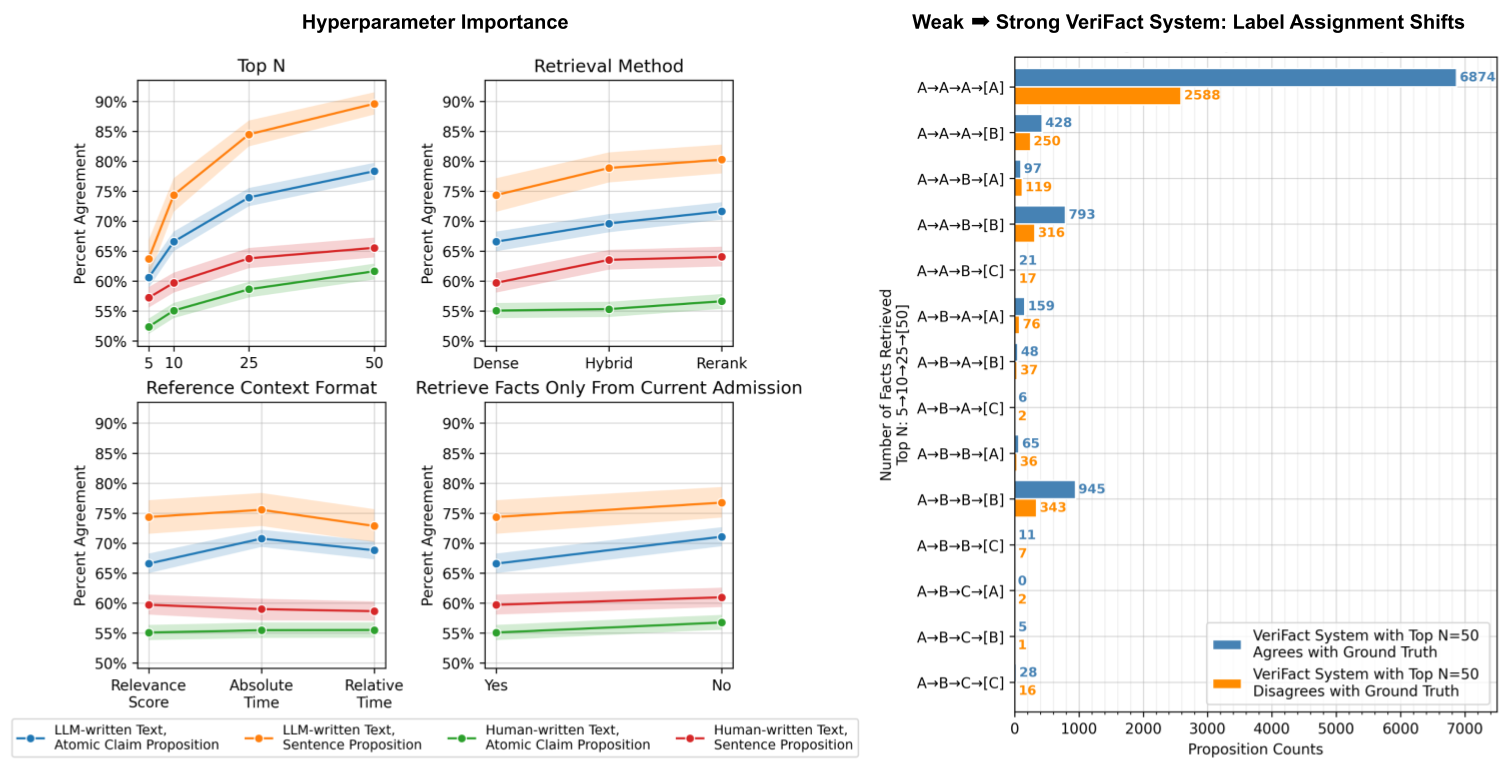}
    \caption{\textbf{Left:} Relative importance of hyperparameters as measured by the Percent Agreement achieved between \textit{VeriFact} and the human clinician ground truth labels. The plots show the effect of varying Top N (number of retrieved EHR facts), retrieval method, reference context format, and whether to limit retrieval scope to current admission. Each plot shows a sensitivity analysis of one of the four hyperparameters that is varied along the plot’s x-axis while the other three hyperparameters are fixed to a default value. The default values are: Top N = 10, Retrieval Method = “Dense”, Reference Context Format = “Relevance Score”, Retrieve Facts Only From Current Admission = “No”. Each line represents \textit{VeriFact} performance on a specific combination of author type and proposition type. Area behind the line depicts 95\% confidence intervals. \textbf{Right:} A depiction of how label assignment changes when transitioning from a weak \textit{VeriFact} system (Top N=5) to a strong \textit{VeriFact} system (Top N=50). “A” represents the label assigned by the weakest \textit{VeriFact} system. The next stronger \textit{VeriFact} system may assign the same label, or may assign a different label as denoted by “B” and “C”. The strongest \textit{VeriFact} system is denoted in brackets and its label is compared against the Ground Truth and the resulting proposition counts are shown as bars. A majority of propositions are assigned the same label regardless of the amount of information retrieved from the EHR. However, a significant fraction of propositions have a change in label assignment as more facts are retrieved, and more often than not, this change in label assignment results in alignment with the human clinician ground truth label. In this analysis, propositions are pooled across author types and proposition types.}
    \label{fig:figure5}
\end{figure}

Multiple hyperparameters can affect \textit{VeriFact} system’s performance: (1) text embedding representation \& information retrieval method, (2) number of facts retrieved from EHR vector database for each proposition, (3) fact-ordering and formatting of facts in the reference context, and (4) whether fact retrieval should be constrained to the current hospital encounter or allowed to span all encounters in the patient’s EHR. \textit{VeriFact} achieves better agreement with the ground truth labels when more facts are retrieved from the EHR, when more advanced hybrid or rerank retrieval techniques were utilized, and when the scope of fact retrieval was not limited to just the current hospital admission but spanned the patient’s entire EHR. The most important hyperparameter was the number of facts retrieved. \hyperref[fig:figure5]{Figure 5} shows that increased fact retrieval leads to label assignment changes that are better aligned with the ground truth labels. The best \textit{VeriFact} systems used hybrid retrieval with a re-ranker model to retrieve 50 facts from across the patient’s entire EHR with either Absolute Time or Relevance Score reference context formats. The lack of superiority between reference context formats suggested that the LLM-as-a-Judge was insensitive to ordering or temporal indicators in the reference context.

%%============%%
%% Discussion %%
%%============%%

\section{Discussion}\label{discussion}

\textit{VeriFact} can evaluate any long-form text discussing a patient by fact-checking the text against the patient’s EHR. At the present, all non-boilerplate clinical text required for clinical care and billing is authored by human clinicians, consuming a considerable amount of time and mental resources. There is promise in using LLMs for clinical text generation tasks to offload this burden and increase the amount of time clinicians can participate in direct patient care activities. However, LLM-written text needs to be evaluated for inaccuracies and hallucinations before it is committed as a part of the patient’s permanent medical record. \textit{VeriFact} can perform this critical guardrailing function with clinician-level performance in a fully automated and scalable fashion. 

Text being evaluated by \textit{VeriFact} can span multiple facts from different documents across a patient’s health record as the system can smartly select a subset of facts for comparison. \textit{VeriFact} can also evaluate statements that do not appear in the EHR verbatim, but require clinical interpretation of facts across multiple documents. This capability arises from using LLM-as-a-Judge to read and interpret both the text being evaluated as well as the reference context of facts prior to assigning a verdict label. \textit{VeriFact} generates an evaluation score sheet with high-level and detailed explanations to alert clinicians to potential factual errors and the reasons why errors may be present (\hyperref[fig:figure2]{Figure 2}). \textit{VeriFact} scores can be used as an internal quality metric in LLM-based software applications where scores can be used to determine whether to accept or reject and revise LLM-written text prior to presenting it to a clinician.

\textit{VeriFact} is intentionally implemented with publicly available open-source general-domain models without specialized fine-tuning, enabling any researcher or hospital system to immediately replicate results and deploy it as a fact verification and chart review system. Since we it does not use medicine-specific models, all results achieved in our studies are due to the intrinsic out-of-domain generalization capabilities of the models and the \textit{VeriFact} system design. Thus applying \textit{VeriFact} to private hospital-specific EHR data should yield similar performance. We demonstrate \textit{VeriFact’s} utility on discharge summary narratives, but \textit{VeriFact} can be used to guardrail any kind of generative text application in clinical care where a chart review is needed for corroboration. In theory, the same approach can also be used to perform fact verification on document corpora that is not the EHR or outside of the medicine domain altogether, though we do not study the performance of these use cases. 

\textit{VeriFact-BHC} is intentionally designed using the publicly available and de-identified MIMIC-III dataset to enable full reproducibility by any researcher without requiring affiliation with a hospital to access patient records. The availability of this dataset allows the research community to develop more sophisticated fact verification systems that may overcome \textit{VeriFact’s} limitations. The presence of de-identification tokens in the MIMIC-III text did obscure meaning and resulted in some poorly-extracted claims and erroneous label assignments; these de-identification tokens do not exist in real-world EHRs.

A 3-label classification task (\textit{Supported}, \textit{Not Supported}, \textit{Not Addressed}) was chosen because of the clinical utility in distinguishing these three categories. \textit{Supported} propositions are fully corroborated by existing information in the EHR. \textit{Not Supported} propositions are either contradictions or statements that are only partially supported. \textit{Not Addressed} propositions are statements not mentioned in the EHR, meaning the statement is either a hallucination or it is new information a clinician wishes to introduce into the patient’s records. The actions a clinician would take for \textit{Not Supported} versus \textit{Not Addressed} are different; however, if this distinction is not needed, these categories can be combined to form a binary classification task, resulting in improved \textit{VeriFact} agreement (\hyperref[fig:figure4]{Figure 4}, \hyperref[suptables_sub2]{Appendix B.2}, \hyperref[suptables_sub3]{Appendix B.3}, \hyperref[suptables_sub4]{Appendix B.4}, \hyperref[suptables_sub5]{Appendix B.5}). 

\textit{VeriFact’s} performance is most impacted by the number of retrieved facts followed by information retrieval method. Hardware constraints limited our experiments to 50 facts for each reference context, but \hyperref[fig:figure5]{Figure 5} indicates performance has not saturated and that additional fact retrieval may continue to improve performance. Our findings concur with prior research showing that hybrid retrieval and second-stage reranking helps overcome domain mismatch between general-domain retrieval models and specialized-domains such as medicine\cite{Lassance2022-qs,Chen2024-mi, Nogueira2019-ql, Dejean2024-hd, Chen2024-vn, Thakur2021-qd, Formal2021-qk, Kong2023-vs}. The formulation of the evaluation task in \textit{VeriFact} as a fact comparison against a reference context is also motivated by prior research showing that LLMs excel at comparison tasks but may fail at true logical reasoning or compositional knowledge tasks\cite{Wang2024-ao,Song2024-la,Parmar2024-ep, Pan2023-kj, Wan2024-op, Patel2024-bc}. Future research is needed to understand whether compositional logical reasoning with LLMs can be improved to be used in \textit{VeriFact}.

Earlier fact-checking literature has advocated for atomic claims as a unit of retrieval and analysis but does not provide a strict definition for atomic claim extraction\cite{Min2023-zk, Kamoi2023-tm, Chen2024-lm, Wei2024-ur, Munnangi2024-wl}. We impose a definition motivated by first-order formal logic and knowledge graphs, resulting in a high proportion of valid propositions derived from text narratives (\hyperref[fig:figure3]{Figure 3}). Prior findings have indicated that atomic claims improve information retrieval for RAG\cite{Chen2024-lm}, but we find that sentence propositions achieve better performance at 92.7\% agreement compared to atomic claim propositions at 88.8\% agreement with LLM-written summaries. However, this comes with the caveat of sentences containing roughly 2.2 times more information than the same number of atomic claims, meaning that reference contexts composed from sentence facts may contain substantially more information than those composed from atomic facts (\hyperref[fig:figure3]{Figure 3}). Conversely, atomic facts are more targeted and specific to the proposition that is being evaluated, explaining why atomic claim propositions still retain strong performance in our experiments despite being disadvantaged in absolute information content.

Additional problems with sentence propositions were encountered: human clinicians struggle to agree upon a single ground truth and nearly all invalid propositions are sentence propositions. Sentence propositions can contain multiple logical claims, each of which can be \textit{Supported}, \textit{Not Supported}, and \textit{Not Addressed}. Yet because a single label is assigned to each proposition, the evaluation task becomes underspecified. Atomic claims are a way to overcome this problem, but new challenges arise in how to properly and consistently extract all atomic claims from the candidate input text. For example, LLMs may perform incorrect atomic claim extraction across compound nouns and compound objects asserting many-to-many associations. Consider an LLM-written summary: \textit{“... the patient underwent various diagnostic tests, including a chest X-ray, ECG, CT scans, and a transthoracic echocardiogram, which showed no evidence of acute cardiopulmonary process, endocarditis, or orbital cellulitis.”} One extracted atomic claim proposition was \textit{“The transthoracic echocardiogram showed no evidence of orbital cellulitis.”}, but this is a nonsensical claim because a transthoracic echocardiogram is an ultrasound of the heart and cannot be used to evaluate an infection around the eyes. The scientific community also continues to debate on how to best decompose text into atomic units of information with some proposing alternatives such as molecular facts\cite{Gunjal2024-mi, Wanner2024-ts, Choi2021-zu, Wei2024-ur}. Overall, our findings support atomic claim propositions as a unit of evaluation with high agreement among both human clinicians and AI systems, though further research is needed on how to consistently extract logically correct and unambiguous atomic claims grounded in commonsense.

Human-written BHC narratives illustrate the information asymmetry between text generator and evaluator. Our experiments show that \textit{VeriFact} has lower agreement with the ground truth labels due to the majority of disagreement arising in how \textit{VeriFact} assigns \textit{Not Supported} and \textit{Not Addressed} labels (Figure 5, Extended Data Table 3). Binarizing the labels reveals that \textit{VeriFact} is able to classify \textit{Supported} propositions with agreement of 79.2\% and 78.5\% for atomic claim and sentence propositions, respectively. This agreement is still lower than with LLM-written texts, implying that if a human were to edit an LLM-generated draft of a clinical note and increase information asymmetry in the editing process, \textit{VeriFact’s} performance may degrade. In this scenario, agreement on atomic claim propositions would drop from 88.8\% towards 79.2\%. However, \textit{VeriFact’s} LLM-generated reasons for each verdict would still be able to provide the end-user interpretable explanations of why text fails to be \textit{Supported} (\hyperref[fig:figure2]{Figure 2}). Better calibration of models, prompts, task specification, and system design may help better align \textit{VeriFact’s} judgment with human clinicians on the negative labels and improve \textit{VeriFact’s} ability to handle evaluation in information asymmetry scenarios.

There are several limitations to our investigation. We do not study the choice of different retrieval models, reranking models, or LLMs. We also do not study the effect of fine-tuning models for the medical domain or preference-tuning to enhance the LLM’s performance in atomic claim extraction or as an LLM–as-a-Judge evaluator, both of which are avenues for future investigation\cite{Wang2024-hv, Bavaresco2024-gt}. We utilize a fixed set of prompts for atomic claim extraction and LLM-as-a-Judge; further prompt optimization and task-specification may also result in better performance. \textit{VeriFact} can only evaluate information that is present in the text and does not address errors of omission where important information is missing from the text. Additionally, \textit{VeriFact} treats any data about the patient documented in the EHR as the truth and entire scope of knowledge about the patient. New patients to a health system who do not have their care records imported into the EHR will result in \textit{VeriFact} labeling a significant fraction of text as \textit{Not Addressed} due to a dearth of records, even if the information is known to patients and clinicians. EHRs have also been noted to contain erroneous information due to misdiagnosis, miscommunication, or copy-pasting outdated information\cite{Bell2020-md}. Even if errors are discovered and corrected by clinicians in newer clinical notes, the errors may persist in older documents which can still be retrieved by \textit{VeriFact} and treated as truths about the patient. Thus \textit{VeriFact’s} accuracy and utility is limited by how well clinicians collectively curate the information within each patient’s EHR.

In conclusion, we present \textit{VeriFact} as a framework for guardrailing LLM-generated text for clinical care applications using a patient’s own EHR. \textit{VeriFact} provides a quick summary of which parts of the text can be verified by the patient’s records paired with both high-level and detailed explanations. Furthermore, \textit{VeriFact} achieves high agreement with human clinicians and is a promising proxy for manual clinician evaluation of LLM-generated text. We study discharge summary narratives as an illustration of \textit{VeriFact’s} capabilities, but \textit{VeriFact} can be applied to guardrail any text discussing a patient in a clinical care setting. We publicly release our clinician-annotated dataset \textit{VeriFact-BHC} to further accelerate research focused on fact checking using the EHR as a patient-specific reference.

%%=========%%
%% Methods %%
%%=========%%

\section{Methods}\label{sec11}
This study adheres to the TRIPOD+AI reporting guidelines\cite{Collins2024-ey}.

\subsection{Foundation Models}\label{methods_sub1}
\subsubsection{Large Language Models}\label{methods_sub1sub1}
The \textit{vLLM} inference engine\cite{Kwon2023-gr} was used to locally host an adaptive-weight quantized Llama 3.1 70B model\cite{Dubey2024-gu}, specifically the \textit{hugging-quants/Meta-Llama-3.1-70B-Instruct-AWQ-INT4} model from Hugging Face Hub52,53. This LLM was used to generate the LLM-written summary with \textit{temperature=0.5} to encourage diversity in writing. 

The same LLM was also used in all steps in \textit{VeriFact} that require an LLM, which include atomic claim extraction, assigning the \textit{Supported}, \textit{Not Supported} and \textit{Not Addressed} label using LLM-as-a-Judge, and generating explanations for why each label is assigned (\hyperref[fig:figure1]{Figure 1}). In \textit{VeriFact}, the LLM is set to a default \textit{temperature=0.1} because low-temperature sampling results in consistent judgements and has been shown to outperform human experts\cite{Zhang2024-tz}. However, low-temperature next-token sampling on smaller or quantized models occasionally resulted in generating a repeated nonsensical text pattern that overflowed the model’s context window, resulting in outputs that could not be properly parsed and passed to downstream steps in \textit{VeriFact}. In these situations, the same prompt was retried with the LLM’s temperature incremented by \textit{0.1} up to a ceiling value of \textit{1.0} until a valid output is obtained. The desire to avoid these repeated nonsensical generations is also the reason why we use a default \textit{temperature=0.1} and not \textit{temperature=0}.

The same LLM with \textit{temperature=0.1} was also used to determine proposition validity using LLM-assisted classification to produce the initial binary classification labels for each of the invalid proposition categories prior to manual human review.

\subsubsection{Structured Output Generation}\label{methods_sub1sub2}
Structured output generation (also known as guided or constrained generation) was implemented using Outlines\cite{Willard2023-wp} to force LLMs’ text output to conform to custom JSON schemas. These JSON outputs were then parsed and passed to downstream \textit{VeriFact} logic, which may include a pipeline of additional structured output generation steps. Structured output inferencing pipelines also contained a back-up self-healing mechanism whereby if the LLM failed to return valid parsable JSON, the output was given back to the LLM with instructions to correct the output to valid JSON. Fixed in-context demonstrations were included with each prompt to increase task success rate and proper response format\cite{Brown2020-sb, Olsson2022-oc}. 

Specific prompts were used to instruct LLMs to perform tasks with narrowly-defined behavior so that LLMs would act as simple agents. The prompts for generating LLM-written summaries are shown in \hyperref[supfigs_sub2]{Appendix A.2}. The prompts for atomic claim extraction from BHC summaries to form propositions and from EHR notes to store as facts in the vector database are shown in \hyperref[supfigs_sub3]{Appendix A.3}. The prompts for LLM-as-a-Judge are shown in \hyperref[supfigs_sub4]{Appendix A.4}. The prompts for generating initial draft labels for proposition validity are shown in \hyperref[supfigs_sub5]{Appendix A.5}, though a human clinician was able to override these labels and curate the final labels for proposition validity.

\subsubsection{Embedding and Re-ranking Models}\label{methods_sub1sub3}
The \textit{Infinity} inference engine\cite{FeilUnknown-ni} was used to locally host the \textit{BAAI/bge-m3} (\textit{M3 Embedding}) model\cite{Chen2024-vn} from Hugging Face Hub. The \textit{M3 Embedding} model generates both dense and sparse text embedding representations, which are used in \textit{VeriFact} to encode propositions and retrieve relevant facts from the EHR. \textit{Infinity} was also used to locally host the \textit{BAAI/bge-reranker-v2-m3} (\textit{M3 Reranker}) cross-encoder language model\cite{Li2023-ow} to perform second-stage reranking of facts retrieved from the EHR vector database. 

\subsubsection{Compute Infrastructure}\label{methods_sub1sub4}
All foundation model inference was performed using a single machine with four NVIDIA L40S GPUs. Asynchronous task parallelism is used to route different inference tasks to appropriate models.

\subsection{\textit{VeriFact-BHC} Dataset Construction}\label{methods_sub2}
\subsubsection{Patient Cohort Selection}\label{methods_sub2sub1}
This is a retrospective study on de-identified electronic health records data which were originally collected for routine clinical care. Experiments were conducted using a subset of the MIMIC-III Clinical Database v1.4 dataset which is a freely available and de-identified EHR dataset for researchers derived from a tertiary/quaternary care hospital, the Beth Israel Deaconess Medical Center in Boston, Massachusetts, USA\cite{Johnson2016-bt}. MIMIC-III contains longitudinal clinical note data extracted from a real EHR with ICU clinician notes, radiology \& cardiology study reports, and discharge summaries, but it is missing other notes such as non-ICU inpatient clinical notes and outpatient clinic notes. Each patient may have multiple hospital admissions in the MIMIC-III dataset. As part of the de-identification process, dates in MIMIC-III are randomly shifted to be in a year between 2100-2200 with all relative date and time intervals within a single patient preserved.

The dataset in this study comprised of a random sample of 100 patients from MIMIC-III that meet the following inclusion criteria: 
\begin{enumerate}
    \item The patient’s last hospital admission must have a discharge summary with Brief Hospital Course (BHC) section that can be extracted using regular expressions.
    \item The last hospital admission should have at least 2 physician notes other than the discharge summary.
    \item The patient’s EHR must contain at least 10 total notes prior to the discharge summary, which may be from current admission or prior hospital encounters.
\end{enumerate}

\subsubsection{EHR Reference Creation}\label{methods_sub2sub2}
Patients may have multiple hospital admissions in their EHR, so the last hospital admission was selected as the hospital admission of interest. For each patient, the discharge summary for the target hospital admission was separated from the rest of the clinical notes. The clinical notes contain factual knowledge known about the patient that was entered into the chart prior to the end of the hospital admission when a clinician would typically write a discharge summary. Thus the rest of the patient’s clinical notes served as a patient-specific reference that can be used to verify facts written in a BHC.

\subsubsection{Human-written Brief Hospital Course}\label{methods_sub2sub3}
A human-written BHC was extracted from the discharge summary of the target hospital admission using regular expressions for each of the 100 patients. This long-form human-written narrative represents the current status quo in clinical care where human clinicians obtain data from chart review and care team discussions to generate new clinical text. This narrative was written by a clinician who likely had access to additional data sources than what is present in the patient-specific reference. This is due to MIMIC-III consisting mainly of ICU notes, radiology reports, and cardiology reports, and because not all bedside patient care discussions between clinical staff and patients are captured in EHR clinical notes.

\subsubsection{LLM-written Brief Hospital Course}\label{methods_sub2sub4}
An LLM-written BHC or LLM-written summary was also created for each of the 100 patients by using LLMs to iteratively summarize all the available notes from the patient’s current admission with the exception of the original human-written discharge summary. The LLM-written narrative represents a use case where LLMs are used to generate text in clinical care applications. The LLM-written summary was created using the following procedure: 
\begin{enumerate}
    \item Each note from the target hospital admission was selected to be source material for the final LLM-written summary.
    \item Each note was split into text chunks of 1000 or fewer tokens (1000 tokens is roughly 750 words). Each text chunk was summarized into 4 or fewer sentences using an LLM. The summaries were subsequently combined using an LLM to yield a summary of each note with a length of 4 or fewer sentences. This tree-summarization scheme for each clinical note ensures a balance between high-level summary and detail.
    \item A rolling update summarization scheme was then applied across all individual note summaries to yield a single summary. Note summaries were first ordered in chronological order and gathered into a series of 5000-token note summary bundles such that each bundle contained summarized versions of 5 or more consecutive notes. The first 5000-token note summary bundle was summarized into an initial target LLM-written summary. All subsequent note summary bundles were iteratively merged and aggregated into the target LLM-written summary. This process was repeated until all 5000-token note summary bundles were consumed.
    \item Whenever the target LLM-written summary exceeded 1000 tokens in length, an LLM was used to rewrite and compress the summary while attempting to maintain the most important details. This ensured the final LLM-written summary never exceeded the length of 1000 tokens.
    \item The final LLM-written summary is one of the studied input text types for \textit{VeriFact}. Prompts used to generate the LLM-written summary are depicted in \hyperref[supfigs_sub2]{Appendix A.2}.
\end{enumerate}

The described procedure ensures the LLM-written summaries are able to be generated from a set of EHR clinical notes no matter the length of each clinical note or how many clinical notes were written during the patient’s hospital admission. The procedure also enables LLMs to focus on details in each subsection of every note while providing LLM’s the autonomy to discard what it considers to be less important information. The LLM-written summaries were intended to represent a real-world clinical use case where LLMs are used to generate text for clinical care scenarios. However, no significant effort was made to further optimize the content of LLM-written summaries since clinical summary-generation was not the focus of this investigation.

\subsubsection{Formation of the Annotated Proposition Dataset}\label{methods_sub2sub5}
Both LLM-written and human-written BHC narratives were decomposed into propositions and these propositions are annotated by human clinicians to create human clinician ground truth labels which \textit{VeriFact} can be compared against. The text transformation process is the same for BHC narratives being evaluated and the EHR clinical notes which are used to compose reference contexts. The transformation pipeline is described in detail in the following paragraphs. Once propositions are obtained, verdict annotations of \textit{Supported}, \textit{Not Supported}, and \textit{Not Addressed} for each proposition were created by human clinicians as well as the \textit{VeriFact} system, which are described in the subsequent steps. Additionally, the validity of each proposition based on first-order formal logic was assessed. All of these proposition annotations form the final \textit{VeriFact-BHC} dataset.

\subsection{Forming Propositions and Facts}\label{methods_sub3}
\subsubsection{Transformation of Text to Propositions and Facts}\label{methods_sub3sub1}
\textit{VeriFact} performs evaluation on text inputs such as the human-written or LLM-written summaries by transforming the text into proposition units for more fine-grained fact checking. Both atomic claim and sentence statements are considered as a unit of information representation for propositions and facts. Atomic claims and sentences were referred to as propositions if they originate from the candidate input text being evaluated by \textit{VeriFact}; they were referred to as facts if they were extracted from the patient’s EHR clinical notes and stored in the vector database for reference context formation.

The same text transformation pipeline was used to extract propositions from candidate input text being evaluated by \textit{VeriFact} as was used to extract facts from each of the patient’s EHR notes:
\begin{enumerate}
    \item Input text is split into text chunks of 128 or fewer tokens (roughly 100 words) using recursive semantic parsing. This involved creating an embedding representation for each sliding window of 3 sentences, computing cosine distance between sequential embedding representations, and generating text splits when the embedding representations exceed the 90th percentile of distances within the input text. The dense encoder from the \textit{M3 Embedding} model was used to generate embedding representations. If the resultant text chunk exceeded 128 tokens, this process was recursively repeated to ensure all final text chunks were 128 tokens or fewer.
    \item To obtain sentence propositions or facts, each text chunk was split using the NLTK sentence tokenizer\cite{Bird2009-nr}. Each resultant sentence chunk was used as a proposition or fact in the \textit{VeriFact} framework.
    \item To obtain atomic claim propositions or facts, each text chunk was passed to an LLM with a prompt for atomic claim extraction. The prompt defines atomic claims as simple sentences that resemble first-order predicate logic statements and encapsulate a single Subject-Object-Predicate relationship. Each atomic claim was then used as a proposition or fact in the \textit{VeriFact} framework. Prompts used in atomic claim extraction are in \hyperref[supfigs_sub3]{Appendix A.3}.
\end{enumerate}

In all experiments, the proposition and fact representation type (sentences versus atomic claims) are always chosen to be the same. For example, if the input text is decomposed into atomic claim propositions, the corresponding reference context of EHR facts is composed of retrieved atomic claim facts and not sentence facts.

\subsubsection{Determining the Validity of Propositions}\label{methods_sub3sub2}

In first-order formal logic, valid propositions must assert a specific claim about the state of the world and have a truth-value–either true or false; as a result, valid propositions are often declarative statements that assert something about a subject\cite{Russell2009-tm, Smith2012-us}. Invalid propositions involve statements where a truth value cannot be properly assigned, which includes imperative statements (instructions), interrogative statements (questions), incomplete statements that fail to assert a complete thought, and vague statements that do not identify a specific subject (e.g. missing subject or ambiguous pronouns). Since truth values cannot be properly assigned to invalid propositions there is intrinsic ambiguity and disagreement in assignment of the verdict labels: \textit{Supported}, \textit{Not Supported}, and \textit{Not Addressed}.

Invalid propositions were occasionally produced as an unintended byproduct of either sentence or atomic claim extraction. To investigate how often these invalid propositions occurred, we employed a LLM-assisted approach: 
\begin{enumerate}
    \item A LLM-logician was created by prompting a LLM (\textit{temperature=0.1}) to role-play as a logician. The LLM-logician was tasked to perform binary classification for whether a proposition contained an imperative statement, interrogative statement, incomplete statement, or vague statement. Prompts used for proposition validity classification are in \hyperref[supfigs_sub5]{Appendix A.5}.
    \item A clinician then manually reviewed all the classified outputs from the LLM-logician with the ability to override and correct the assigned label if it was incorrect.
    \item The union of classification labels was used to determine whether a proposition was valid. If the proposition was classified as an imperative statement, interrogative statement, incomplete statement, or vague statement, then it was considered an invalid proposition. Otherwise it was a valid proposition.
\end{enumerate}

This study relaxed some of the strict definitions for claims and propositions in formal logic due to idiosyncrasies in EHR clinical notes and the MIMIC-III dataset. Specifically, redacted name placeholders in the de-identified MIMIC-III dataset are assumed to be specific names or dates and are not a primary reason for propositions to be invalid. Special grammatical and style conventions used in the EHR are accepted as valid and grammatical incorrectness was not a primary reason for propositions to be invalid as long as a clinician can discern a complete claim in the proposition. Since the EHR is a descriptive record of the patient, if a proposition lacks an explicit subject but the predicate and object relate to the patient, the subject is assumed to be the patient. If a proposition contains multiple claims, we consider the proposition valid if at least one of the claims is valid.

\subsection{Human Clinician Annotator Study}\label{methods_sub4}
\subsubsection{Annotation, Adjudication and Ground Truth Creation}\label{methods_sub4sub1}
Each proposition was annotated by three clinicians who assigned a verdict label of \textit{Supported}, \textit{Not Supported}, or \textit{Not Addressed}. \textit{Supported} propositions must be fully supported by the patient’s EHR. \textit{Not Supported} propositions are not supported, partially supported, or contradicted by the EHR. \textit{Not Addressed} propositions are not mentioned in the EHR.

A majority vote was then taken to denoise the labels and establish a human clinician ground truth set of labels for each proposition. In scenarios where all three clinicians disagreed, propositions were independently reviewed by two additional clinicians and a majority vote was taken from their annotations. Any further disagreement was manually discussed and adjudicated by the two clinicians and the addition of a third clinician to break ties, ultimately yielding a consensus label. The initial annotations were created by 25 clinicians with varying levels of training, experience and specialties who were each randomly assigned to 12 patients and tasked with labeling all propositions for those 12 patients. Adjudication of disagreeing propositions was conducted by 2 additional clinicians. The ultimate goal was to create a human ground truth for each proposition that approximates the judgment of the average clinician, which can serve as a gold standard comparison for \textit{VeriFact}.

Since the MIMIC-III dataset was de-identified and all clinician annotators were study authors with no recruitment of study subjects, this investigation was not human subjects research and did not require institutional review board approval.

\subsubsection{Annotation Workflow}\label{methods_sub4sub2}
Clinician annotators performed labeling using \textit{Label Studio}\cite{UnknownUnknown-nb} with a custom annotation interface and were tasked to assign a \textit{Supported}, \textit{Not Supported}, or \textit{Not Addressed} label to each proposition (\hyperref[supfigs_sub7sub1]{Appendix A.7.1} \& \hyperref[supfigs_sub7sub2]{Appendix A.7.2}). Clinician annotators were presented with the propositions one at a time and were not explicitly informed whether the proposition was an atomic claim or sentence proposition. Clinician annotators also did not have access to the original BHC narrative from which propositions were derived and were blinded to whether the proposition was derived from a human-written or LLM-written Brief Hospital Course narrative.

Annotators were provided each patient’s EHR clinical notes in Portable Document Format (PDF) document format, which contained all clinical notes for the patient with the exception of the final discharge summary (\hyperref[supfigs_sub7sub3]{Appendix A.7.3} \& \hyperref[supfigs_sub7sub4]{Appendix A.7.4}). This EHR PDF contains the same source text from which facts stored in the EHR Vector Database are derived, thereby providing human clinician annotators access to the same information as the \textit{VeriFact} system. The EHR PDF was generated using the \textit{fpdf} version 2.7.9 python library. To approximate how clinicians perform chart review in modern real-world EHR systems, clinicians were allowed to use any PDF reader software to read and search the EHR PDF for information, including text search. The amount of time taken to complete the annotation task was recorded by \textit{Label Studio}.

\subsection{The \textit{VeriFact} System}\label{methods_sub5}
\subsubsection{Vector Search Methods for Fact Retrieval}\label{methods_sub5sub1}
Each patient’s clinical notes were transformed into sets of atomic claim facts and sentence facts in the same manner that propositions were derived. Facts were then stored in a vector database for subsequent retrieval. \textit{LlamaIndex} version 0.11.14 was used as a framework for RAG and \textit{Qdrant} version 1.10 was used as the vector database with default indexing and search settings. \textit{Qdrant} employs a filtrable hierarchical small world navigation search algorithm, which provides a mechanism to restrict retrieval of facts to those derived from clinical notes from the current hospital admission versus all admissions in the EHR\cite{Malkov2016-tb}.

We explored three methods for retrieving EHR facts from the vector database: 
\begin{itemize}
    \item \textbf{Dense:} The proposition was encoded into a vector query by the \textit{M3 Embedding} model’s dense encoder and dot-product similarity was used to determine the most relevant facts to retrieve for each proposition.
    \item \textbf{Hybrid:} Dense search with the addition of performing a parallel sparse search by encoding the proposition using the \textit{M3 Embedding} model’s sparse token encoder. Results from dense and sparse search methods were subsequently fused using distribution-based score fusion\cite{Mazzeschi2023-ov}.
    \item \textbf{Rerank:} Hybrid search followed by the \textit{M3 Reranker} to rescore and rerank the search results discovered by Hybrid search.
\end{itemize}

\subsubsection{Number of Facts Retrieved}\label{methods_sub5sub2}
We studied how \textit{VeriFact} performance was affected by the number of EHR facts used in the reference context passed to the LLM-as-a-Judge. This is the top \textit{N} parameter for which we studied \textit{N = 5, 10, 25, and 50} facts included in the reference context. 

Hybrid and rerank retrieval methods produce more search results than dense retrieval, so we utilized the following scheme to equalize the number of facts in the reference context to enable direct comparison between the retrieval methods. Let \textit{K} be the number of facts retrieved from either a dense or sparse search query. Let \textit{N} be the top \textit{N} items selected from the final search results which will be included in the reference context. The mechanism for selecting the number of facts for the reference context for each retrieval method is as follows:
\begin{itemize}
    \item \textbf{Dense:} \textit{N=K} and all \textit{K} retrieved facts from the EHR vector database are considered for the reference context. It is commonplace for EHR note text to be templated or copy-pasted, resulting in many duplicated facts stored in the EHR Vector Database. To avoid much of the reference context being duplicate text, exact string match facts are deduplicated prior to inserting the final list of facts into the reference context.
    \item \textbf{Hybrid:} \textit{N=K} for dense retrieval and \textit{N=K} for sparse retrieval, resulting in \textit{2K} search results. These retrievals are independent of one another, meaning that some retrieved facts may be present in both search result lists. During distribution-based score fusion these search results become deduplicated, meaning that the final fused search result list had $\leq$\textit{2K} facts. Similar to the dense method, exact string match facts were then deduplicated. After deduplication, only the top \textit{N} facts ranked by the distribution-based score fusion method were used in the reference context with all other facts discarded.
    \item \textbf{Rerank:} Same as hybrid, but after the final fused search result list of length $\leq$\textit{2K} facts is obtained and exact string match facts are deduplicated, the \textit{M3 Reranker} model was applied to rescore each fact’s relevance to the proposition. This resulted in re-ordering of the search results. Then the top N facts ranked by the \textit{M3 Reranker} were used in the reference context with all other items discarded.
\end{itemize}

\subsubsection{Reference Context Formation}\label{methods_sub5sub3}
After the set of top \textit{N} facts was finalized, we explored different ways to format the reference context. Scores generated from retrieval serve as a way to rank how relevant each retrieved fact is for each proposition. However, the EHR is a temporal record and a list of facts sorted by relevance score can result in historical and outdated information ranked higher in the list of facts than newer and more pertinent information. Since it was unclear how to optimally arrange the top \textit{N} facts, we studied the following three approaches: 
\begin{itemize}
    \item \textbf{Relevance Score:} The top \textit{N} facts were ordered by relevance score, which was dot product similarity for dense retrieval, distribution-based score fusion score for hybrid retrieval, and M3 Reranker score for rerank retrieval. Facts were formatted as a numbered list with fact text presented alongside retrieval score, note category and note description metadata of the clinical note where the fact originated.
    \item \textbf{Absolute Time:} The top \textit{N} facts were ordered chronologically and formatted as a numbered list with fact text and the timestamp, note category \& note description metadata of the clinical note where the fact originated. To contextualize the timestamped facts, the hospital admission start and hospital admission end timestamps were pre-pended to the reference context.
    \item \textbf{Relative Time:} Same as absolute time, but all timestamps were converted into relative times with the current time set to the hospital admission end. Timestamps in MIMIC-III are randomly shifted to occur between years 2100 and 2200 as part of the de-identification process, resulting in unnatural timestamps that may be outside of the training distribution for LLMs and may result in abnormal LLM behavior. Conversion of absolute timestamps to relative time in hours and days before the end of hospital admission removes the effect of unnatural timestamps and provides a clearer chronology of events.
\end{itemize}
Representative examples of the different types of reference contexts are depicted in \hyperref[supfigs_sub8]{Appendix A.8}.

\subsubsection{LLM-as-a-Judge}\label{methods_sub5sub4}
Each proposition and its reference context of facts retrieved from the EHR was then passed to an LLM-as-a-Judge. The LLM-as-a-Judge was instructed to assign a verdict on whether the proposition was \textit{Supported}, \textit{Not Supported}, or \textit{Not Addressed} by the reference context along with a reason explaining why the verdict was assigned. After verdicts are assigned to all propositions derived from input text, the percentage of each label is computed and the reasons for each verdict were aggregated into a summary. This design is inspired by prior fact evaluation work leveraging text decomposition and RAG evaluation\cite{UnknownUnknown-ku, Min2023-zk, Wei2024-ur}. This results in the \textit{VeriFact} output which includes:
\begin{enumerate}
    \item A high-level score sheet denoting the proportion of the input text that is \textit{Supported}, \textit{Not Supported}, or \textit{Not Addressed} and a high-level summary explaining aspects of the input text that contribute to each label.
    \item Detailed verdicts and explanations for individual propositions.
\end{enumerate}
A depiction of these \textit{VeriFact} outputs is shown in \hyperref[fig:figure2]{Figure 2}. Prompts used in the LLM-as-a-Judge to generate the outputs are shown in \hyperref[supfigs_sub4]{Appendix A.4}.

\subsection{Statistical Analysis}\label{methods_sub6}
Inter-rater agreement for the 3-label proposition classification task (\textit{Supported}, \textit{Not Supported}, \textit{Not Addressed}) and binarized labels (\textit{Supported}, \textit{Not Supported or Addressed}) is measured using Percentage Agreement as the primary metric and Gwet’s Agreement Coefficient 1 (AC1) as a secondary metric. Percentage Agreement is the average agreement between all annotator pairs for all propositions and is not a chance-corrected metric, but is simple to interpret. Gwet’s AC1 is a chance-corrected agreement coefficient that is relatively insensitive to label imbalance and the kappa paradox\cite{Gwet2008-ga}. Percentage Agreement is computed in python and Gwet’s AC1 is computed using \textit{irrCAC} version 0.4.3 python package with 95\% confidence intervals estimated using 1000 bootstrap iterations. The same agreement metrics are applied to measure human clinician inter-annotator agreement as well as agreement between the denoised human clinician ground truth labels and each hyperparameter variation of the \textit{VeriFact} system. Once the best \textit{VeriFact} system was identified using Agreement Percentage and Gwet’s AC1, sensitivity (true positive rate), specificity (true negative rate, recall), positive predictive value (PPV, precision), and negative predictive value (NPV) are calculated for each label using the denoised human clinician ground truth labels as a gold standard reference.

\section*{Data Availability}\label{data_availability}
The \textit{VeriFact-BHC} dataset used in this study is derived from the MIMIC-III Clinical Database v1.4 and is made publicly available on Physionet per credentialed access policy and data use agreement.

\section*{Code Availability}\label{code_availability}
Code for the \textit{VeriFact} system, \textit{VeriFact-BHC} dataset creation, and all experimental analysis is publicly available at \href{https://github.com/philipchung/verifact}{https://github.com/philipchung/verifact}

\section*{Acknowledgments}\label{acknowledgements}
P.C. is supported by the Foundation for Anesthesia Education and Research (FAER) Mentored Research Training Grant. This investigation is also supported by Anesthesia Research Grant from Stanford Department of Anesthesiology, Perioperative \& Pain Medicine, and National Institutes of Health R35GM138353. We thank all members in Nima Aghaeepour's laboratory and members in Nigam Shah’s laboratory for early discussions and feedback related to the project. We thank Akeem Daly, Gerald Wen, and Kelvin Leung from Stanford Department of Anesthesiology, Perioperative \& Pain Medicine IT for assistance with computing infrastructure.

\section*{Competing Interests}\label{competing_interests}
N.A. is a member of the Scientific Advisory Boards of January AI, Parallel Bio, Celine Therapeutics, WellSim Biomedical Technologies, and Medeloop AI, is a cofounder of Takeoff, and is a paid consultant for MaraBio Systems. E.Y.W. is on the board of Invincikids, a nonprofit organization that seeks to distribute immersive technologies to improve pediatric care. She receives no compensation for her role.

\section*{Author Contributions}\label{author_contributions}
P.C. contributed to study conception and design, wrote code for software and dataset creation, contributed to dataset annotation, analyzing and interpreting the data, drafting and revising the manuscript; A.S., A.J.G., Y.K., S.M.R, L.H., B.D., M.A.S., A.A., M.G., D.S., A.A.L., C.E.C., B.B. M.A.S., H.J.H., T.P.N., M.R.R., K.K., M.A.B., J.C.M., R.S., S.P.M., D.D., J.X., E.Y.W., C.A.S. all  contributed to dataset annotation, data interpretation, and revising the manuscript; N.S. and N.A. contributed to study design, data analysis, data interpretation, and revising the manuscript.

%%=======================%%
%% Supplementary Figures %%
%%=======================%%

\clearpage
\begin{appendices}
\section{Supplementary Figures}\label{supfigs}

\subsection{Decomposition of LLM-written Brief Hospital Course}\label{supfigs_sub1}

\subsubsection{LLM-written Brief Hospital Course}\label{supfigs_sub1sub1}
\begin{lstlisting}[style=prompt]
The patient, a 77-year-old male, was admitted with a diagnosis of acute cholangitis, presenting with nausea, vomiting, and abdominal pain, and a history of coronary artery disease, atrial fibrillation, cardiomyopathy, and chronic obstructive pulmonary disease. Initial imaging revealed gallstones, and the patient underwent a series of interventions, including fluid boluses, IV antibiotics, and a central line placement. Due to worsening septic shock, the patient required vasopressors and was intubated. An ERCP procedure was performed, which revealed pus, sludge, and stones in the common bile duct, and a stent was placed for drainage. Post-procedure, the patient's condition improved, with stabilization of blood pressure, improvement in abdominal pain, and eventual weaning off vasopressors. The patient was extubated and transferred to the floor, where he continued to receive treatment for sepsis, acute renal failure, and pain control, with plans for repeat ERCP in 4 weeks.
\end{lstlisting}
An example of an LLM-written Brief Hospital Course text that is generated by the LLM-writer, which utilizes all of the clinical notes from the latest hospital admission to generate the summary. This is the same LLM-written Brief Hospital Course text shown in Figure \hyperref[fig:figure2]{Figure 2}.

\subsubsection{Sentence Propositions from LLM-written Brief Hospital Course}\label{supfigs_sub1sub2}
\begin{lstlisting}[style=prompt_linenum]
The patient, a 77-year-old male, was admitted with a diagnosis of acute cholangitis, presenting with nausea, vomiting, and abdominal pain, and a history of coronary artery disease, atrial fibrillation, cardiomyopathy, and chronic obstructive pulmonary disease.
Initial imaging revealed gallstones, and the patient underwent a series of interventions, including fluid boluses, IV antibiotics, and a central line placement.
Due to worsening septic shock, the patient required vasopressors and was intubated.
An ERCP procedure was performed, which revealed pus, sludge, and stones in the common bile duct, and a stent was placed for drainage.
Post-procedure, the patient's condition improved, with stabilization of blood pressure, improvement in abdominal pain, and eventual weaning off vasopressors. 
The patient was extubated and transferred to the floor, where he continued to receive treatment for sepsis, acute renal failure, and pain control, with plans for repeat ERCP in 4 weeks.
\end{lstlisting}
The corresponding decomposition of the LLM-written Brief Hospital Course into sentence propositions.

\subsubsection{Atomic Claim Propositions from LLM-written Brief Hospital Course}\label{supfigs_sub1sub3}
\begin{lstlisting}[style=prompt_linenum]
The patient was a 77-year-old male.
The patient was admitted with a diagnosis of acute cholangitis.
The patient presented with nausea.
The patient presented with vomiting.
The patient presented with abdominal pain.
The patient had a history of coronary artery disease.
The patient had a history of atrial fibrillation.
The patient had a history of cardiomyopathy.
The patient had a history of chronic obstructive pulmonary disease.
Initial imaging revealed gallstones.
The patient underwent fluid boluses.
The patient underwent IV antibiotics.
The patient underwent a central line placement.
The patient required vasopressors due to worsening septic shock.
The patient was intubated due to worsening septic shock.
An ERCP procedure was performed.
The ERCP procedure revealed pus in the common bile duct.
The ERCP procedure revealed sludge in the common bile duct.
The ERCP procedure revealed stones in the common bile duct.
A stent was placed for drainage.
The patient's condition improved post-procedure.
The patient's blood pressure stabilized post-procedure.
The patient experienced improvement in abdominal pain post-procedure.
The patient was weaned off vasopressors post-procedure.
The patient was extubated post-procedure.
The patient was transferred to the floor post-procedure.
The patient continued to receive treatment for sepsis on the floor.
The patient continued to receive treatment for acute renal failure on the floor.
The patient continued to receive treatment for pain control on the floor.
The patient had plans for repeat ERCP in 4 weeks.
\end{lstlisting}
The corresponding decomposition of the LLM-written Brief Hospital Course into atomic claim propositions.

\clearpage
\subsection{Prompts for Generating LLM-written Brief Hospital Course}\label{supfigs_sub2}
Prompts used to transform all clinical notes from a hospital admission to an LLM-written Brief Hospital Course summary. The same system prompt is used across all LLM invocations for the LLM-writer. Prompts are Jinja templates with \{\{text\}\} brackets indicating variables which are derived from the dataset or an intermediate text output.
\subsubsection{LLM-writer System Prompt}\label{supfigs_sub2sub1}
\begin{lstlisting}[style=prompt]
You are a physician who takes care of patients in a hospital.
\end{lstlisting}

\subsubsection{Prompt for Note Chunk Summarization}\label{supfigs_sub2sub2}
\begin{lstlisting}[style=prompt]
## Task:
Summarize the text in 1-4 sentences making sure to include the key points. Return only the summary text. Do not state it is a summary.

## Text:
{{text}}
    
## Summary:
\end{lstlisting}

\subsubsection{Prompt for Combining Note Chunk Summaries}\label{supfigs_sub2sub3}
\begin{lstlisting}[style=prompt]
## Task:
Combine the summaries into a single cohesive summary that is at most 4 sentences long, making sure to include the key points. Return only the combined summary text. Do not state it is a combined summary.

## Summaries:
{{text}}

## Combined Summary:
\end{lstlisting}

\subsubsection{Prompt for Writing Initial Brief Hospital Course}\label{supfigs_sub2sub4}
\begin{lstlisting}[style=prompt]
## Task:
Write a "Brief Hospital Course" section for a patient's discharge summary using the provided information from the patient's electronic health record from this hospital admission. Respond only with content for the "Brief Hospital course" section without any elaboration or explanation.

## Information from Patient's Electronic Health Record:
{{text}}

## Brief Hospital Course:
\end{lstlisting}

\subsubsection{Prompt for Refining Brief Hospital Course with New Information}\label{supfigs_sub2sub5}
\begin{lstlisting}[style=prompt]
## Task:
You are preparing a patient for discharge from the hospital and have written part of the "Brief Hospital Course" section for the patient's discharge summary. Refine the "Brief Hospital Course" section by incorporating the new information from the patient's electronic health record. Respond only with content for the "Brief Hospital course" section without any elaboration or explanation.

## Existing Draft of "Brief Hospital Course":
{{brief_hospital_course}}

## New Information from Patient's Electronic Health Record:
{{new_text}}

## Refined "Brief Hospital Course":
\end{lstlisting}

\subsubsection{Prompt for Compacting Brief Hospital Course}\label{supfigs_sub2sub6}
\begin{lstlisting}[style=prompt]
## Task:
You are preparing a patient for discharge from the hospital and have written a "Brief Hospital Course" section for the patient's discharge summary. However, the text is too long. Please make it more concise while retaining the key points.

## Existing Draft of "Brief Hospital Course":
{{brief_hospital_course}}

## Refined "Brief Hospital Course":
\end{lstlisting}

\clearpage
\subsection{Prompts for Atomic Claim Extraction in \textit{VeriFact} Pipeline}\label{supfigs_sub3}
Prompts used to detect and extract atomic claims from text chunks in order to create propositions from the input candidate text being evaluated and facts from the EHR clinical notes. The same system prompt is used across all LLM invocations related to atomic claim extraction. Prompts are Jinja templates with \{\{text\}\} brackets indicating variables. In these prompts, \{\{text\}\} is replaced with a text chunk generated by the semantic text splitter which splits candidate input text and EHR notes into smaller text chunks.

\subsubsection{Atomic Claim Extraction System Prompt}\label{supfigs_sub3sub1}
\begin{lstlisting}[style=prompt]
You are a physician tasked with extracting atomic claims from different text reference sources to represent the information in the text as a list of atomic claims.
\end{lstlisting}

\subsubsection{Prompt for Checking for Presence of Atomic Claim in Text}\label{supfigs_sub3sub2}
\begin{lstlisting}[style=prompt]
## Task Definition
Determine whether the text contains any atomic claims.

## Atomic Claim Definition
An atomic claim is a phrase or sentence that makes a single assertion. The assertion may be factual or may be a hypothesis posed by the text. Atomic claims are indivisible and cannot be decomposed into more fundamental claims. More complex facts and statements can be composed from atomic facts. Atomic claims should have a subject, object, and predicate. The predicate relates the subject to the object.

## Detailed Instructions
1. Read the text from the "text" key in the input JSON object. Determine if the text contains any atomic claims as defined above. If the text contains at least one atomic claim, output "contains_atomic_claim" as True. If the text does not contain any atomic claims, output "contains_atomic_claim" as False.
2. Format your response as a JSON object with "claims" key with the boolean value. Return only JSON. No explanation is needed.

## Examples
Input JSON: 
{
    "text": "Einstein won the noble prize in 1968 for his discovery of the photoelectric effect."
}
Output JSON: 
{
    "contains_atomic_claim": true
}

Input JSON:
{
    "text":  "I ate a"
}
Output JSON:
{
    "contains_atomic_claim": false
}

## Actual Task
Input JSON:
{
    "text": "{{text}}"
}
Output JSON:
\end{lstlisting}

\subsubsection{Prompt for Extracting Atomic Claims from Text}\label{supfigs_sub3sub3}
\begin{lstlisting}[style=prompt]
## Task Definition
Generate a list of unique atomic claims that can be inferred from the provided text.

## Atomic Claim Definition
An atomic claim is a phrase or sentence that makes a single assertion. The assertion may be factual or may be a hypothesis posed by the text. Atomic claims are indivisible and cannot be decomposed into more fundamental claims. More complex facts and statements can be composed from atomic facts. Atomic claims should have a subject, object, and predicate. The predicate relates the subject to the object.

## Detailed Instructions
1. Extract a list of claims from the text in "text" key from the JSON object. Each claim should have a subject, object, and predicate and conform to the Atomic Claim Definition provided above. Claims should be unambiguous if they are read in isolation, which means you should avoid pronouns and ambiguous references.
2. The list of claims should be comprehensive and cover all information in the text. Claims you extract should include the full context it was presented in, NOT cherry picked facts. You should NOT include any prior knowledge, and take the text at face value when extracting claims.
3. Format all the claims as a JSON object with "claims" key with values as a list of string claims. Return only JSON. No explanation is needed.

## Examples
Input JSON: 
{
    "text": "Einstein won the noble prize in 1968 for his discovery of the photoelectric effect."
}
Output JSON: 
{
    "claims": [
        "Einstein won the noble prize for his discovery of the photoelectric effect.",
        "Einstein won the noble prize in 1968."
    ]  
}

## Actual Task
Input JSON:
{
    "text": "{{text}}"
}
Output JSON:
\end{lstlisting}

\clearpage
\subsection{Prompts for LLM-as-a-Judge in \textit{VeriFact} Pipeline}\label{supfigs_sub4}
Prompts used to assign a verdict label (\textit{Supported}, \textit{Not Supported}, \textit{Not Addressed}) to propositions and generate a reason for the verdict. The same system prompt (A) is used across all LLM invocations related to the LLM-as-a-Judge. Prompts are Jinja templates with \{\{text\}\} brackets indicating variables. 

\subsubsection{LLM-as-a-Judge System Prompt}\label{supfigs_sub4sub1}
\begin{lstlisting}[style=prompt]
You are judging whether text written about a patient is supported by information from the patient's electronic health record.
\end{lstlisting}

\subsubsection{Prompt for Assigning Verdict Label to a Proposition}\label{supfigs_sub4sub2}
\begin{lstlisting}[style=prompt]
## Task:
Using the provided reference from the patient's electronic health record, determine whether the text is supported, not supported, or not addressed by the reference context.

## Detailed Instructions:
1. Compare the text and reference, which is provided in the "text" and "reference" keys of the input JSON object.
2. Provide a verdict in the output JSON object under the key "verdict".
    a. Determine whether the text is addressed or not addressed by the reference context.
    b. If the text is not addressed by the reference context, output the verdict "Not Addressed".
    c. If the text is addressed and fully supported by the reference context, output the verdict "Supported".
    d. If the text is addressed and not fully supported by the reference context, output the verdict "Not Supported". This includes cases where the text is only partially supported by the reference context.
    e. The only valid outputs are "Supported", "Not Supported", or "Not Addressed".
    d. When determining a verdict, consider the information in the reference context to be correct and the source of truth.
3. Provide a reason for your verdict in the output JSON object under the key "reason". Try to make your reason concise and to the point. You must provide a reason. Do not leave it blank.
4. Format your response as a JSON object with the keys "verdict" and "reason". Return only JSON. No explanation is needed.

## Examples
Input JSON:
{
    "text": "Jenny's Hemoglobin A1c indicated she has poorly controlled diabetes."
    "reference": "Jenny's Hemoglobin A1c was 6.8% on admission. Danny has a Hemoglobin A1c of 9.2%."
}
Output JSON:
{
    "verdict": "Not Supported",
    "reason": "The text states that Jenny has poorly controlled diabetes, but her Hemoglobin A1c was 6.8% on admission which is within normal range."
}

Input JSON:
{
    "text": "Mr. Sanchez has severe cardiac disease."
    "reference": "Carlos Sanchez had a LVEF 23% on transthoracic echocardiogram."
}
Output JSON:
{
    "verdict": "Supported",
    "reason": "The text states that Mr. Sanchez has severe cardiac disease which is supported by his transthoracic echocardiogram showing a low LVEF of 23%."
}

Input JSON:
{
    "text": "The patient is allergic to penicillin."
    "reference": "The patient is allergic to latex and cefazolin."
}
Output JSON:
{
    "verdict": "Not Addressed",
    "reason": "The reference context does not mention the patient's allergy to penicillin. The patient is allergic to cefazolin, but this is a cephalosporin antibiotic and not a penicillin and chances of cross-reactivity are extremely low."
}

## Actual Task
Input JSON:
{{json}}
Output JSON:
\end{lstlisting}

In this prompt, \{\{json\}\} is replaced with a JSON object that contains a text proposition and a reference context derived from EHR facts. The generated output JSON contains both a verdict as well as an LLM-generated reason for that verdict.

\subsubsection{Prompt for Generating an Overall Summary for Each Verdict Label}\label{supfigs_sub4sub3}
\begin{lstlisting}[style=prompt]
## Task Definition
Summarize the reasons for which the text was {{label}} by the reference context. Output the summary as a JSON object with the key "summary".

## Reasons
{{json}}

## Summary
\end{lstlisting}
In this prompt, \{\{json\}\} is replaced with a JSON object that includes a list of reasons from all proposition verdicts that were assigned the target label (\textit{Supported}, \textit{Not Supported}, \textit{Not Addressed}) and the output JSON is a summary of those reasons.

\clearpage
\subsection{Prompts for Assessing Proposition Validity}\label{supfigs_sub5}
Prompts used to determine if a proposition is an imperative statement, interrogative statement, incomplete statement, or vague statement. The output from the LLM is used as an initial draft label which is then reviewed by a human clinician who may override the assignment to generate final proposition validity labels.

\subsubsection{Proposition Validity System Prompt}\label{supfigs_sub5sub1}
\begin{lstlisting}[style=prompt]
You are a logician tasked with judging whether a given text contains a valid claim.
\end{lstlisting}

\subsubsection{Determine if Proposition is an Imperative Statement}\label{supfigs_sub5sub2}
\begin{lstlisting}[style=prompt]
## Task Definition
Determine whether the text contains an imperative statement (instruction or command).

## Imperative Statement Definition
Imperative statements give orders or instructions, asking someone to perform an action. They do not assert a fact about the world. Imperative statements do not have a truth value; they are neither true nor false. They are simply directives for action.

## Detailed Instructions
1. Read the text from the "text" key in the input JSON object. Determine if the text contains any imperative statements as defined above. If the text contains at least one imperative statement, output "contains_imperative_statement" as True. If the text does not contain any imperative statements, output "contains_imperative_statement" as False.
2. Format your response as a JSON object with "contains_imperative_statement" key with the boolean value. Return only JSON. No explanation is needed.

## Examples
Input JSON: 
{
    "text": "Einstein won the noble prize in 1968 for his discovery of the photoelectric effect."
}
Output JSON: 
{
    "contains_imperative_statement": false
}

Input JSON:
{
    "text":  "Close the door!"
}
Output JSON:
{
    "contains_imperative_statement": true
}

Input JSON:
{
    "text":  "Read this book."
}
Output JSON:
{
    "contains_imperative_statement": true
}

## Actual Task
Input JSON:
{
    "text": "{{text}}"
}
Output JSON:
\end{lstlisting}

\subsubsection{Determine if Proposition is an Interrogative Statement}\label{supfigs_sub5sub3}
\begin{lstlisting}[style=prompt]
## Task Definition
Determine whether the text contains an interrogative statement (question).

## Interrogative Statement Definition
Questions ask for information or clarification, and are not making any assertions about the world. Questions cannot be evaluated as true or false; they are requests for information rather than claims about the state of affairs.

## Detailed Instructions
1. Read the text from the "text" key in the input JSON object. Determine if the text contains any interrogative statements as defined above. If the text contains at least one interrogative statement, output "contains_interrogative_statement" as True. If the text does not contain any interrogative statements, output "contains_interrogative_statement" as False.
2. Format your response as a JSON object with "contains_interrogative_statement" key with the boolean value. Return only JSON. No explanation is needed.

## Examples
Input JSON: 
{
    "text": "Einstein won the noble prize in 1968 for his discovery of the photoelectric effect."
}
Output JSON: 
{
    "contains_interrogative_statement": false
}

Input JSON:
{
    "text":  "Is the door open?"
}
Output JSON:
{
    "contains_interrogative_statement": true
}

Input JSON:
{
    "text":  "Who is the president?"
}
Output JSON:
{
    "contains_interrogative_statement": true
}

## Actual Task
Input JSON:
{
    "text": "{{text}}"
}
Output JSON:
\end{lstlisting}

\subsubsection{Determine if Proposition is an Incomplete Statement}\label{supfigs_sub5sub4}
\begin{lstlisting}[style=prompt]
## Task Definition
Determine whether the text contains an incomplete or fragmentary statement.

## Incomplete Statement Definition
Incomplete or fragmentary sentences do not make a full assertion and therefore cannot be evaluated as true or false. These sentences lack a complete thought or claim, so they cannot be judged true or false without additional context.

## Detailed Instructions
1. Read the text from the "text" key in the input JSON object. Determine if the text contains any incomplete statements as defined above. If the text contains at least one incomplete statement, output "contains_incomplete_statement" as True. If the text does not contain any incomplete statements, output "contains_incomplete_statement" as False.
2. Format your response as a JSON object with "contains_incomplete_statement" key with the boolean value. Return only JSON. No explanation is needed.

## Examples
Input JSON: 
{
    "text": "Einstein won the noble prize in 1968 for his discovery of the photoelectric effect."
}
Output JSON: 
{
    "contains_incomplete_statement": false
}

Input JSON:
{
    "text":  "Because he was late."
}
Output JSON:
{
    "contains_incomplete_statement": true
}

Input JSON:
{
    "text":  "When the sun sets."
}
Output JSON:
{
    "contains_incomplete_statement": true
}

Input JSON:
{
    "text":  "I ate a"
}
Output JSON:
{
    "contains_incomplete_statement": true
}

## Actual Task
Input JSON:
{
    "text": "{{text}}"
}
Output JSON:
\end{lstlisting}

\subsubsection{Determine if Proposition is a Vague Statement}\label{supfigs_sub5sub5}
\begin{lstlisting}[style=prompt]
## Task Definition
Determine whether the text contains an vague statement.

## Vague Statement Definition
Vague statements are ones that cannot be formalized as a claim or proposition in formal logic due to ambiguity, lack of specificity, or unclear subjects or
entities. Vague statements fail to describe the full context of the situation and require additional information or clarification to be properly formalized as a claim in formal logic.

## Detailed Instructions
1. Read the text from the "text" key in the input JSON object. Determine if the text contains any vague statements as defined above. If the text contains at least one vague statement, output "contains_vague_statement" as True. If the text does not contain any vague statements, output "contains_vague_statement" as False.
2. Format your response as a JSON object with "contains_vague_statement" key with the boolean value. Return only JSON. No explanation is needed.

## Examples
Input JSON: 
{
    "text": "Einstein won the noble prize in 1968 for his discovery of the photoelectric effect."
}
Output JSON: 
{
    "contains_vague_statement": false
}

Input JSON:
{
    "text":  "He was late to this."
}
Output JSON:
{
    "contains_vague_statement": true
}

Input JSON:
{
    "text":  "They are scheduled to visit Dr."
}
Output JSON:
{
    "contains_vague_statement": true
}

Input JSON:
{
    "text":  "Found to be supratherapeutic."
}
Output JSON:
{
    "contains_vague_statement": true
}

## Actual Task
Input JSON:
{
    "text": "{{text}}"
}
Output JSON:
\end{lstlisting}

\clearpage
\subsection{Agreement Between Clinician Annotator Pairs}\label{supfigs_sub6}
\subsubsection{Pairwise Annotator Percent Agreement}\label{supfigs_sub6sub1}
\begin{figure}[hbt]
    \centering
    \includegraphics[width=1\linewidth]{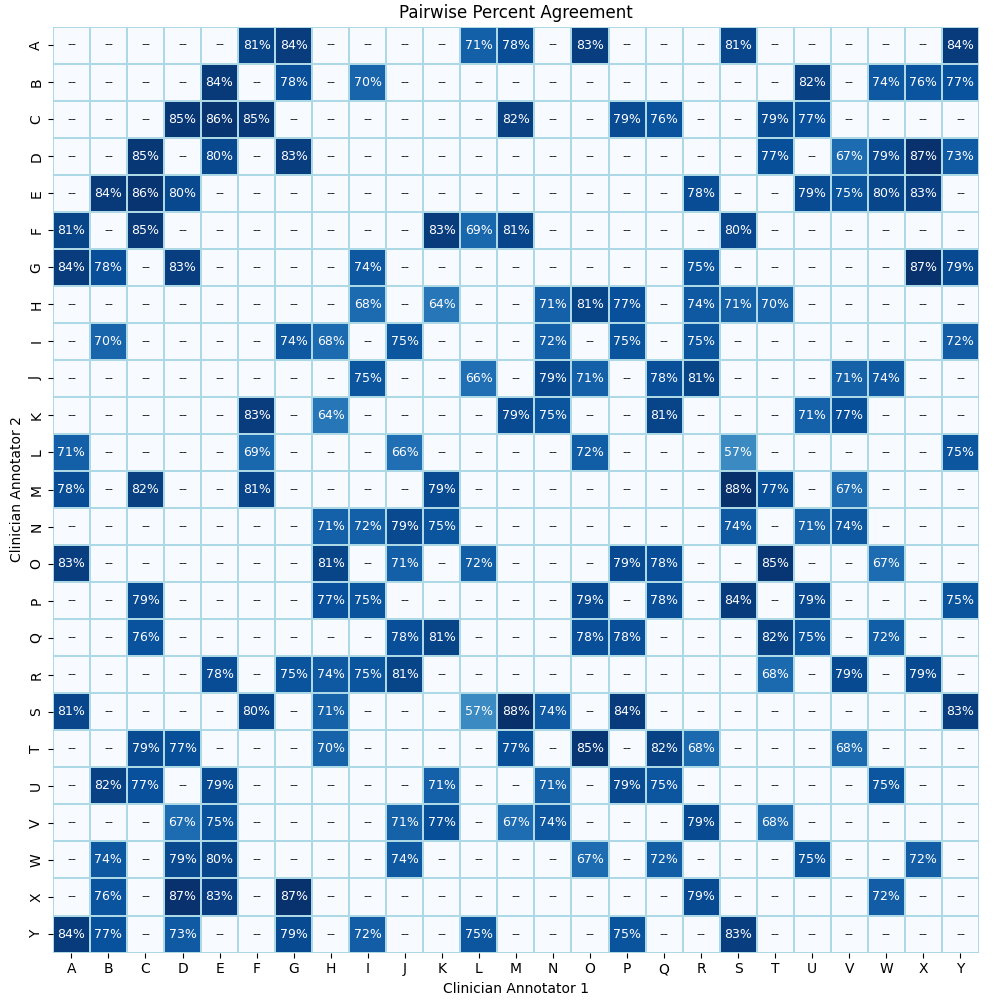}
    \caption{Heatmap showing Percent Agreement between each pair of clinician annotators. Empty cells indicate the two clinicians did not rate the same propositions.}
    \label{fig:supfig6a}
\end{figure}

\clearpage
\subsubsection{Pairwise Annotator Gwet's AC1}\label{supfigs_sub6sub2}
\begin{figure}[hbt]
    \centering
    \includegraphics[width=1\linewidth]{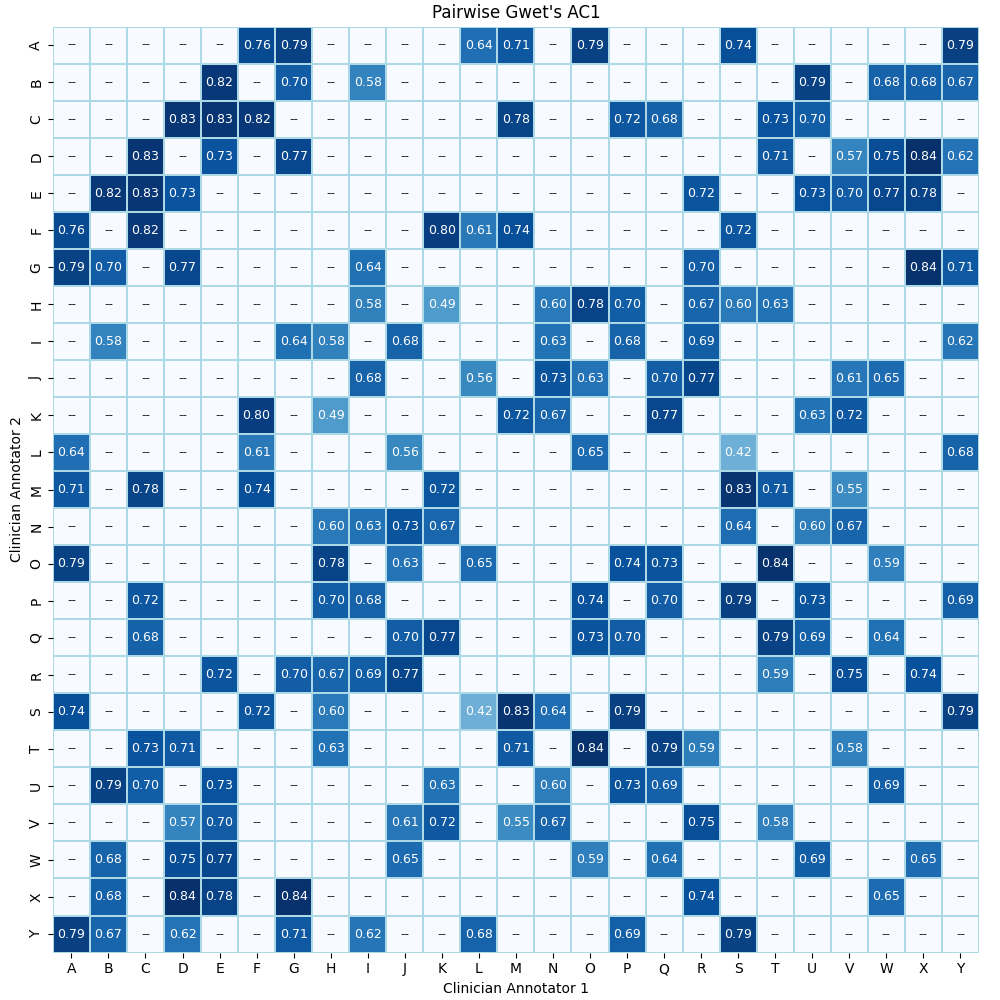}
    \caption{Heatmap showing Gwet’s AC1 between each pair of clinician annotators. Empty cells indicate the two clinicians did not rate the same propositions.}
    \label{fig:supfig6b}
\end{figure}

\clearpage
\subsection{Labeling Guidelines \& Annotation Interface}\label{supfigs_sub7}

\subsubsection{Labeling Interface}\label{supfigs_sub7sub1}
\begin{figure}[hbt]
    \centering
    \includegraphics[width=1\linewidth]{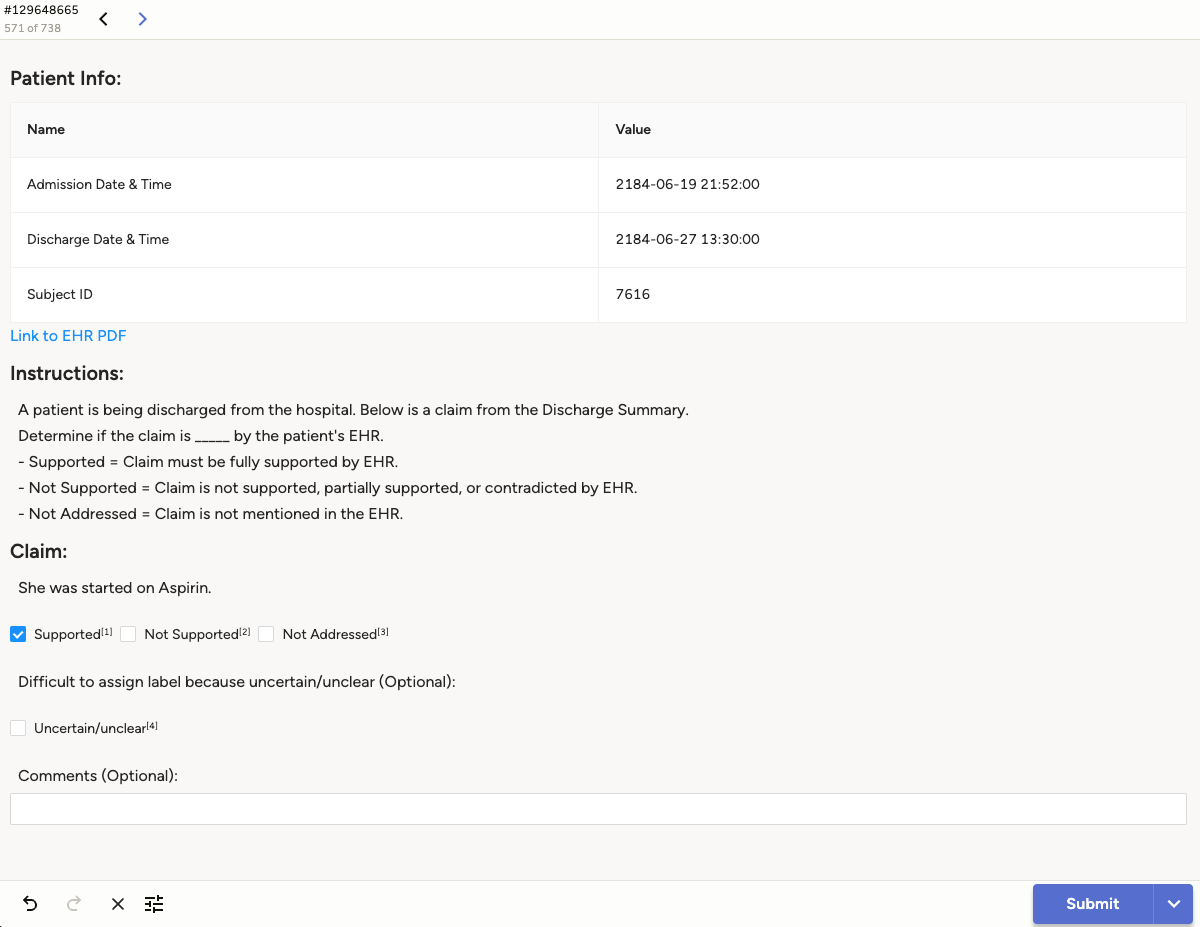}
    \caption{Custom labeling interface created within \textit{Label Studio}. Clinician annotators were required to provide a label on whether the proposition/claim was \textit{Supported}, \textit{Not Supported}, or \textit{Not Addressed}. Each proposition/claim was independently considered and clinician annotators were instructed to not assume any dependence between adjacent claims. They were optionally allowed to mark whether they found the claim “Uncertain/unclear” and free-text comments. EHR = Electronic Health Record.}
    \label{fig:supfig7a}
\end{figure}

\clearpage
\subsubsection{Labeling Algorithm}\label{supfigs_sub7sub2}
\begin{figure}[hbt]
    \centering
    \includegraphics[width=0.5\linewidth]{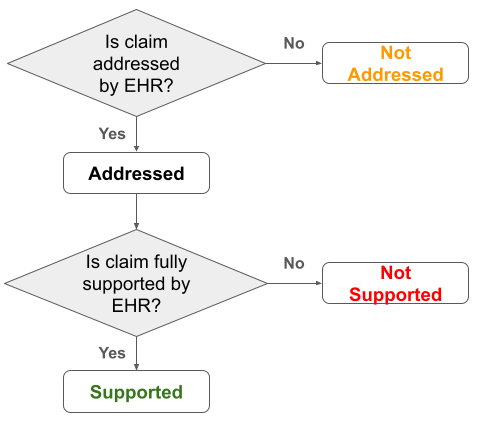}
    \caption{Guidelines provided to clinician annotators on how to categorize each proposition as \textit{Supported}, \textit{Not Supported}, or \textit{Not Addressed}. EHR = Electronic Health Record.}
    \label{fig:supfig7b}
\end{figure}

\clearpage
\subsubsection{Electronic Health Record PDF - Table of Contents}\label{supfigs_sub7sub3}
\begin{figure}[hbt]
    \centering
    \includegraphics[width=1\linewidth]{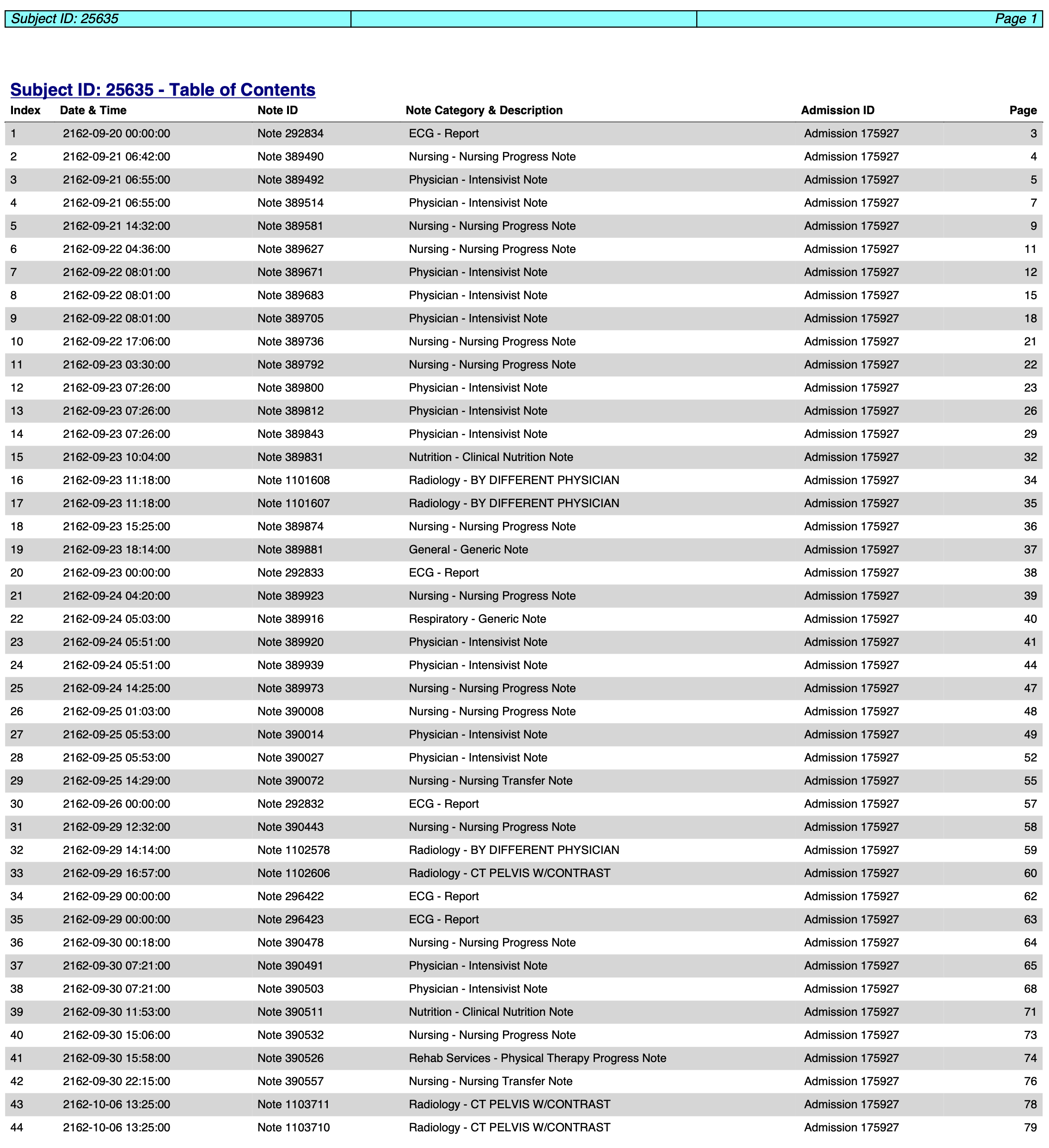}
    \caption{Table of contents for a patient’s Electronic Health Record PDF}
    \label{fig:supfig7c}
\end{figure}

\subsubsection{Electronic Health Record PDF - Clinical Note}\label{supfigs_sub7sub4}
\begin{figure}
    \centering
    \includegraphics[width=1\linewidth]{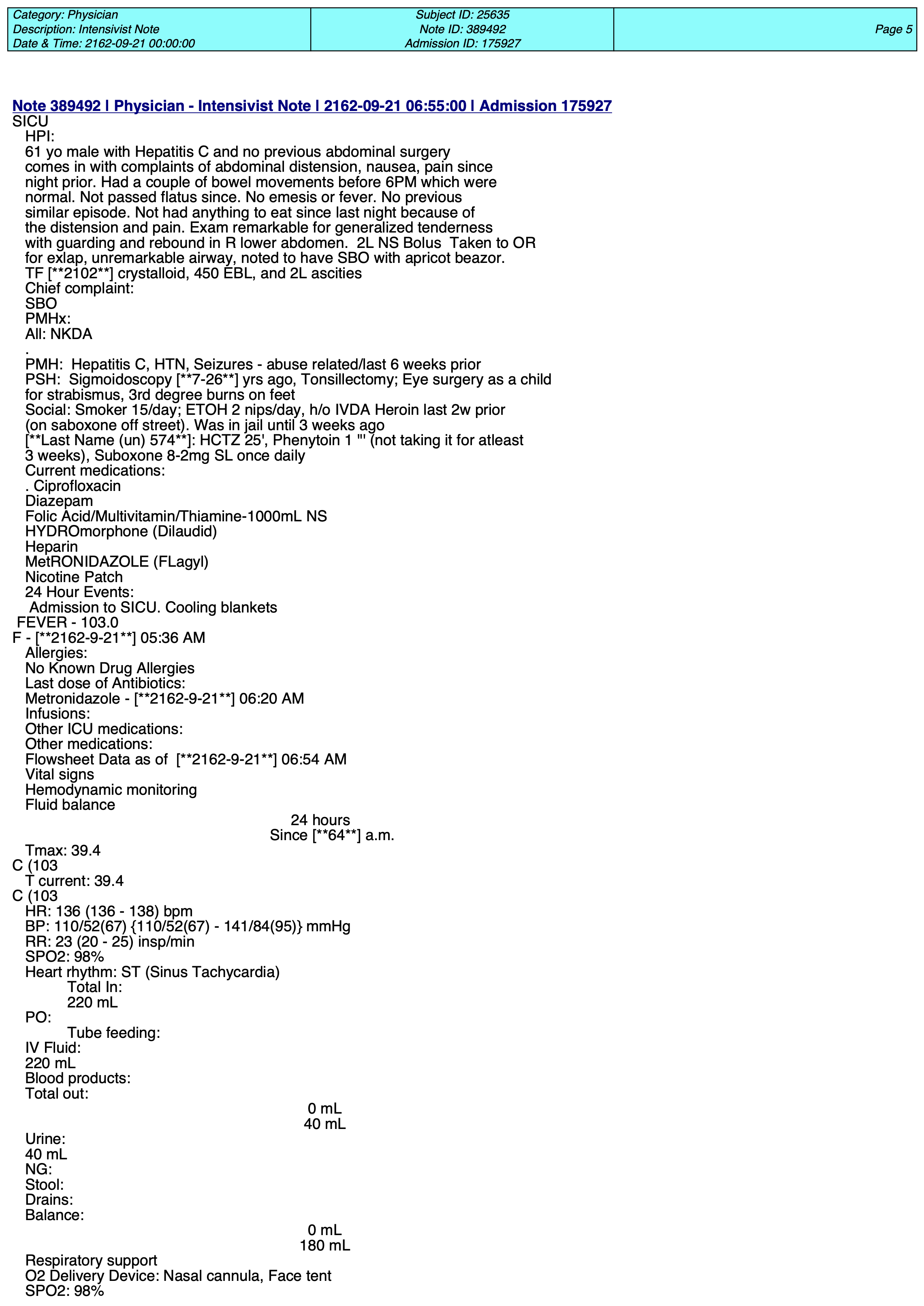}
    \caption{Clinical note in a patient’s Electronic Health Record PDF. Keyword search can be performed over the entire PDF using a PDF reader's text search features.
}
    \label{fig:supfig7d}
\end{figure}

\clearpage
\subsection{Sample Reference Contexts}\label{supfigs_sub8}
Representative examples of different types of reference contexts provided to the LLM-as-a-Judge for the same proposition. In this example, the verdict label and reason that \textit{VeriFact} assigns changes depending on the reference context. The reference contexts shown in this figure is \textit{VeriFact} with atomic claim propositions and dense retrieval of top N=10 facts across the entire EHR.

\subsubsection{Relevance Score Reference Context}\label{supfigs_sub8sub1}
The score assigned by the fact retriever is presented in this reference context and is also used to order the facts.
\\
\textbf{Proposition:}
\begin{lstlisting}[style=prompt]
Taking Ambien instead of methadone led to the patient's altered mental state.
\end{lstlisting}
\textbf{Reference Context:}
\begin{lstlisting}[style=prompt]
Electronic Health Record Context
Ordered by Relevance Score (Highest to Lowest):
1. Score: 0.83, Note Category: Nursing, Note Description: Nursing Transfer Note | Text: He took ambien instead of his methadone yesterday afternoon.
2. Score: 0.76, Note Category: Physician, Note Description: Physician Resident Admission Note | Text: The patient had altered mental status.
3. Score: 0.72, Note Category: Physician, Note Description: Physician Attending Progress Note | Text: The patient presents with altered mental status.
4. Score: 0.71, Note Category: Physician, Note Description: Physician Attending Progress Note | Text: The patient's altered mental status is likely related to infection.
5. Score: 0.71, Note Category: Physician, Note Description: Physician Resident Admission Note | Text: The altered mental status was from infection.
6. Score: 0.70, Note Category: Nursing, Note Description: Nursing Progress Note | Text: The patient has altered mental status, not delirium.
7. Score: 0.69, Note Category: Physician, Note Description: Physician Resident Admission Note | Text: The patient's complex pain medication regimen at home may have complicated the altered mental status.
\end{lstlisting}
\textbf{Verdict:}
\begin{lstlisting}[style=prompt]
Not Supported
\end{lstlisting}
\textbf{Reason:}
\begin{lstlisting}[style=prompt]
The text states that taking Ambien instead of methadone led to the patient's altered mental state, but the reference context suggests that the patient's altered mental status is likely related to infection.
\end{lstlisting}

\subsubsection{Absolute Time Reference Context}\label{supfigs_sub8sub2}
Retrieved facts are re-ordered in chronological order based on the timestamp of the clinical note from which the fact was extracted. Timestamps are also included in the reference context.
\\
\textbf{Proposition:}
\begin{lstlisting}[style=prompt]
Taking Ambien instead of methadone led to the patient's altered mental state.
\end{lstlisting}
\textbf{Reference Context:}
\begin{lstlisting}[style=prompt]
Hospital Admission Start: 2198-08-02 17:41:00
Hospital Admission End: 2198-08-04 12:12:00
Electronic Health Record Context
Ordered by Time (Earliest to Latest):
1. Date: 2198-08-02, Time: 19:30:00, Note Category: Physician, Note Description: Physician Resident Admission Note | Text: The patient had altered mental status.
2. Date: 2198-08-02, Time: 19:30:00, Note Category: Physician, Note Description: Physician Resident Admission Note | Text: The altered mental status was from infection.
3. Date: 2198-08-02, Time: 19:30:00, Note Category: Physician, Note Description: Physician Resident Admission Note | Text: The patient's complex pain medication regimen at home may have complicated the altered mental status.
4. Date: 2198-08-03, Time: 05:13:00, Note Category: Nursing, Note Description: Nursing Progress Note | Text: The patient has altered mental status, not delirium.
5. Date: 2198-08-03, Time: 09:11:00, Note Category: Physician, Note Description: Physician Attending Progress Note | Text: The patient's altered mental status is likely related to infection.
6. Date: 2198-08-03, Time: 09:11:00, Note Category: Physician, Note Description: Physician Attending Progress Note | Text: The patient presents with altered mental status.
7. Date: 2198-08-03, Time: 15:27:00, Note Category: Nursing, Note Description: Nursing Transfer Note | Text: He took ambien instead of his methadone yesterday afternoon.
\end{lstlisting}
\textbf{Verdict:}
\begin{lstlisting}[style=prompt]
Supported
\end{lstlisting}
\textbf{Reason:}
\begin{lstlisting}[style=prompt]
The text states that taking Ambien instead of methadone led to the patient's altered mental state, which is supported by the nursing transfer note that the patient took Ambien instead of his methadone yesterday afternoon.
\end{lstlisting}

\subsubsection{Relative Time Reference Context}\label{supfigs_sub8sub3}
Retrieved facts are re-ordered in chronological order based on the timestamp of the clinical note from which the fact was extracted. Time is expressed as relative hours and days from discharge time.
\\
\textbf{Proposition:}
\begin{lstlisting}[style=prompt]
Taking Ambien instead of methadone led to the patient's altered mental state.
\end{lstlisting}
\textbf{Reference Context:}
\begin{lstlisting}[style=prompt]
Hospital Admission Start: 1 days 18 hours ago
Hospital Admission End: Now
Electronic Health Record Context
Ordered by Time (Earliest to Latest):
1. When: 2 days 12 hours ago, Note Category: Physician, Note Description: Physician Resident Admission Note | Text: The patient had altered mental status.
2. When: 2 days 12 hours ago, Note Category: Physician, Note Description: Physician Resident Admission Note | Text: The altered mental status was from infection.
3. When: 2 days 12 hours ago, Note Category: Physician, Note Description: Physician Resident Admission Note | Text: The patient's complex pain medication regimen at home may have complicated the altered mental status.
4. When: 1 days 12 hours ago, Note Category: Nursing, Note Description: Nursing Progress Note | Text: The patient has altered mental status, not delirium.
5. When: 1 days 12 hours ago, Note Category: Physician, Note Description: Physician Attending Progress Note | Text: The patient's altered mental status is likely related to infection.
6. When: 1 days 12 hours ago, Note Category: Physician, Note Description: Physician Attending Progress Note | Text: The patient presents with altered mental status.
7. When: 1 days 12 hours ago, Note Category: Nursing, Note Description: Nursing Transfer Note | Text: He took ambien instead of his methadone yesterday afternoon.
\end{lstlisting}
\textbf{Verdict:}
\begin{lstlisting}[style=prompt]
Supported
\end{lstlisting}
\textbf{Reason:}
\begin{lstlisting}[style=prompt]
The text states that taking Ambien instead of methadone led to the patient's altered mental state, which is supported by the nursing transfer note that the patient took Ambien instead of his methadone yesterday afternoon.
\end{lstlisting}

%%======================%%
%% Supplementary Tables %%
%%======================%%

\clearpage
\begin{landscape}
\section{Supplementary Tables}\label{suptables}

\subsection{\textit{VeriFact-BHC} Dataset Statistics and Composition}\label{suptables_sub1}

\begin{table}[!h]
    \begin{tabular}{@{}clcll@{}}
    \toprule
    \multicolumn{3}{c}{\textbf{Table B1: \textit{VeriFact-BHC} Dataset Statistics and Composition}} &  &  \\ \midrule
    \multirow{8}{*}{\textbf{Patient Demographics}} & \textbf{Patients, no.} & 100 &  &  \\
     & \textbf{White, no.} & 77 &  &  \\
     & \textbf{Black/African American, no.} & 9 &  &  \\
     & \textbf{Asian, no.} & 4 &  &  \\
     & \textbf{Hispanic/Latino, no.} & 2 &  &  \\
     & \textbf{Unknown/Declined to Answer, no.} & 8 &  &  \\
     & \textbf{Age, years, mean ± std} & 65.05 ± 18.15 &  &  \\
     & \textbf{Admission Length of Stay, days, mean ± std} & 7.77 ± 7.50 &  &  \\ \midrule
    \multirow{3}{*}{\textbf{EHR Notes}} & \textbf{EHR Notes Count   (Current Admission), median (IQR)} & 22 (15, 36) &  &  \\
     & \textbf{EHR Notes Count (All), median (IQR)} & 25 (16, 47) &  &  \\
     & \textbf{Average Words per EHR Note, median (IQR)} & 375 (322, 444) &  &  \\ \midrule
    \multirow{6}{*}{\textbf{Brief Hospital Course}} & \textbf{Words in   Human-written summary, median (IQR)} & 334 (215, 502) &  &  \\
     & \textbf{Words in LLM-written summary, median (IQR)} & 164 (140, 248) &  &  \\
     & \textbf{Atomic Claims in Human-written summary, median (IQR)} & 49 (34, 76) &  &  \\
     & \textbf{Sentences in Human-written summary, median (IQR)} & 24 (17, 43) &  &  \\
     & \textbf{Atomic Claims in LLM-written summary, median (IQR)} & 28 (22, 43) &  &  \\
     & \textbf{Sentences in LLM-written summary, median (IQR)} & 8 (7, 12) &  &  \\ \midrule 
    \end{tabular}
    \caption{Patient demographics and note statistics. Human-written summary refers to the original Brief Hospital Course (BHC) section of the discharge summary note from the MIMIC-III dataset. The LLM-written summary is generated only from the patient’s current admission EHR notes that are available in the MIMIC-III dataset. However, the Human-written summary was not constrained to using the patient's EHR notes in MIMIC-III as a source of knowledge as the original clinicians taking care of the patient had access to additional clinical notes and patient discussions during routine clinical care. }
    \label{tab:suptable1}
\end{table}

\clearpage
\subsection{Percent Agreement for \textit{VeriFact} vs. Ground Truth with LLM-written Brief Hospital Course}\label{suptables_sub2}
% Atomic Claim Propositions | Original Label Space
\begin{table}[h]
\footnotesize
    \begin{tabular}{@{}ccllllll@{}}
    \toprule
    \multicolumn{8}{c}{\textbf{\shortstack{Table B2: Percent Agreement for LLM-written Brief Hospital Course \\ with Atomic Claim Propositions \& Original 3-Label Prediction Task}}} \\ \midrule
    \multicolumn{1}{l}{} & \textbf{\begin{tabular}[c]{@{}c@{}}Reference Context \\ Format\end{tabular}} & \multicolumn{2}{c}{\textbf{Relevance Score}} & \multicolumn{2}{c}{\textbf{Absolute Time}} & \multicolumn{2}{c}{\textbf{Relative Time}} \\ \cmidrule{2-8}
    \multicolumn{1}{l}{} & \textbf{\begin{tabular}[c]{@{}c@{}}Retrieve Facts \\ Only From \\ Current Admission\end{tabular}} & \multicolumn{1}{c}{\textbf{Yes}} & \multicolumn{1}{c}{\textbf{No}} & \multicolumn{1}{c}{\textbf{Yes}} & \multicolumn{1}{c}{\textbf{No}} & \multicolumn{1}{c}{\textbf{Yes}} & \multicolumn{1}{c}{\textbf{No}} \\ \midrule
    \textbf{\begin{tabular}[c]{@{}c@{}}Retrieval \\ Method\end{tabular}} & \textbf{Top   N} &  &  &  &  &  &  \\ \midrule
     & \textbf{5} & \cellcolor[HTML]{F7FBFF}60.6\% (58.9-62.2\%) & \cellcolor[HTML]{DAE8F6}65.4\% (63.8-66.9\%) & \cellcolor[HTML]{EBF3FB}62.5\% (60.8-64.1\%) & \cellcolor[HTML]{C7DCEF}68.4\% (66.8-70.0\%) & \cellcolor[HTML]{F2F7FD}61.5\% (59.9-63.2\%) & \cellcolor[HTML]{D5E5F4}66.2\% (64.6-67.8\%) \\
     & \textbf{10} & \cellcolor[HTML]{D3E3F3}66.6\% (65.0-68.3\%) & \cellcolor[HTML]{AED1E7}71.1\% (69.5-72.7\%) & \cellcolor[HTML]{B2D2E8}70.7\% (69.4-72.2\%) & \cellcolor[HTML]{7DB8DA}75.2\% (73.8-76.7\%) & \cellcolor[HTML]{C4DAEE}68.8\% (67.3-70.3\%) & \cellcolor[HTML]{94C4DF}73.5\% (72.1-74.9\%) \\
     & \textbf{25} & \cellcolor[HTML]{8DC1DD}74.0\% (72.5-75.5\%) & \cellcolor[HTML]{57A0CE}{\color[HTML]{F1F1F1} 78.6\% (77.2-80.0\%)} & \cellcolor[HTML]{4B98CA}{\color[HTML]{F1F1F1} 79.8\% (78.4-81.1\%)} & \cellcolor[HTML]{2676B8}{\color[HTML]{F1F1F1} 84.0\% (82.9-85.2\%)} & \cellcolor[HTML]{6CAED6}{\color[HTML]{F1F1F1} 76.6\% (75.1-77.9\%)} & \cellcolor[HTML]{3989C1}{\color[HTML]{F1F1F1} 81.8\% (80.5-83.1\%)} \\
    \multirow{-4}{*}{\textbf{Dense}} & \textbf{50} & \cellcolor[HTML]{5AA2CF}{\color[HTML]{F1F1F1} 78.3\% (76.9-79.7\%)} & \cellcolor[HTML]{2D7DBB}{\color[HTML]{F1F1F1} 83.2\% (82.0-84.5\%)} & \cellcolor[HTML]{2F7FBC}{\color[HTML]{F1F1F1} 83.0\% (81.7-84.2\%)} & \cellcolor[HTML]{0E59A2}{\color[HTML]{F1F1F1} 87.6\% (86.6-88.7\%)} & \cellcolor[HTML]{3C8CC3}{\color[HTML]{F1F1F1} 81.3\% (80.0-82.7\%)} & \cellcolor[HTML]{1663AA}{\color[HTML]{F1F1F1} 86.4\% (85.3-87.5\%)} \\ \midrule
     & \textbf{5} & \cellcolor[HTML]{E7F0FA}63.3\% (61.7-64.8\%) & \cellcolor[HTML]{CADEF0}67.9\% (66.4-69.5\%) & \cellcolor[HTML]{D9E8F5}65.4\% (63.8-66.9\%) & \cellcolor[HTML]{B2D2E8}70.7\% (69.1-72.2\%) & \cellcolor[HTML]{E5EFF9}63.6\% (62.0-65.2\%) & \cellcolor[HTML]{C4DAEE}68.8\% (67.3-70.4\%) \\
     & \textbf{10} & \cellcolor[HTML]{BDD7EC}69.6\% (68.1-71.2\%) & \cellcolor[HTML]{87BDDC}74.4\% (72.9-75.9\%) & \cellcolor[HTML]{9DCAE1}72.8\% (71.3-74.3\%) & \cellcolor[HTML]{65AAD4}{\color[HTML]{F1F1F1} 77.2\% (75.9-78.6\%)} & \cellcolor[HTML]{AACFE5}71.5\% (70.0-72.9\%) & \cellcolor[HTML]{6DAFD7}{\color[HTML]{F1F1F1} 76.5\% (75.1-77.9\%)} \\
     & \textbf{25} & \cellcolor[HTML]{7DB8DA}75.3\% (73.9-76.7\%) & \cellcolor[HTML]{4D99CA}{\color[HTML]{F1F1F1} 79.6\% (78.3-80.9\%)} & \cellcolor[HTML]{3F8FC5}{\color[HTML]{F1F1F1} 81.1\% (79.8-82.4\%)} & \cellcolor[HTML]{206FB4}{\color[HTML]{F1F1F1} 84.9\% (83.7-86.0\%)} & \cellcolor[HTML]{549FCD}{\color[HTML]{F1F1F1} 78.8\% (77.6-80.1\%)} & \cellcolor[HTML]{2777B8}{\color[HTML]{F1F1F1} 83.9\% (82.6-85.1\%)} \\
    \multirow{-4}{*}{\textbf{Hybrid}} & \textbf{50} & \cellcolor[HTML]{529DCC}{\color[HTML]{F1F1F1} 79.1\% (77.8-80.4\%)} & \cellcolor[HTML]{2E7EBC}{\color[HTML]{F1F1F1} 83.1\% (81.9-84.4\%)} & \cellcolor[HTML]{2777B8}{\color[HTML]{F1F1F1} 83.9\% (82.7-85.1\%)} & \cellcolor[HTML]{09529D}{\color[HTML]{FFFFFF} 88.6\% (87.5-89.6\%)} & \cellcolor[HTML]{3F8FC5}{\color[HTML]{F1F1F1} 81.0\% (79.6-82.2\%)} & \cellcolor[HTML]{1561A9}{\color[HTML]{F1F1F1} 86.7\% (85.6-87.9\%)} \\ \midrule
     & \textbf{5} & \cellcolor[HTML]{D6E6F4}65.9\% (64.4-67.5\%) & \cellcolor[HTML]{B3D3E8}70.5\% (69.1-72.1\%) & \cellcolor[HTML]{D9E7F5}65.6\% (64.0-67.1\%) & \cellcolor[HTML]{AFD1E7}70.9\% (69.4-72.6\%) & \cellcolor[HTML]{E3EEF9}63.8\% (62.1-65.4\%) & \cellcolor[HTML]{C4DAEE}68.8\% (67.3-70.3\%) \\
     & \textbf{10} & \cellcolor[HTML]{A9CFE5}71.6\% (70.2-73.1\%) & \cellcolor[HTML]{6FB0D7}{\color[HTML]{F1F1F1} 76.4\% (74.9-77.8\%)} & \cellcolor[HTML]{9CC9E1}72.8\% (71.4-74.3\%) & \cellcolor[HTML]{64A9D3}{\color[HTML]{F1F1F1} 77.3\% (76.0-78.7\%)} & \cellcolor[HTML]{ADD0E6}71.2\% (69.8-72.7\%) & \cellcolor[HTML]{6CAED6}{\color[HTML]{F1F1F1} 76.6\% (75.2-78.0\%)} \\
     & \textbf{25} & \cellcolor[HTML]{64A9D3}{\color[HTML]{F1F1F1} 77.3\% (75.9-78.8\%)} & \cellcolor[HTML]{3484BF}{\color[HTML]{F1F1F1} 82.4\% (81.1-83.7\%)} & \cellcolor[HTML]{3E8EC4}{\color[HTML]{F1F1F1} 81.1\% (79.8-82.4\%)} & \cellcolor[HTML]{1D6CB1}{\color[HTML]{F1F1F1} 85.3\% (84.1-86.4\%)} & \cellcolor[HTML]{549FCD}{\color[HTML]{F1F1F1} 78.8\% (77.6-80.1\%)} & \cellcolor[HTML]{2979B9}{\color[HTML]{F1F1F1} 83.8\% (82.6-85.0\%)} \\
    \multirow{-4}{*}{\textbf{Rerank}} & \textbf{50} & \cellcolor[HTML]{3B8BC2}{\color[HTML]{F1F1F1} 81.6\% (80.3-82.9\%)} & \cellcolor[HTML]{1967AD}{\color[HTML]{F1F1F1} 86.0\% (84.8-87.1\%)} & \cellcolor[HTML]{2A7AB9}{\color[HTML]{F1F1F1} 83.6\% (82.4-84.9\%)} & \cellcolor[HTML]{08509B}{\color[HTML]{FFFF00} \textbf{88.8\% (87.7-89.8\%)}} & \cellcolor[HTML]{3F8FC5}{\color[HTML]{F1F1F1} 81.1\% (79.7-82.3\%)} & \cellcolor[HTML]{135FA7}{\color[HTML]{F1F1F1} 86.9\% (85.7-88.0\%)} \\ \bottomrule
    \end{tabular}
    \caption{Percent Agreement for \textit{VeriFact} vs. Ground Truth computed across all propositions extracted from LLM-written Brief Hospital Course (BHC) from 100 patients. This table shows results when decompose the BHC as Atomic Claim Propositions and using \textit{VeriFact} to make a prediction in the Original 3-Label Prediction Task. The table displays the evaluation results for each combination of the following experimental variables: Top N (number of facts retrieved), Retrieval Method, Reference Context Format, and whether to retrieve facts only from current admission versus the entire EHR. Background cell color intensity is darker blue with higher agreement. The highest agreement is highlighted with yellow text. 95\% confidence intervals shown in parenthesis.}
    \label{tab:suptable2_sub1}
\end{table}

% Sentence Propositions | Original Label Space
\begin{table}[h]
    \footnotesize
    \begin{tabular}{@{}ccllllll@{}}
    \toprule
    \multicolumn{8}{c}{\textbf{\shortstack{Table B3: Percent Agreement for LLM-written Brief Hospital Course \\ with Sentence Propositions \& Original 3-Label Prediction Task}}} \\ \midrule
    \multicolumn{1}{l}{} & \textbf{\begin{tabular}[c]{@{}c@{}}Reference Context \\ Format\end{tabular}} & \multicolumn{2}{c}{\textbf{Relevance Score}} & \multicolumn{2}{c}{\textbf{Absolute Time}} & \multicolumn{2}{c}{\textbf{Relative Time}} \\ \cmidrule{2-8}
    \multicolumn{1}{l}{} & \textbf{\begin{tabular}[c]{@{}c@{}}Retrieve Facts \\ Only From \\ Current Admission\end{tabular}} & \multicolumn{1}{c}{\textbf{Yes}} & \multicolumn{1}{c}{\textbf{No}} & \multicolumn{1}{c}{\textbf{Yes}} & \multicolumn{1}{c}{\textbf{No}} & \multicolumn{1}{c}{\textbf{Yes}} & \multicolumn{1}{c}{\textbf{No}} \\ \midrule
    \textbf{\begin{tabular}[c]{@{}c@{}}Retrieval \\ Method\end{tabular}} & \textbf{Top   N} &  &  &  &  &  &  \\ \midrule
     & \textbf{5} & \cellcolor[HTML]{E3EEF9}63.7\%   (60.6-66.8\%) & \cellcolor[HTML]{DFEBF7}64.5\% (61.5-67.7\%) & \cellcolor[HTML]{ECF4FB}62.4\% (59.2-65.4\%) & \cellcolor[HTML]{E9F2FA}62.9\% (59.6-65.9\%) & \cellcolor[HTML]{F0F6FD}61.7\% (58.5-64.7\%) & \cellcolor[HTML]{E2EDF8}64.0\% (61.0-66.9\%) \\
     & \textbf{10} & \cellcolor[HTML]{89BEDC}74.3\% (71.5-77.2\%) & \cellcolor[HTML]{6AAED6}{\color[HTML]{F1F1F1} 76.8\% (74.2-79.4\%)} & \cellcolor[HTML]{79B5D9}75.6\% (72.8-78.4\%) & \cellcolor[HTML]{58A1CF}{\color[HTML]{F1F1F1} 78.5\% (76.1-81.0\%)} & \cellcolor[HTML]{9CC9E1}72.8\% (70.1-75.7\%) & \cellcolor[HTML]{5BA3D0}{\color[HTML]{F1F1F1} 78.3\% (75.7-80.8\%)} \\
     & \textbf{25} & \cellcolor[HTML]{2373B6}{\color[HTML]{F1F1F1} 84.5\% (82.5-86.8\%)} & \cellcolor[HTML]{1562A9}{\color[HTML]{F1F1F1} 86.6\% (84.6-88.7\%)} & \cellcolor[HTML]{1B69AF}{\color[HTML]{F1F1F1} 85.7\% (83.5-87.9\%)} & \cellcolor[HTML]{0A539E}{\color[HTML]{F1F1F1} 88.5\% (86.3-90.5\%)} & \cellcolor[HTML]{2777B8}{\color[HTML]{F1F1F1} 83.9\% (81.7-86.3\%)} & \cellcolor[HTML]{0A539E}{\color[HTML]{F1F1F1} 88.5\% (86.6-90.5\%)} \\
    \multirow{-4}{*}{\textbf{Dense}} & \textbf{50} & \cellcolor[HTML]{084A91}{\color[HTML]{F1F1F1} 89.6\% (87.8-91.5\%)} & \cellcolor[HTML]{083C7D}{\color[HTML]{F1F1F1} 91.2\% (89.5-92.8\%)} & \cellcolor[HTML]{08509B}{\color[HTML]{F1F1F1} 88.8\% (86.9-90.8\%)} & \cellcolor[HTML]{083979}{\color[HTML]{F1F1F1} 91.6\% (89.8-93.3\%)} & \cellcolor[HTML]{0A539E}{\color[HTML]{F1F1F1} 88.5\% (86.6-90.5\%)} & \cellcolor[HTML]{083B7C}{\color[HTML]{F1F1F1} 91.3\% (89.6-93.0\%)} \\ \midrule
     & \textbf{5} & \cellcolor[HTML]{B7D4EA}70.2\% (67.3-73.1\%) & \cellcolor[HTML]{AACFE5}71.4\% (68.8-74.3\%) & \cellcolor[HTML]{C6DBEF}68.7\% (65.8-71.5\%) & \cellcolor[HTML]{AFD1E7}70.9\% (68.0-73.7\%) & \cellcolor[HTML]{C9DDF0}68.2\% (65.2-71.2\%) & \cellcolor[HTML]{A9CFE5}71.5\% (68.6-74.3\%) \\
     & \textbf{10} & \cellcolor[HTML]{549FCD}{\color[HTML]{F1F1F1} 78.9\% (76.5-81.5\%)} & \cellcolor[HTML]{3888C1}{\color[HTML]{F1F1F1} 81.9\% (79.7-84.2\%)} & \cellcolor[HTML]{519CCC}{\color[HTML]{F1F1F1} 79.3\% (76.9-81.8\%)} & \cellcolor[HTML]{2E7EBC}{\color[HTML]{F1F1F1} 83.2\% (81.0-85.6\%)} & \cellcolor[HTML]{56A0CE}{\color[HTML]{F1F1F1} 78.8\% (76.4-81.3\%)} & \cellcolor[HTML]{2F7FBC}{\color[HTML]{F1F1F1} 83.1\% (80.8-85.5\%)} \\
     & \textbf{25} & \cellcolor[HTML]{115CA5}{\color[HTML]{F1F1F1} 87.3\% (85.3-89.3\%)} & \cellcolor[HTML]{08488E}{\color[HTML]{F1F1F1} 89.8\% (88.0-91.6\%)} & \cellcolor[HTML]{125DA6}{\color[HTML]{F1F1F1} 87.2\% (85.3-89.2\%)} & \cellcolor[HTML]{083B7C}{\color[HTML]{F1F1F1} 91.4\% (89.6-93.1\%)} & \cellcolor[HTML]{1562A9}{\color[HTML]{F1F1F1} 86.6\% (84.6-88.8\%)} & \cellcolor[HTML]{084488}{\color[HTML]{F1F1F1} 90.3\% (88.4-92.1\%)} \\
    \multirow{-4}{*}{\textbf{Hybrid}} & \textbf{50} & \cellcolor[HTML]{08468B}{\color[HTML]{F1F1F1} 90.1\% (88.4-91.9\%)} & \cellcolor[HTML]{083776}{\color[HTML]{F1F1F1} 91.9\% (90.2-93.6\%)} & \cellcolor[HTML]{08458A}{\color[HTML]{F1F1F1} 90.2\% (88.4-92.1\%)} & \cellcolor[HTML]{083370}{\color[HTML]{F1F1F1} 92.4\% (90.7-93.9\%)} & \cellcolor[HTML]{0B559F}{\color[HTML]{F1F1F1} 88.2\% (86.3-90.3\%)} & \cellcolor[HTML]{083877}{\color[HTML]{F1F1F1} 91.7\% (89.9-93.4\%)} \\ \midrule
     & \textbf{5} & \cellcolor[HTML]{AED1E7}71.1\% (68.2-73.8\%) & \cellcolor[HTML]{9CC9E1}72.8\% (70.2-75.5\%) & \cellcolor[HTML]{C2D9EE}69.0\% (66.2-71.8\%) & \cellcolor[HTML]{B4D3E9}70.4\% (67.6-73.1\%) & \cellcolor[HTML]{C9DDF0}68.1\% (65.0-71.1\%) & \cellcolor[HTML]{A6CEE4}71.8\% (68.9-74.5\%) \\
     & \textbf{10} & \cellcolor[HTML]{4695C8}{\color[HTML]{F1F1F1} 80.3\% (78.0-82.8\%)} & \cellcolor[HTML]{3282BE}{\color[HTML]{F1F1F1} 82.7\% (80.4-85.1\%)} & \cellcolor[HTML]{4E9ACB}{\color[HTML]{F1F1F1} 79.5\% (77.1-82.0\%)} & \cellcolor[HTML]{2C7CBA}{\color[HTML]{F1F1F1} 83.5\% (81.3-85.8\%)} & \cellcolor[HTML]{5CA4D0}{\color[HTML]{F1F1F1} 78.2\% (75.7-80.8\%)} & \cellcolor[HTML]{2E7EBC}{\color[HTML]{F1F1F1} 83.2\% (80.9-85.6\%)} \\
     & \textbf{25} & \cellcolor[HTML]{08509B}{\color[HTML]{F1F1F1} 88.9\% (87.1-90.9\%)} & \cellcolor[HTML]{083C7D}{\color[HTML]{F1F1F1} 91.2\% (89.4-93.0\%)} & \cellcolor[HTML]{115CA5}{\color[HTML]{F1F1F1} 87.4\% (85.5-89.4\%)} & \cellcolor[HTML]{083979}{\color[HTML]{F1F1F1} 91.6\% (89.9-93.3\%)} & \cellcolor[HTML]{1562A9}{\color[HTML]{F1F1F1} 86.6\% (84.5-88.8\%)} & \cellcolor[HTML]{084082}{\color[HTML]{F1F1F1} 90.8\% (89.0-92.6\%)} \\
    \multirow{-4}{*}{\textbf{Rerank}} & \textbf{50} & \cellcolor[HTML]{084082}{\color[HTML]{F1F1F1} 90.9\% (89.1-92.7\%)} & \cellcolor[HTML]{08306B}{\color[HTML]{FFFF00} \textbf{92.7\% (91.0-94.2\%)}} & \cellcolor[HTML]{08478D}{\color[HTML]{F1F1F1} 90.0\% (88.2-91.9\%)} & \cellcolor[HTML]{083573}{\color[HTML]{F1F1F1} 92.1\% (90.4-93.7\%)} & \cellcolor[HTML]{0E58A2}{\color[HTML]{F1F1F1} 87.8\% (85.8-89.9\%)} & \cellcolor[HTML]{083471}{\color[HTML]{F1F1F1} 92.2\% (90.5-93.9\%)} \\ \bottomrule
    \end{tabular}
    \caption{Percent Agreement for \textit{VeriFact} vs. Ground Truth computed across all propositions extracted from LLM-written Brief Hospital Course (BHC) from 100 patients. This table shows results when decompose the BHC as Sentence Propositions and using \textit{VeriFact} to make a prediction in the Original 3-Label Prediction Task. The table displays the evaluation results for each combination of the following experimental variables: Top N (number of facts retrieved), Retrieval Method, Reference Context Format, and whether to retrieve facts only from current admission versus the entire EHR. Background cell color intensity is darker blue with higher agreement. The highest agreement is highlighted with yellow text. 95\% confidence intervals shown in parenthesis.}
    \label{tab:suptable2_sub2}
\end{table}

% Atomic Claim Propositions | Binarized Label Space
\begin{table}[h]
    \footnotesize
    \begin{tabular}{@{}ccllllll@{}}
    \toprule
    \multicolumn{8}{c}{\textbf{\shortstack{Table B4: Percent Agreement for LLM-written Brief Hospital Course \\ with Atomic Claim Propositions \& Binarized 2-Label Prediction Task}}} \\ \midrule
    \multicolumn{1}{l}{} & \textbf{\begin{tabular}[c]{@{}c@{}}Reference Context \\ Format\end{tabular}} & \multicolumn{2}{c}{\textbf{Relevance Score}} & \multicolumn{2}{c}{\textbf{Absolute Time}} & \multicolumn{2}{c}{\textbf{Relative Time}} \\ \cmidrule{2-8}
    \multicolumn{1}{l}{} & \textbf{\begin{tabular}[c]{@{}c@{}}Retrieve Facts \\ Only From \\ Current Admission\end{tabular}} & \multicolumn{1}{c}{\textbf{Yes}} & \multicolumn{1}{c}{\textbf{No}} & \multicolumn{1}{c}{\textbf{Yes}} & \multicolumn{1}{c}{\textbf{No}} & \multicolumn{1}{c}{\textbf{Yes}} & \multicolumn{1}{c}{\textbf{No}} \\ \midrule
    \textbf{\begin{tabular}[c]{@{}c@{}}Retrieval \\ Method\end{tabular}} & \textbf{Top   N} &  &  &  &  &  &  \\ \midrule
     & \textbf{5} & \cellcolor[HTML]{EEF5FC}62.0\%   (60.4-63.6\%) & \cellcolor[HTML]{D2E3F3}66.6\% (65.1-68.3\%) & \cellcolor[HTML]{E3EEF8}63.9\% (62.3-65.5\%) & \cellcolor[HTML]{BDD7EC}69.5\% (68.0-71.1\%) & \cellcolor[HTML]{E9F2FA}62.9\% (61.1-64.6\%) & \cellcolor[HTML]{CDE0F1}67.4\% (65.9-69.0\%) \\
     & \textbf{10} & \cellcolor[HTML]{CCDFF1}67.7\% (66.1-69.3\%) & \cellcolor[HTML]{A3CCE3}72.2\% (70.7-73.8\%) & \cellcolor[HTML]{A9CFE5}71.6\% (70.3-73.1\%) & \cellcolor[HTML]{72B2D8}{\color[HTML]{F1F1F1} 76.1\% (74.7-77.5\%)} & \cellcolor[HTML]{BAD6EB}69.8\% (68.4-71.4\%) & \cellcolor[HTML]{87BDDC}74.4\% (73.1-75.8\%) \\
     & \textbf{25} & \cellcolor[HTML]{82BBDB}74.9\% (73.5-76.4\%) & \cellcolor[HTML]{4F9BCB}{\color[HTML]{F1F1F1} 79.3\% (78.0-80.7\%)} & \cellcolor[HTML]{4493C7}{\color[HTML]{F1F1F1} 80.5\% (79.2-81.9\%)} & \cellcolor[HTML]{2171B5}{\color[HTML]{F1F1F1} 84.7\% (83.5-85.9\%)} & \cellcolor[HTML]{63A8D3}{\color[HTML]{F1F1F1} 77.5\% (76.2-78.8\%)} & \cellcolor[HTML]{3383BE}{\color[HTML]{F1F1F1} 82.5\% (81.2-83.7\%)} \\
    \multirow{-4}{*}{\textbf{Dense}} & \textbf{50} & \cellcolor[HTML]{519CCC}{\color[HTML]{F1F1F1} 79.2\% (77.9-80.5\%)} & \cellcolor[HTML]{2777B8}{\color[HTML]{F1F1F1} 83.9\% (82.6-85.2\%)} & \cellcolor[HTML]{2A7AB9}{\color[HTML]{F1F1F1} 83.7\% (82.4-84.9\%)} & \cellcolor[HTML]{0B559F}{\color[HTML]{F1F1F1} 88.2\% (87.1-89.2\%)} & \cellcolor[HTML]{3787C0}{\color[HTML]{F1F1F1} 82.0\% (80.7-83.3\%)} & \cellcolor[HTML]{125EA6}{\color[HTML]{F1F1F1} 87.1\% (85.9-88.2\%)} \\ \midrule
     & \textbf{5} & \cellcolor[HTML]{DEEBF7}64.7\% (63.1-66.3\%) & \cellcolor[HTML]{C1D9ED}69.2\% (67.8-70.8\%) & \cellcolor[HTML]{D2E3F3}66.7\% (65.2-68.3\%) & \cellcolor[HTML]{A8CEE4}71.7\% (70.2-73.3\%) & \cellcolor[HTML]{DDEAF7}64.8\% (63.2-66.4\%) & \cellcolor[HTML]{B9D6EA}69.9\% (68.4-71.5\%) \\
     & \textbf{10} & \cellcolor[HTML]{B0D2E7}70.8\% (69.4-72.4\%) & \cellcolor[HTML]{7AB6D9}75.5\% (74.2-77.0\%) & \cellcolor[HTML]{91C3DE}73.8\% (72.4-75.2\%) & \cellcolor[HTML]{5CA4D0}{\color[HTML]{F1F1F1} 78.2\% (76.9-79.5\%)} & \cellcolor[HTML]{A0CBE2}72.4\% (71.1-73.9\%) & \cellcolor[HTML]{64A9D3}{\color[HTML]{F1F1F1} 77.3\% (76.0-78.7\%)} \\
     & \textbf{25} & \cellcolor[HTML]{71B1D7}{\color[HTML]{F1F1F1} 76.3\% (75.0-77.7\%)} & \cellcolor[HTML]{4594C7}{\color[HTML]{F1F1F1} 80.4\% (79.1-81.8\%)} & \cellcolor[HTML]{3989C1}{\color[HTML]{F1F1F1} 81.8\% (80.6-83.1\%)} & \cellcolor[HTML]{1B69AF}{\color[HTML]{F1F1F1} 85.7\% (84.4-86.8\%)} & \cellcolor[HTML]{4D99CA}{\color[HTML]{F1F1F1} 79.6\% (78.3-80.9\%)} & \cellcolor[HTML]{2272B6}{\color[HTML]{F1F1F1} 84.5\% (83.3-85.7\%)} \\
    \multirow{-4}{*}{\textbf{Hybrid}} & \textbf{50} & \cellcolor[HTML]{4997C9}{\color[HTML]{F1F1F1} 80.0\% (78.7-81.2\%)} & \cellcolor[HTML]{2979B9}{\color[HTML]{F1F1F1} 83.8\% (82.7-85.0\%)} & \cellcolor[HTML]{2373B6}{\color[HTML]{F1F1F1} 84.4\% (83.2-85.6\%)} & \cellcolor[HTML]{084E98}{\color[HTML]{F1F1F1} 89.0\% (88.0-90.0\%)} & \cellcolor[HTML]{3989C1}{\color[HTML]{F1F1F1} 81.8\% (80.4-83.0\%)} & \cellcolor[HTML]{115CA5}{\color[HTML]{F1F1F1} 87.3\% (86.2-88.4\%)} \\ \midrule
     & \textbf{5} & \cellcolor[HTML]{CEE0F2}67.3\% (65.8-68.9\%) & \cellcolor[HTML]{A6CEE4}71.8\% (70.4-73.3\%) & \cellcolor[HTML]{D1E2F3}66.8\% (65.3-68.3\%) & \cellcolor[HTML]{A4CCE3}72.1\% (70.5-73.6\%) & \cellcolor[HTML]{DCE9F6}65.1\% (63.5-66.7\%) & \cellcolor[HTML]{B9D6EA}70.0\% (68.5-71.4\%) \\
     & \textbf{10} & \cellcolor[HTML]{9CC9E1}72.8\% (71.5-74.3\%) & \cellcolor[HTML]{63A8D3}{\color[HTML]{F1F1F1} 77.5\% (76.1-78.9\%)} & \cellcolor[HTML]{91C3DE}73.8\% (72.4-75.3\%) & \cellcolor[HTML]{5BA3D0}{\color[HTML]{F1F1F1} 78.2\% (76.9-79.5\%)} & \cellcolor[HTML]{A3CCE3}72.2\% (70.8-73.6\%) & \cellcolor[HTML]{63A8D3}{\color[HTML]{F1F1F1} 77.5\% (76.1-78.9\%)} \\
     & \textbf{25} & \cellcolor[HTML]{5BA3D0}{\color[HTML]{F1F1F1} 78.2\% (76.9-79.7\%)} & \cellcolor[HTML]{2E7EBC}{\color[HTML]{F1F1F1} 83.2\% (82.0-84.5\%)} & \cellcolor[HTML]{3888C1}{\color[HTML]{F1F1F1} 81.9\% (80.7-83.2\%)} & \cellcolor[HTML]{1966AD}{\color[HTML]{F1F1F1} 86.0\% (84.8-87.1\%)} & \cellcolor[HTML]{4D99CA}{\color[HTML]{F1F1F1} 79.6\% (78.4-80.9\%)} & \cellcolor[HTML]{2272B6}{\color[HTML]{F1F1F1} 84.5\% (83.3-85.6\%)} \\
    \multirow{-4}{*}{\textbf{Rerank}} & \textbf{50} & \cellcolor[HTML]{3484BF}{\color[HTML]{F1F1F1} 82.4\% (81.1-83.6\%)} & \cellcolor[HTML]{1562A9}{\color[HTML]{F1F1F1} 86.6\% (85.4-87.7\%)} & \cellcolor[HTML]{2575B7}{\color[HTML]{F1F1F1} 84.1\% (82.9-85.4\%)} & \cellcolor[HTML]{084D96}{\color[HTML]{FFFF00} \textbf{89.3\% (88.2-90.2\%)}} & \cellcolor[HTML]{3989C1}{\color[HTML]{F1F1F1} 81.8\% (80.5-83.1\%)} & \cellcolor[HTML]{105BA4}{\color[HTML]{F1F1F1} 87.4\% (86.2-88.5\%)} \\ \bottomrule
    \end{tabular}
    \caption{Percent Agreement for \textit{VeriFact} vs. Ground Truth computed across all propositions extracted from LLM-written Brief Hospital Course (BHC) from 100 patients. This table shows results when decompose the BHC as Atomic Claim Propositions and using \textit{VeriFact} to make a prediction in the Binarized 2-Label Prediction Task. The table displays the evaluation results for each combination of the following experimental variables: Top N (number of facts retrieved), Retrieval Method, Reference Context Format, and whether to retrieve facts only from current admission versus the entire EHR. Background cell color intensity is darker blue with higher agreement. The highest agreement is highlighted with yellow text. 95\% confidence intervals shown in parenthesis.}
    \label{tab:suptable2_sub3}
\end{table}

% Sentence Claim Propositions | Binarized Label Space
\begin{table}[h]
    \footnotesize
    \begin{tabular}{@{}ccllllll@{}}
    \toprule
    \multicolumn{8}{c}{\textbf{\shortstack{Table B5: Percent Agreement for LLM-written Brief Hospital Course \\ with Sentence Propositions \& Binarized 2-Label Prediction Task}}} \\ \midrule
    \multicolumn{1}{l}{} & \textbf{\begin{tabular}[c]{@{}c@{}}Reference Context \\ Format\end{tabular}} & \multicolumn{2}{c}{\textbf{Relevance Score}} & \multicolumn{2}{c}{\textbf{Absolute Time}} & \multicolumn{2}{c}{\textbf{Relative Time}} \\ \cmidrule{2-8}
    \multicolumn{1}{l}{} & \textbf{\begin{tabular}[c]{@{}c@{}}Retrieve Facts \\ Only From \\ Current Admission\end{tabular}} & \multicolumn{1}{c}{\textbf{Yes}} & \multicolumn{1}{c}{\textbf{No}} & \multicolumn{1}{c}{\textbf{Yes}} & \multicolumn{1}{c}{\textbf{No}} & \multicolumn{1}{c}{\textbf{Yes}} & \multicolumn{1}{c}{\textbf{No}} \\ \midrule
    \textbf{\begin{tabular}[c]{@{}c@{}}Retrieval \\ Method\end{tabular}} & \textbf{Top   N} &  &  &  &  &  &  \\ \midrule
     & \textbf{5} & \cellcolor[HTML]{DCEAF6}64.9\%  (62.0-67.9\%) & \cellcolor[HTML]{D7E6F5}65.8\% (62.8-68.9\%) & \cellcolor[HTML]{E5EFF9}63.5\% (60.4-66.5\%) & \cellcolor[HTML]{E3EEF8}63.9\% (60.5-66.9\%) & \cellcolor[HTML]{EAF2FB}62.8\% (59.7-65.8\%) & \cellcolor[HTML]{DCE9F6}65.0\% (62.0-67.8\%) \\
     & \textbf{10} & \cellcolor[HTML]{7DB8DA}75.3\% (72.4-78.1\%) & \cellcolor[HTML]{61A7D2}{\color[HTML]{F1F1F1} 77.7\% (75.1-80.2\%)} & \cellcolor[HTML]{71B1D7}{\color[HTML]{F1F1F1} 76.3\% (73.6-79.1\%)} & \cellcolor[HTML]{529DCC}{\color[HTML]{F1F1F1} 79.2\% (76.8-81.6\%)} & \cellcolor[HTML]{92C4DE}73.6\% (70.8-76.5\%) & \cellcolor[HTML]{539ECD}{\color[HTML]{F1F1F1} 79.1\% (76.6-81.5\%)} \\
     & \textbf{25} & \cellcolor[HTML]{2070B4}{\color[HTML]{F1F1F1} 84.8\% (82.8-87.1\%)} & \cellcolor[HTML]{1460A8}{\color[HTML]{F1F1F1} 86.8\% (84.8-88.9\%)} & \cellcolor[HTML]{1967AD}{\color[HTML]{F1F1F1} 85.9\% (83.7-88.0\%)} & \cellcolor[HTML]{08519C}{\color[HTML]{F1F1F1} 88.7\% (86.5-90.7\%)} & \cellcolor[HTML]{2676B8}{\color[HTML]{F1F1F1} 84.1\% (82.1-86.4\%)} & \cellcolor[HTML]{08519C}{\color[HTML]{F1F1F1} 88.7\% (86.8-90.7\%)} \\
    \multirow{-4}{*}{\textbf{Dense}} & \textbf{50} & \cellcolor[HTML]{084990}{\color[HTML]{F1F1F1} 89.7\% (87.9-91.6\%)} & \cellcolor[HTML]{083B7C}{\color[HTML]{F1F1F1} 91.3\% (89.5-92.9\%)} & \cellcolor[HTML]{084F99}{\color[HTML]{F1F1F1} 89.0\% (87.1-90.9\%)} & \cellcolor[HTML]{083776}{\color[HTML]{F1F1F1} 91.8\% (90.0-93.5\%)} & \cellcolor[HTML]{09529D}{\color[HTML]{F1F1F1} 88.6\% (86.6-90.6\%)} & \cellcolor[HTML]{083B7C}{\color[HTML]{F1F1F1} 91.4\% (89.7-93.1\%)} \\ \midrule
     & \textbf{5} & \cellcolor[HTML]{B0D2E7}70.8\% (68.0-73.7\%) & \cellcolor[HTML]{A4CCE3}72.1\% (69.5-74.8\%) & \cellcolor[HTML]{C1D9ED}69.1\% (66.2-72.0\%) & \cellcolor[HTML]{AACFE5}71.4\% (68.4-74.1\%) & \cellcolor[HTML]{C1D9ED}69.1\% (66.2-72.1\%) & \cellcolor[HTML]{A3CCE3}72.2\% (69.3-75.1\%) \\
     & \textbf{10} & \cellcolor[HTML]{529DCC}{\color[HTML]{F1F1F1} 79.2\% (76.8-81.8\%)} & \cellcolor[HTML]{3585BF}{\color[HTML]{F1F1F1} 82.3\% (80.2-84.6\%)} & \cellcolor[HTML]{4E9ACB}{\color[HTML]{F1F1F1} 79.6\% (77.2-82.1\%)} & \cellcolor[HTML]{2C7CBA}{\color[HTML]{F1F1F1} 83.4\% (81.2-85.7\%)} & \cellcolor[HTML]{519CCC}{\color[HTML]{F1F1F1} 79.3\% (76.9-81.8\%)} & \cellcolor[HTML]{2C7CBA}{\color[HTML]{F1F1F1} 83.5\% (81.2-85.9\%)} \\
     & \textbf{25} & \cellcolor[HTML]{0F5AA3}{\color[HTML]{F1F1F1} 87.6\% (85.7-89.5\%)} & \cellcolor[HTML]{08488E}{\color[HTML]{F1F1F1} 89.9\% (88.1-91.7\%)} & \cellcolor[HTML]{115CA5}{\color[HTML]{F1F1F1} 87.3\% (85.3-89.3\%)} & \cellcolor[HTML]{083A7A}{\color[HTML]{F1F1F1} 91.5\% (89.8-93.2\%)} & \cellcolor[HTML]{1561A9}{\color[HTML]{F1F1F1} 86.7\% (84.7-88.8\%)} & \cellcolor[HTML]{084285}{\color[HTML]{F1F1F1} 90.6\% (88.7-92.4\%)} \\
    \multirow{-4}{*}{\textbf{Hybrid}} & \textbf{50} & \cellcolor[HTML]{08468B}{\color[HTML]{F1F1F1} 90.1\% (88.4-91.9\%)} & \cellcolor[HTML]{083776}{\color[HTML]{F1F1F1} 91.9\% (90.2-93.6\%)} & \cellcolor[HTML]{08458A}{\color[HTML]{F1F1F1} 90.2\% (88.4-92.1\%)} & \cellcolor[HTML]{083370}{\color[HTML]{F1F1F1} 92.4\% (90.7-93.9\%)} & \cellcolor[HTML]{0A549E}{\color[HTML]{F1F1F1} 88.3\% (86.4-90.4\%)} & \cellcolor[HTML]{083776}{\color[HTML]{F1F1F1} 91.9\% (90.1-93.6\%)} \\ \midrule
     & \textbf{5} & \cellcolor[HTML]{A8CEE4}71.7\% (68.9-74.4\%) & \cellcolor[HTML]{92C4DE}73.6\% (71.0-76.3\%) & \cellcolor[HTML]{BED8EC}69.4\% (66.6-72.2\%) & \cellcolor[HTML]{AFD1E7}70.9\% (68.0-73.6\%) & \cellcolor[HTML]{C3DAEE}68.9\% (66.0-72.0\%) & \cellcolor[HTML]{9FCAE1}72.5\% (69.6-75.3\%) \\
     & \textbf{10} & \cellcolor[HTML]{4292C6}{\color[HTML]{F1F1F1} 80.7\% (78.4-83.2\%)} & \cellcolor[HTML]{2F7FBC}{\color[HTML]{F1F1F1} 83.1\% (80.8-85.5\%)} & \cellcolor[HTML]{4B98CA}{\color[HTML]{F1F1F1} 79.8\% (77.4-82.3\%)} & \cellcolor[HTML]{2A7AB9}{\color[HTML]{F1F1F1} 83.7\% (81.5-85.9\%)} & \cellcolor[HTML]{56A0CE}{\color[HTML]{F1F1F1} 78.8\% (76.3-81.3\%)} & \cellcolor[HTML]{2C7CBA}{\color[HTML]{F1F1F1} 83.5\% (81.2-85.9\%)} \\
     & \textbf{25} & \cellcolor[HTML]{084D96}{\color[HTML]{F1F1F1} 89.2\% (87.4-91.1\%)} & \cellcolor[HTML]{083A7A}{\color[HTML]{F1F1F1} 91.5\% (89.7-93.2\%)} & \cellcolor[HTML]{105BA4}{\color[HTML]{F1F1F1} 87.5\% (85.6-89.5\%)} & \cellcolor[HTML]{083877}{\color[HTML]{F1F1F1} 91.7\% (90.0-93.4\%)} & \cellcolor[HTML]{1460A8}{\color[HTML]{F1F1F1} 86.8\% (84.8-88.9\%)} & \cellcolor[HTML]{083D7F}{\color[HTML]{F1F1F1} 91.1\% (89.3-92.9\%)} \\
    \multirow{-4}{*}{\textbf{Rerank}} & \textbf{50} & \cellcolor[HTML]{083E81}{\color[HTML]{F1F1F1} 91.0\% (89.2-92.7\%)} & \cellcolor[HTML]{08306B}{\color[HTML]{FFFF00} \textbf{92.8\% (91.1-94.4\%)}} & \cellcolor[HTML]{08478D}{\color[HTML]{F1F1F1} 90.0\% (88.2-91.9\%)} & \cellcolor[HTML]{083573}{\color[HTML]{F1F1F1} 92.1\% (90.4-93.7\%)} & \cellcolor[HTML]{0D57A1}{\color[HTML]{F1F1F1} 88.0\% (86.0-90.1\%)} & \cellcolor[HTML]{083370}{\color[HTML]{F1F1F1} 92.4\% (90.7-94.1\%)} \\ \bottomrule
    \end{tabular}
    \caption{Percent Agreement for \textit{VeriFact} vs. Ground Truth computed across all propositions extracted from LLM-written Brief Hospital Course (BHC) from 100 patients. This table shows results when decompose the BHC as Sentence Propositions and using \textit{VeriFact} to make a prediction in the Binarized 2-Label Prediction Task. The table displays the evaluation results for each combination of the following experimental variables: Top N (number of facts retrieved), Retrieval Method, Reference Context Format, and whether to retrieve facts only from current admission versus the entire EHR. Background cell color intensity is darker blue with higher agreement. The highest agreement is highlighted with yellow text. 95\% confidence intervals shown in parenthesis.}
    \label{tab:suptable2_sub4}
\end{table}

\clearpage
\subsection{Percent Agreement for \textit{VeriFact} vs. Ground Truth with Human-written Brief Hospital Course}\label{suptables_sub3}
% Atomic Claim Propositions | Original Label Space
\begin{table}[h]
    \footnotesize
    \begin{tabular}{@{}ccllllll@{}}
    \toprule
    \multicolumn{8}{c}{\textbf{\shortstack{Table B6: Percent Agreement for Human-written Brief Hospital Course \\ with Atomic Claim Propositions \& Original 3-Label Prediction Task}}} \\ \midrule
    \multicolumn{1}{l}{} & \textbf{\begin{tabular}[c]{@{}c@{}}Reference Context \\ Format\end{tabular}} & \multicolumn{2}{c}{\textbf{Relevance Score}} & \multicolumn{2}{c}{\textbf{Absolute Time}} & \multicolumn{2}{c}{\textbf{Relative Time}} \\ \cmidrule{2-8}
    \multicolumn{1}{l}{} & \textbf{\begin{tabular}[c]{@{}c@{}}Retrieve Facts \\ Only From \\ Current Admission\end{tabular}} & \multicolumn{1}{c}{\textbf{Yes}} & \multicolumn{1}{c}{\textbf{No}} & \multicolumn{1}{c}{\textbf{Yes}} & \multicolumn{1}{c}{\textbf{No}} & \multicolumn{1}{c}{\textbf{Yes}} & \multicolumn{1}{c}{\textbf{No}} \\ \midrule
    \textbf{\begin{tabular}[c]{@{}c@{}}Retrieval \\ Method\end{tabular}} & \textbf{Top   N} &  &  &  &  &  &  \\ \midrule
     & \textbf{5} & \cellcolor[HTML]{EFF6FC}52.4\%   (51.2-53.7\%) & \cellcolor[HTML]{E5EFF9}53.8\% (52.5-55.2\%) & \cellcolor[HTML]{F5F9FE}51.5\% (50.2-52.8\%) & \cellcolor[HTML]{E9F2FA}53.2\% (52.0-54.5\%) & \cellcolor[HTML]{F4F9FE}51.7\% (50.4-53.0\%) & \cellcolor[HTML]{E9F2FA}53.2\% (51.9-54.5\%) \\
     & \textbf{10} & \cellcolor[HTML]{DCE9F6}55.1\% (53.8-56.3\%) & \cellcolor[HTML]{D0E2F2}56.8\% (55.6-58.0\%) & \cellcolor[HTML]{D9E7F5}55.5\% (54.2-56.8\%) & \cellcolor[HTML]{CCDFF1}57.4\% (56.2-58.6\%) & \cellcolor[HTML]{D9E7F5}55.5\% (54.2-56.8\%) & \cellcolor[HTML]{CCDFF1}57.4\% (56.2-58.8\%) \\
     & \textbf{25} & \cellcolor[HTML]{C1D9ED}58.6\% (57.3-59.9\%) & \cellcolor[HTML]{A5CDE3}61.1\% (59.7-62.3\%) & \cellcolor[HTML]{B4D3E9}59.8\% (58.5-61.1\%) & \cellcolor[HTML]{9AC8E0}62.0\% (60.7-63.2\%) & \cellcolor[HTML]{B5D4E9}59.6\% (58.3-60.9\%) & \cellcolor[HTML]{97C6DF}62.1\% (60.8-63.3\%) \\
    \multirow{-4}{*}{\textbf{Dense}} & \textbf{50} & \cellcolor[HTML]{9FCAE1}61.6\% (60.3-62.9\%) & \cellcolor[HTML]{84BCDB}63.5\% (62.2-64.7\%) & \cellcolor[HTML]{9FCAE1}61.6\% (60.3-62.9\%) & \cellcolor[HTML]{7AB6D9}64.1\% (62.9-65.3\%) & \cellcolor[HTML]{A5CDE3}61.0\% (59.7-62.3\%) & \cellcolor[HTML]{81BADB}63.7\% (62.5-64.9\%) \\ \midrule
     & \textbf{5} & \cellcolor[HTML]{EEF5FC}52.5\% (51.2-53.7\%) & \cellcolor[HTML]{E3EEF8}54.0\% (52.7-55.4\%) & \cellcolor[HTML]{F5F9FE}51.5\% (50.2-52.8\%) & \cellcolor[HTML]{E3EEF9}54.0\% (52.7-55.3\%) & \cellcolor[HTML]{F7FBFF}51.3\% (50.0-52.5\%) & \cellcolor[HTML]{E7F0FA}53.6\% (52.3-54.8\%) \\
     & \textbf{10} & \cellcolor[HTML]{DAE8F6}55.3\% (54.0-56.5\%) & \cellcolor[HTML]{CADEF0}57.5\% (56.3-58.8\%) & \cellcolor[HTML]{D4E4F4}56.2\% (55.0-57.5\%) & \cellcolor[HTML]{C1D9ED}58.6\% (57.4-60.0\%) & \cellcolor[HTML]{D9E7F5}55.6\% (54.3-56.8\%) & \cellcolor[HTML]{C6DBEF}58.2\% (57.0-59.4\%) \\
     & \textbf{25} & \cellcolor[HTML]{B5D4E9}59.7\% (58.3-60.9\%) & \cellcolor[HTML]{A0CBE2}61.5\% (60.2-62.7\%) & \cellcolor[HTML]{A5CDE3}61.0\% (59.8-62.3\%) & \cellcolor[HTML]{85BCDC}63.4\% (62.1-64.6\%) & \cellcolor[HTML]{A9CFE5}60.7\% (59.5-62.0\%) & \cellcolor[HTML]{87BDDC}63.3\% (62.0-64.6\%) \\
    \multirow{-4}{*}{\textbf{Hybrid}} & \textbf{50} & \cellcolor[HTML]{A9CFE5}60.7\% (59.4-62.0\%) & \cellcolor[HTML]{8FC2DE}62.8\% (61.5-64.0\%) & \cellcolor[HTML]{8FC2DE}62.7\% (61.4-63.9\%) & \cellcolor[HTML]{6FB0D7}{\color[HTML]{FFFF00} \textbf{64.9\% (63.6-66.1\%)}} & \cellcolor[HTML]{9CC9E1}61.9\% (60.6-63.2\%) & \cellcolor[HTML]{75B4D8}64.5\% (63.3-65.8\%) \\ \midrule
     & \textbf{5} & \cellcolor[HTML]{EAF3FB}53.0\% (51.7-54.2\%) & \cellcolor[HTML]{DCEAF6}54.9\% (53.7-56.2\%) & \cellcolor[HTML]{F4F9FE}51.7\% (50.4-53.0\%) & \cellcolor[HTML]{E4EFF9}53.9\% (52.6-55.1\%) & \cellcolor[HTML]{F7FBFF}51.2\% (49.9-52.5\%) & \cellcolor[HTML]{E7F0FA}53.5\% (52.2-54.7\%) \\
     & \textbf{10} & \cellcolor[HTML]{D1E2F3}56.6\% (55.4-57.8\%) & \cellcolor[HTML]{BED8EC}58.9\% (57.6-60.1\%) & \cellcolor[HTML]{D4E4F4}56.1\% (54.9-57.4\%) & \cellcolor[HTML]{C1D9ED}58.7\% (57.5-60.0\%) & \cellcolor[HTML]{D7E6F5}55.7\% (54.5-57.1\%) & \cellcolor[HTML]{C4DAEE}58.4\% (57.1-59.6\%) \\
     & \textbf{25} & \cellcolor[HTML]{AACFE5}60.7\% (59.4-61.9\%) & \cellcolor[HTML]{8DC1DD}62.8\% (61.5-64.0\%) & \cellcolor[HTML]{A5CDE3}61.0\% (59.8-62.3\%) & \cellcolor[HTML]{87BDDC}63.2\% (61.9-64.5\%) & \cellcolor[HTML]{AACFE5}60.6\% (59.5-61.9\%) & \cellcolor[HTML]{8CC0DD}62.9\% (61.6-64.1\%) \\
    \multirow{-4}{*}{\textbf{Rerank}} & \textbf{50} & \cellcolor[HTML]{97C6DF}62.2\% (61.0-63.4\%) & \cellcolor[HTML]{7AB6D9}64.1\% (63.0-65.3\%) & \cellcolor[HTML]{91C3DE}62.6\% (61.4-63.8\%) & \cellcolor[HTML]{6FB0D7}{\color[HTML]{FFFF00} \textbf{64.9\% (63.7-66.1\%)}} & \cellcolor[HTML]{97C6DF}62.1\% (60.8-63.4\%) & \cellcolor[HTML]{75B4D8}64.4\% (63.2-65.8\%) \\ \bottomrule
    \end{tabular}
    \caption{Percent Agreement for \textit{VeriFact} vs. Ground Truth computed across all propositions extracted from Human-written Brief Hospital Course (BHC) from 100 patients. This table shows results when decompose the BHC as Atomic Claim Propositions and using \textit{VeriFact} to make a prediction in the Original 3-Label Prediction Task. The table displays the evaluation results for each combination of the following experimental variables: Top N (number of facts retrieved), Retrieval Method, Reference Context Format, and whether to retrieve facts only from current admission versus the entire EHR. Background cell color intensity is darker blue with higher agreement. The highest agreement is highlighted with yellow text. 95\% confidence intervals shown in parenthesis.}
    \label{tab:suptable3_sub1}
\end{table}

% Sentence Propositions | Original Label Space
\begin{table}[h]
    \footnotesize
    \begin{tabular}{@{}ccllllll@{}}
    \toprule
    \multicolumn{8}{c}{\textbf{\shortstack{Table B7: Percent Agreement for Human-written Brief Hospital Course \\ with Sentence Propositions \& Original 3-Label Prediction Task}}} \\ \midrule
    \multicolumn{1}{l}{} & \textbf{\begin{tabular}[c]{@{}c@{}}Reference Context \\ Format\end{tabular}} & \multicolumn{2}{c}{\textbf{Relevance Score}} & \multicolumn{2}{c}{\textbf{Absolute Time}} & \multicolumn{2}{c}{\textbf{Relative Time}} \\ \cmidrule{2-8}
    \multicolumn{1}{l}{} & \textbf{\begin{tabular}[c]{@{}c@{}}Retrieve Facts \\ Only From \\ Current Admission\end{tabular}} & \multicolumn{1}{c}{\textbf{Yes}} & \multicolumn{1}{c}{\textbf{No}} & \multicolumn{1}{c}{\textbf{Yes}} & \multicolumn{1}{c}{\textbf{No}} & \multicolumn{1}{c}{\textbf{Yes}} & \multicolumn{1}{c}{\textbf{No}} \\ \midrule
    \textbf{\begin{tabular}[c]{@{}c@{}}Retrieval \\ Method\end{tabular}} & \textbf{Top   N} &  &  &  &  &  &  \\ \midrule
     & \textbf{5} & \cellcolor[HTML]{CDDFF1}57.2\%   (55.6-59.0\%) & \cellcolor[HTML]{CCDFF1}57.4\% (55.7-59.1\%) & \cellcolor[HTML]{D9E7F5}55.5\% (53.9-57.3\%) & \cellcolor[HTML]{D6E6F4}55.9\% (54.1-57.7\%) & \cellcolor[HTML]{DBE9F6}55.2\% (53.4-57.0\%) & \cellcolor[HTML]{D5E5F4}56.1\% (54.2-57.8\%) \\
     & \textbf{10} & \cellcolor[HTML]{B5D4E9}59.7\% (58.1-61.4\%) & \cellcolor[HTML]{A6CEE4}61.0\% (59.3-62.6\%) & \cellcolor[HTML]{BDD7EC}59.0\% (57.1-60.7\%) & \cellcolor[HTML]{B4D3E9}59.8\% (58.0-61.6\%) & \cellcolor[HTML]{C1D9ED}58.6\% (57.0-60.3\%) & \cellcolor[HTML]{B4D3E9}59.8\% (58.1-61.7\%) \\
     & \textbf{25} & \cellcolor[HTML]{7FB9DA}63.8\% (62.2-65.5\%) & \cellcolor[HTML]{72B2D8}{\color[HTML]{F1F1F1} 64.7\% (63.1-66.4\%)} & \cellcolor[HTML]{94C4DF}62.4\% (60.7-64.1\%) & \cellcolor[HTML]{7AB6D9}64.1\% (62.4-65.7\%) & \cellcolor[HTML]{94C4DF}62.4\% (60.7-64.0\%) & \cellcolor[HTML]{82BBDB}63.6\% (61.9-65.3\%) \\
    \multirow{-4}{*}{\textbf{Dense}} & \textbf{50} & \cellcolor[HTML]{66ABD4}{\color[HTML]{F1F1F1} 65.5\% (63.9-67.2\%)} & \cellcolor[HTML]{60A7D2}{\color[HTML]{F1F1F1} 66.1\% (64.5-67.8\%)} & \cellcolor[HTML]{7CB7DA}64.0\% (62.3-65.6\%) & \cellcolor[HTML]{77B5D9}64.3\% (62.6-66.0\%) & \cellcolor[HTML]{7CB7DA}64.0\% (62.3-65.6\%) & \cellcolor[HTML]{66ABD4}{\color[HTML]{F1F1F1} 65.5\% (63.8-67.1\%)} \\ \midrule
     & \textbf{5} & \cellcolor[HTML]{AACFE5}60.7\% (59.0-62.3\%) & \cellcolor[HTML]{ADD0E6}60.4\% (58.8-62.1\%) & \cellcolor[HTML]{B8D5EA}59.4\% (57.8-61.0\%) & \cellcolor[HTML]{B8D5EA}59.4\% (57.8-61.1\%) & \cellcolor[HTML]{B8D5EA}59.5\% (57.8-61.2\%) & \cellcolor[HTML]{B5D4E9}59.7\% (58.0-61.5\%) \\
     & \textbf{10} & \cellcolor[HTML]{84BCDB}63.5\% (61.9-65.2\%) & \cellcolor[HTML]{75B4D8}64.4\% (62.7-66.1\%) & \cellcolor[HTML]{99C7E0}62.1\% (60.4-63.8\%) & \cellcolor[HTML]{8CC0DD}63.0\% (61.4-64.6\%) & \cellcolor[HTML]{95C5DF}62.2\% (60.6-63.9\%) & \cellcolor[HTML]{8DC1DD}62.8\% (61.2-64.5\%) \\
     & \textbf{25} & \cellcolor[HTML]{69ADD5}{\color[HTML]{F1F1F1} 65.4\% (63.6-67.0\%)} & \cellcolor[HTML]{5BA3D0}{\color[HTML]{F1F1F1} 66.5\% (64.9-68.1\%)} & \cellcolor[HTML]{81BADB}63.8\% (62.0-65.3\%) & \cellcolor[HTML]{6DAFD7}{\color[HTML]{F1F1F1} 65.0\% (63.4-66.7\%)} & \cellcolor[HTML]{7DB8DA}63.9\% (62.2-65.4\%) & \cellcolor[HTML]{6CAED6}{\color[HTML]{F1F1F1} 65.1\% (63.4-66.8\%)} \\
    \multirow{-4}{*}{\textbf{Hybrid}} & \textbf{50} & \cellcolor[HTML]{68ACD5}{\color[HTML]{F1F1F1} 65.5\% (63.8-67.1\%)} & \cellcolor[HTML]{5CA4D0}{\color[HTML]{F1F1F1} 66.5\% (64.7-68.0\%)} & \cellcolor[HTML]{84BCDB}63.5\% (61.7-65.2\%) & \cellcolor[HTML]{74B3D8}64.5\% (62.8-66.3\%) & \cellcolor[HTML]{7AB6D9}64.2\% (62.4-65.8\%) & \cellcolor[HTML]{6CAED6}{\color[HTML]{F1F1F1} 65.1\% (63.3-66.7\%)} \\ \midrule
     & \textbf{5} & \cellcolor[HTML]{A3CCE3}61.3\% (59.7-62.9\%) & \cellcolor[HTML]{A6CEE4}61.0\% (59.5-62.8\%) & \cellcolor[HTML]{B9D6EA}59.4\% (57.7-61.1\%) & \cellcolor[HTML]{B9D6EA}59.4\% (57.8-61.2\%) & \cellcolor[HTML]{B9D6EA}59.4\% (57.7-61.1\%) & \cellcolor[HTML]{B9D6EA}59.3\% (57.7-61.1\%) \\
     & \textbf{10} & \cellcolor[HTML]{7CB7DA}64.0\% (62.4-65.7\%) & \cellcolor[HTML]{7DB8DA}63.9\% (62.2-65.6\%) & \cellcolor[HTML]{97C6DF}62.2\% (60.5-63.9\%) & \cellcolor[HTML]{8DC1DD}62.8\% (61.1-64.5\%) & \cellcolor[HTML]{92C4DE}62.5\% (60.9-64.1\%) & \cellcolor[HTML]{8DC1DD}62.8\% (61.1-64.5\%) \\
     & \textbf{25} & \cellcolor[HTML]{69ADD5}{\color[HTML]{F1F1F1} 65.3\% (63.6-67.0\%)} & \cellcolor[HTML]{5BA3D0}{\color[HTML]{F1F1F1} 66.5\% (64.9-68.1\%)} & \cellcolor[HTML]{84BCDB}63.5\% (61.8-65.2\%) & \cellcolor[HTML]{6DAFD7}{\color[HTML]{F1F1F1} 65.0\% (63.3-66.7\%)} & \cellcolor[HTML]{7FB9DA}63.9\% (62.1-65.4\%) & \cellcolor[HTML]{71B1D7}{\color[HTML]{F1F1F1} 64.8\% (63.2-66.6\%)} \\
    \multirow{-4}{*}{\textbf{Rerank}} & \textbf{50} & \cellcolor[HTML]{66ABD4}{\color[HTML]{F1F1F1} 65.5\% (63.8-67.1\%)} & \cellcolor[HTML]{58A1CF}{\color[HTML]{FFFF00} \textbf{66.8\% (65.1-68.5\%)}} & \cellcolor[HTML]{7FB9DA}63.8\% (62.0-65.5\%) & \cellcolor[HTML]{75B4D8}64.4\% (62.8-66.1\%) & \cellcolor[HTML]{79B5D9}64.2\% (62.5-65.8\%) & \cellcolor[HTML]{6FB0D7}{\color[HTML]{F1F1F1} 64.9\% (63.1-66.6\%)} \\ \bottomrule
    \end{tabular}
    \caption{Percent Agreement for \textit{VeriFact} vs. Ground Truth computed across all propositions extracted from Human-written Brief Hospital Course (BHC) from 100 patients. This table shows results when decompose the BHC as Sentence Propositions and using \textit{VeriFact} to make a prediction in the Original 3-Label Prediction Task. The table displays the evaluation results for each combination of the following experimental variables: Top N (number of facts retrieved), Retrieval Method, Reference Context Format, and whether to retrieve facts only from current admission versus the entire EHR. Background cell color intensity is darker blue with higher agreement. The highest agreement is highlighted with yellow text. 95\% confidence intervals shown in parenthesis.}
    \label{tab:suptable3_sub2}
\end{table}

% Atomic Claim Propositions | Binarized Label Space
\begin{table}[h]
    \footnotesize
    \begin{tabular}{@{}ccllllll@{}}
    \toprule
    \multicolumn{8}{c}{\textbf{\shortstack{Table B8: Percent Agreement for Human-written Brief Hospital Course \\ with Atomic Claim Propositions \& Binarized 2-Label Prediction Task}}} \\ \midrule
    \multicolumn{1}{l}{} & \textbf{\begin{tabular}[c]{@{}c@{}}Reference Context \\ Format\end{tabular}} & \multicolumn{2}{c}{\textbf{Relevance Score}} & \multicolumn{2}{c}{\textbf{Absolute Time}} & \multicolumn{2}{c}{\textbf{Relative Time}} \\ \cmidrule{2-8}
    \multicolumn{1}{l}{} & \textbf{\begin{tabular}[c]{@{}c@{}}Retrieve Facts \\ Only From \\ Current Admission\end{tabular}} & \multicolumn{1}{c}{\textbf{Yes}} & \multicolumn{1}{c}{\textbf{No}} & \multicolumn{1}{c}{\textbf{Yes}} & \multicolumn{1}{c}{\textbf{No}} & \multicolumn{1}{c}{\textbf{Yes}} & \multicolumn{1}{c}{\textbf{No}} \\ \midrule
    \textbf{\begin{tabular}[c]{@{}c@{}}Retrieval \\ Method\end{tabular}} & \textbf{Top   N} &  &  &  &  &  &  \\ \midrule
     & \textbf{5} & \cellcolor[HTML]{4896C8}{\color[HTML]{F1F1F1} 68.2\%   (67.0-69.5\%)} & \cellcolor[HTML]{3787C0}{\color[HTML]{F1F1F1} 69.8\% (68.6-70.9\%)} & \cellcolor[HTML]{4A98C9}{\color[HTML]{F1F1F1} 68.0\% (66.7-69.2\%)} & \cellcolor[HTML]{3888C1}{\color[HTML]{F1F1F1} 69.7\% (68.5-70.8\%)} & \cellcolor[HTML]{4B98CA}{\color[HTML]{F1F1F1} 67.9\% (66.6-69.1\%)} & \cellcolor[HTML]{3B8BC2}{\color[HTML]{F1F1F1} 69.3\% (68.1-70.5\%)} \\
     & \textbf{10} & \cellcolor[HTML]{3383BE}{\color[HTML]{F1F1F1} 70.2\% (69.0-71.4\%)} & \cellcolor[HTML]{2070B4}{\color[HTML]{F1F1F1} 72.2\% (71.0-73.3\%)} & \cellcolor[HTML]{2A7AB9}{\color[HTML]{F1F1F1} 71.3\% (70.1-72.4\%)} & \cellcolor[HTML]{1966AD}{\color[HTML]{F1F1F1} 73.3\% (72.1-74.4\%)} & \cellcolor[HTML]{2C7CBA}{\color[HTML]{F1F1F1} 71.0\% (69.7-72.2\%)} & \cellcolor[HTML]{1966AD}{\color[HTML]{F1F1F1} 73.3\% (72.1-74.4\%)} \\
     & \textbf{25} & \cellcolor[HTML]{1B69AF}{\color[HTML]{F1F1F1} 72.9\% (71.8-74.1\%)} & \cellcolor[HTML]{0A549E}{\color[HTML]{F1F1F1} 75.2\% (74.1-76.3\%)} & \cellcolor[HTML]{0E58A2}{\color[HTML]{F1F1F1} 74.9\% (73.7-76.0\%)} & \cellcolor[HTML]{08458A}{\color[HTML]{F1F1F1} 77.0\% (75.9-78.1\%)} & \cellcolor[HTML]{0F5AA3}{\color[HTML]{F1F1F1} 74.6\% (73.4-75.8\%)} & \cellcolor[HTML]{08458A}{\color[HTML]{F1F1F1} 76.9\% (75.7-78.0\%)} \\
    \multirow{-4}{*}{\textbf{Dense}} & \textbf{50} & \cellcolor[HTML]{084E98}{\color[HTML]{F1F1F1} 76.0\% (74.9-77.1\%)} & \cellcolor[HTML]{083E81}{\color[HTML]{F1F1F1} 77.6\% (76.4-78.7\%)} & \cellcolor[HTML]{084990}{\color[HTML]{F1F1F1} 76.4\% (75.4-77.6\%)} & \cellcolor[HTML]{08306B}{\color[HTML]{F1F1F1} 79.1\% (78.0-80.2\%)} & \cellcolor[HTML]{08509B}{\color[HTML]{F1F1F1} 75.7\% (74.5-76.8\%)} & \cellcolor[HTML]{083877}{\color[HTML]{F1F1F1} 78.2\% (77.1-79.3\%)} \\ \midrule
     & \textbf{5} & \cellcolor[HTML]{3F8FC5}{\color[HTML]{F1F1F1} 68.9\% (67.7-70.1\%)} & \cellcolor[HTML]{3181BD}{\color[HTML]{F1F1F1} 70.4\% (69.3-71.6\%)} & \cellcolor[HTML]{3C8CC3}{\color[HTML]{F1F1F1} 69.2\% (68.0-70.5\%)} & \cellcolor[HTML]{2979B9}{\color[HTML]{F1F1F1} 71.4\% (70.2-72.5\%)} & \cellcolor[HTML]{4493C7}{\color[HTML]{F1F1F1} 68.5\% (67.3-69.7\%)} & \cellcolor[HTML]{2E7EBC}{\color[HTML]{F1F1F1} 70.8\% (69.6-71.9\%)} \\
     & \textbf{10} & \cellcolor[HTML]{2979B9}{\color[HTML]{F1F1F1} 71.3\% (70.1-72.4\%)} & \cellcolor[HTML]{1764AB}{\color[HTML]{F1F1F1} 73.5\% (72.4-74.6\%)} & \cellcolor[HTML]{1E6DB2}{\color[HTML]{F1F1F1} 72.5\% (71.4-73.7\%)} & \cellcolor[HTML]{0D57A1}{\color[HTML]{F1F1F1} 74.9\% (73.9-76.1\%)} & \cellcolor[HTML]{2373B6}{\color[HTML]{F1F1F1} 71.9\% (70.8-73.1\%)} & \cellcolor[HTML]{125DA6}{\color[HTML]{F1F1F1} 74.3\% (73.2-75.4\%)} \\
     & \textbf{25} & \cellcolor[HTML]{105BA4}{\color[HTML]{F1F1F1} 74.5\% (73.4-75.6\%)} & \cellcolor[HTML]{084B93}{\color[HTML]{F1F1F1} 76.3\% (75.2-77.3\%)} & \cellcolor[HTML]{084A91}{\color[HTML]{F1F1F1} 76.4\% (75.4-77.5\%)} & \cellcolor[HTML]{083370}{\color[HTML]{F1F1F1} 78.7\% (77.7-79.8\%)} & \cellcolor[HTML]{084D96}{\color[HTML]{F1F1F1} 76.0\% (74.9-77.1\%)} & \cellcolor[HTML]{083877}{\color[HTML]{F1F1F1} 78.3\% (77.2-79.3\%)} \\
    \multirow{-4}{*}{\textbf{Hybrid}} & \textbf{50} & \cellcolor[HTML]{08509B}{\color[HTML]{F1F1F1} 75.7\% (74.6-76.9\%)} & \cellcolor[HTML]{083B7C}{\color[HTML]{F1F1F1} 78.0\% (76.8-79.0\%)} & \cellcolor[HTML]{08478D}{\color[HTML]{F1F1F1} 76.8\% (75.6-77.8\%)} & \cellcolor[HTML]{08316D}{\color[HTML]{F1F1F1} 79.0\% (78.0-80.0\%)} & \cellcolor[HTML]{08468B}{\color[HTML]{F1F1F1} 76.8\% (75.7-78.0\%)} & \cellcolor[HTML]{08306B}{\color[HTML]{FFFF00} \textbf{79.2\% (78.1-80.2\%)}} \\ \midrule
     & \textbf{5} & \cellcolor[HTML]{3888C1}{\color[HTML]{F1F1F1} 69.7\% (68.6-70.9\%)} & \cellcolor[HTML]{2676B8}{\color[HTML]{F1F1F1} 71.5\% (70.4-72.7\%)} & \cellcolor[HTML]{3C8CC3}{\color[HTML]{F1F1F1} 69.2\% (68.0-70.5\%)} & \cellcolor[HTML]{2979B9}{\color[HTML]{F1F1F1} 71.4\% (70.2-72.5\%)} & \cellcolor[HTML]{4292C6}{\color[HTML]{F1F1F1} 68.6\% (67.3-69.9\%)} & \cellcolor[HTML]{3080BD}{\color[HTML]{F1F1F1} 70.6\% (69.4-71.7\%)} \\
     & \textbf{10} & \cellcolor[HTML]{1B69AF}{\color[HTML]{F1F1F1} 73.0\% (71.8-74.2\%)} & \cellcolor[HTML]{0D57A1}{\color[HTML]{F1F1F1} 75.0\% (73.9-76.1\%)} & \cellcolor[HTML]{1F6EB3}{\color[HTML]{F1F1F1} 72.5\% (71.4-73.7\%)} & \cellcolor[HTML]{0A549E}{\color[HTML]{F1F1F1} 75.2\% (74.2-76.4\%)} & \cellcolor[HTML]{2070B4}{\color[HTML]{F1F1F1} 72.2\% (71.1-73.5\%)} & \cellcolor[HTML]{105BA4}{\color[HTML]{F1F1F1} 74.5\% (73.3-75.6\%)} \\
     & \textbf{25} & \cellcolor[HTML]{09529D}{\color[HTML]{F1F1F1} 75.5\% (74.3-76.6\%)} & \cellcolor[HTML]{083D7F}{\color[HTML]{F1F1F1} 77.7\% (76.6-78.8\%)} & \cellcolor[HTML]{08478D}{\color[HTML]{F1F1F1} 76.6\% (75.6-77.8\%)} & \cellcolor[HTML]{083370}{\color[HTML]{F1F1F1} 78.8\% (77.7-79.9\%)} & \cellcolor[HTML]{084E98}{\color[HTML]{F1F1F1} 75.9\% (74.7-77.0\%)} & \cellcolor[HTML]{083979}{\color[HTML]{F1F1F1} 78.2\% (77.0-79.2\%)} \\
    \multirow{-4}{*}{\textbf{Rerank}} & \textbf{50} & \cellcolor[HTML]{084184}{\color[HTML]{F1F1F1} 77.4\% (76.3-78.5\%)} & \cellcolor[HTML]{083370}{\color[HTML]{F1F1F1} 78.8\% (77.8-79.9\%)} & \cellcolor[HTML]{08458A}{\color[HTML]{F1F1F1} 76.9\% (75.8-78.0\%)} & \cellcolor[HTML]{08316D}{\color[HTML]{F1F1F1} 79.0\% (77.9-80.0\%)} & \cellcolor[HTML]{084488}{\color[HTML]{F1F1F1} 77.0\% (75.9-78.1\%)} & \cellcolor[HTML]{08316D}{\color[HTML]{F1F1F1} 79.0\% (77.9-80.0\%)} \\ \bottomrule
    \end{tabular}
    \caption{Percent Agreement for \textit{VeriFact} vs. Ground Truth computed across all propositions extracted from Human-written Brief Hospital Course (BHC) from 100 patients. This table shows results when decompose the BHC as Atomic Claim Propositions and using \textit{VeriFact} to make a prediction in the Binarized 2-Label Prediction Task. The table displays the evaluation results for each combination of the following experimental variables: Top N (number of facts retrieved), Retrieval Method, Reference Context Format, and whether to retrieve facts only from current admission versus the entire EHR. Background cell color intensity is darker blue with higher agreement. The highest agreement is highlighted with yellow text. 95\% confidence intervals shown in parenthesis.}
    \label{tab:suptable3_sub3}
\end{table}

% Sentence Propositions | Binarized Label Space
\begin{table}[h]
    \footnotesize
    \begin{tabular}{@{}ccllllll@{}}
    \toprule
    \multicolumn{8}{c}{\textbf{\shortstack{Table B9: Percent Agreement for Human-written Brief Hospital Course \\ with Sentence Propositions \& Binarized 2-Label Prediction Task}}} \\ \midrule
    \multicolumn{1}{l}{} & \textbf{\begin{tabular}[c]{@{}c@{}}Reference Context \\ Format\end{tabular}} & \multicolumn{2}{c}{\textbf{Relevance Score}} & \multicolumn{2}{c}{\textbf{Absolute Time}} & \multicolumn{2}{c}{\textbf{Relative Time}} \\ \cmidrule{2-8}
    \multicolumn{1}{l}{} & \textbf{\begin{tabular}[c]{@{}c@{}}Retrieve Facts \\ Only From \\ Current Admission\end{tabular}} & \multicolumn{1}{c}{\textbf{Yes}} & \multicolumn{1}{c}{\textbf{No}} & \multicolumn{1}{c}{\textbf{Yes}} & \multicolumn{1}{c}{\textbf{No}} & \multicolumn{1}{c}{\textbf{Yes}} & \multicolumn{1}{c}{\textbf{No}} \\ \midrule
    \textbf{\begin{tabular}[c]{@{}c@{}}Retrieval \\ Method\end{tabular}} & \textbf{Top   N} &  &  &  &  &  &  \\ \midrule
     & \textbf{5} & \cellcolor[HTML]{1865AC}{\color[HTML]{F1F1F1} 73.5\%   (72.0-75.1\%)} & \cellcolor[HTML]{135FA7}{\color[HTML]{F1F1F1} 74.1\% (72.7-75.6\%)} & \cellcolor[HTML]{1966AD}{\color[HTML]{F1F1F1} 73.3\% (71.9-74.7\%)} & \cellcolor[HTML]{1865AC}{\color[HTML]{F1F1F1} 73.5\% (72.0-75.0\%)} & \cellcolor[HTML]{1A68AE}{\color[HTML]{F1F1F1} 73.1\% (71.7-74.7\%)} & \cellcolor[HTML]{1764AB}{\color[HTML]{F1F1F1} 73.5\% (72.0-75.1\%)} \\
     & \textbf{10} & \cellcolor[HTML]{0D57A1}{\color[HTML]{F1F1F1} 75.0\% (73.5-76.4\%)} & \cellcolor[HTML]{084C95}{\color[HTML]{F1F1F1} 76.1\% (74.7-77.6\%)} & \cellcolor[HTML]{0A549E}{\color[HTML]{F1F1F1} 75.3\% (73.9-76.8\%)} & \cellcolor[HTML]{084D96}{\color[HTML]{F1F1F1} 76.0\% (74.6-77.5\%)} & \cellcolor[HTML]{0E59A2}{\color[HTML]{F1F1F1} 74.7\% (73.3-76.2\%)} & \cellcolor[HTML]{084F99}{\color[HTML]{F1F1F1} 75.8\% (74.3-77.3\%)} \\
     & \textbf{25} & \cellcolor[HTML]{08468B}{\color[HTML]{F1F1F1} 76.8\% (75.3-78.2\%)} & \cellcolor[HTML]{083C7D}{\color[HTML]{F1F1F1} 77.8\% (76.4-79.3\%)} & \cellcolor[HTML]{08478D}{\color[HTML]{F1F1F1} 76.7\% (75.4-78.2\%)} & \cellcolor[HTML]{083E81}{\color[HTML]{F1F1F1} 77.6\% (76.2-79.1\%)} & \cellcolor[HTML]{084B93}{\color[HTML]{F1F1F1} 76.3\% (74.8-77.7\%)} & \cellcolor[HTML]{084A91}{\color[HTML]{F1F1F1} 76.4\% (74.9-77.9\%)} \\
    \multirow{-4}{*}{\textbf{Dense}} & \textbf{50} & \cellcolor[HTML]{084285}{\color[HTML]{F1F1F1} 77.3\% (76.0-78.7\%)} & \cellcolor[HTML]{083A7A}{\color[HTML]{F1F1F1} 78.0\% (76.5-79.5\%)} & \cellcolor[HTML]{084488}{\color[HTML]{F1F1F1} 77.0\% (75.5-78.4\%)} & \cellcolor[HTML]{084A91}{\color[HTML]{F1F1F1} 76.4\% (74.8-77.9\%)} & \cellcolor[HTML]{084A91}{\color[HTML]{F1F1F1} 76.4\% (74.9-77.9\%)} & \cellcolor[HTML]{084A91}{\color[HTML]{F1F1F1} 76.3\% (74.8-77.8\%)} \\ \midrule
     & \textbf{5} & \cellcolor[HTML]{08478D}{\color[HTML]{F1F1F1} 76.7\% (75.3-78.1\%)} & \cellcolor[HTML]{08458A}{\color[HTML]{F1F1F1} 76.9\% (75.5-78.4\%)} & \cellcolor[HTML]{084A91}{\color[HTML]{F1F1F1} 76.4\% (75.0-77.9\%)} & \cellcolor[HTML]{084990}{\color[HTML]{F1F1F1} 76.5\% (75.2-78.0\%)} & \cellcolor[HTML]{084990}{\color[HTML]{F1F1F1} 76.4\% (75.0-77.9\%)} & \cellcolor[HTML]{08468B}{\color[HTML]{F1F1F1} 76.8\% (75.3-78.3\%)} \\
     & \textbf{10} & \cellcolor[HTML]{083D7F}{\color[HTML]{F1F1F1} 77.6\% (76.3-79.1\%)} & \cellcolor[HTML]{083674}{\color[HTML]{FFFF00} \textbf{78.5\% (77.0-79.9\%)}} & \cellcolor[HTML]{083C7D}{\color[HTML]{F1F1F1} 77.8\% (76.4-79.1\%)} & \cellcolor[HTML]{083877}{\color[HTML]{F1F1F1} 78.2\% (76.9-79.6\%)} & \cellcolor[HTML]{083C7D}{\color[HTML]{F1F1F1} 77.8\% (76.4-79.2\%)} & \cellcolor[HTML]{083979}{\color[HTML]{F1F1F1} 78.1\% (76.7-79.6\%)} \\
     & \textbf{25} & \cellcolor[HTML]{083B7C}{\color[HTML]{F1F1F1} 77.9\% (76.5-79.2\%)} & \cellcolor[HTML]{083573}{\color[HTML]{FFFF00} \textbf{78.5\% (77.2-80.0\%)}} & \cellcolor[HTML]{083E81}{\color[HTML]{F1F1F1} 77.6\% (76.2-79.2\%)} & \cellcolor[HTML]{083573}{\color[HTML]{FFFF00} \textbf{78.5\% (77.1-80.0\%)}} & \cellcolor[HTML]{083B7C}{\color[HTML]{F1F1F1} 77.9\% (76.4-79.2\%)} & \cellcolor[HTML]{083D7F}{\color[HTML]{F1F1F1} 77.7\% (76.3-79.1\%)} \\
    \multirow{-4}{*}{\textbf{Hybrid}} & \textbf{50} & \cellcolor[HTML]{084488}{\color[HTML]{F1F1F1} 77.0\% (75.6-78.5\%)} & \cellcolor[HTML]{083979}{\color[HTML]{F1F1F1} 78.1\% (76.6-79.5\%)} & \cellcolor[HTML]{08488E}{\color[HTML]{F1F1F1} 76.6\% (75.1-78.2\%)} & \cellcolor[HTML]{08468B}{\color[HTML]{F1F1F1} 76.8\% (75.3-78.4\%)} & \cellcolor[HTML]{08488E}{\color[HTML]{F1F1F1} 76.6\% (75.2-78.1\%)} & \cellcolor[HTML]{084A91}{\color[HTML]{F1F1F1} \textbf{76.4\% (74.9-77.8\%)}} \\ \midrule
     & \textbf{5} & \cellcolor[HTML]{084387}{\color[HTML]{F1F1F1} 77.2\% (75.8-78.6\%)} & \cellcolor[HTML]{084082}{\color[HTML]{F1F1F1} 77.4\% (76.1-78.9\%)} & \cellcolor[HTML]{084A91}{\color[HTML]{F1F1F1} 76.4\% (75.0-77.8\%)} & \cellcolor[HTML]{084990}{\color[HTML]{F1F1F1} 76.5\% (75.2-78.0\%)} & \cellcolor[HTML]{084B93}{\color[HTML]{F1F1F1} 76.3\% (74.9-77.8\%)} & \cellcolor[HTML]{08478D}{\color[HTML]{F1F1F1} 76.7\% (75.1-78.3\%)} \\
     & \textbf{10} & \cellcolor[HTML]{083C7D}{\color[HTML]{F1F1F1} 77.8\% (76.4-79.3\%)} & \cellcolor[HTML]{083B7C}{\color[HTML]{F1F1F1} 78.0\% (76.5-79.6\%)} & \cellcolor[HTML]{083E81}{\color[HTML]{F1F1F1} 77.6\% (76.3-79.1\%)} & \cellcolor[HTML]{083A7A}{\color[HTML]{F1F1F1} 78.0\% (76.7-79.5\%)} & \cellcolor[HTML]{083C7D}{\color[HTML]{F1F1F1} 77.8\% (76.4-79.2\%)} & \cellcolor[HTML]{083B7C}{\color[HTML]{F1F1F1} 77.9\% (76.5-79.3\%)} \\
     & \textbf{25} & \cellcolor[HTML]{084184}{\color[HTML]{F1F1F1} 77.4\% (75.8-78.8\%)} & \cellcolor[HTML]{083877}{\color[HTML]{F1F1F1} 78.2\% (76.8-79.6\%)} & \cellcolor[HTML]{084082}{\color[HTML]{F1F1F1} 77.5\% (76.1-78.9\%)} & \cellcolor[HTML]{083776}{\color[HTML]{F1F1F1} 78.3\% (76.9-79.7\%)} & \cellcolor[HTML]{083E81}{\color[HTML]{F1F1F1} 77.6\% (76.2-79.0\%)} & \cellcolor[HTML]{083979}{\color[HTML]{F1F1F1} 78.1\% (76.7-79.5\%)} \\
    \multirow{-4}{*}{\textbf{Rerank}} & \textbf{50} & \cellcolor[HTML]{08478D}{\color[HTML]{F1F1F1} 76.7\% (75.2-78.1\%)} & \cellcolor[HTML]{083B7C}{\color[HTML]{F1F1F1} 77.9\% (76.3-79.4\%)} & \cellcolor[HTML]{08468B}{\color[HTML]{F1F1F1} 76.8\% (75.3-78.3\%)} & \cellcolor[HTML]{08478D}{\color[HTML]{F1F1F1} 76.7\% (75.1-78.2\%)} & \cellcolor[HTML]{08468B}{\color[HTML]{F1F1F1} 76.9\% (75.3-78.3\%)} & \cellcolor[HTML]{084A91}{\color[HTML]{F1F1F1} 76.4\% (74.8-77.9\%)} \\ \bottomrule
    \end{tabular}
    \caption{Percent Agreement for \textit{VeriFact} vs. Ground Truth computed across all propositions extracted from Human-written Brief Hospital Course (BHC) from 100 patients. This table shows results when decompose the BHC as Sentence Propositions and using \textit{VeriFact} to make a prediction in the Binarized 2-Label Prediction Task. The table displays the evaluation results for each combination of the following experimental variables: Top N (number of facts retrieved), Retrieval Method, Reference Context Format, and whether to retrieve facts only from current admission versus the entire EHR. Background cell color intensity is darker blue with higher agreement. The highest agreement is highlighted with yellow text. 95\% confidence intervals shown in parenthesis.}
    \label{tab:suptable3_sub4}
\end{table}

\clearpage
\subsection{Gwet's AC1 for \textit{VeriFact} vs. Ground Truth with LLM-written Brief Hospital Course}\label{suptables_sub4}
% Atomic Claim Propositions | Original Label Space
\begin{table}[h]
    \footnotesize
    \begin{tabular}{@{}ccllllll@{}}
    \toprule
    \multicolumn{8}{c}{\textbf{\shortstack{Table B10: Gwet's AC1 for LLM-written Brief Hospital Course \\ with Atomic Claim Propositions \& Original 3-Label Prediction Task}}} \\ \midrule
    \multicolumn{1}{l}{} & \textbf{\begin{tabular}[c]{@{}c@{}}Reference Context \\ Format\end{tabular}} & \multicolumn{2}{c}{\textbf{Relevance Score}} & \multicolumn{2}{c}{\textbf{Absolute Time}} & \multicolumn{2}{c}{\textbf{Relative Time}} \\ \cmidrule{2-8}
    \multicolumn{1}{l}{} & \textbf{\begin{tabular}[c]{@{}c@{}}Retrieve Facts \\ Only From \\ Current Admission\end{tabular}} & \multicolumn{1}{c}{\textbf{Yes}} & \multicolumn{1}{c}{\textbf{No}} & \multicolumn{1}{c}{\textbf{Yes}} & \multicolumn{1}{c}{\textbf{No}} & \multicolumn{1}{c}{\textbf{Yes}} & \multicolumn{1}{c}{\textbf{No}} \\ \midrule
    \textbf{\begin{tabular}[c]{@{}c@{}}Retrieval \\ Method\end{tabular}} & \textbf{Top   N} &  &  &  &  &  &  \\ \midrule
     & \textbf{5} & \cellcolor[HTML]{CFE1F2}0.52 (0.49-0.54) & \cellcolor[HTML]{AACFE5}0.58 (0.56-0.61) & \cellcolor[HTML]{C3DAEE}0.55 (0.52-0.57) & \cellcolor[HTML]{8CC0DD}0.63 (0.61-0.65) & \cellcolor[HTML]{CADDF0}0.53 (0.51-0.55) & \cellcolor[HTML]{A3CCE3}0.60 (0.58-0.62) \\
     & \textbf{10} & \cellcolor[HTML]{A0CBE2}0.60 (0.58-0.62) & \cellcolor[HTML]{71B1D7}{\color[HTML]{F1F1F1} 0.66 (0.64-0.68)} & \cellcolor[HTML]{74B3D8}0.66 (0.64-0.68) & \cellcolor[HTML]{4D99CA}{\color[HTML]{F1F1F1} 0.72 (0.70-0.73)} & \cellcolor[HTML]{89BEDC}0.63 (0.61-0.65) & \cellcolor[HTML]{5BA3D0}{\color[HTML]{F1F1F1} 0.69 (0.68-0.71)} \\
     & \textbf{25} & \cellcolor[HTML]{57A0CE}{\color[HTML]{F1F1F1} 0.70 (0.68-0.72)} & \cellcolor[HTML]{3484BF}{\color[HTML]{F1F1F1} 0.76 (0.74-0.77)} & \cellcolor[HTML]{2D7DBB}{\color[HTML]{F1F1F1} 0.77 (0.76-0.79)} & \cellcolor[HTML]{1663AA}{\color[HTML]{F1F1F1} 0.82 (0.81-0.84)} & \cellcolor[HTML]{4191C6}{\color[HTML]{F1F1F1} 0.73 (0.71-0.75)} & \cellcolor[HTML]{2070B4}{\color[HTML]{F1F1F1} 0.80 (0.78-0.81)} \\
    \multirow{-4}{*}{\textbf{Dense}} & \textbf{50} & \cellcolor[HTML]{3686C0}{\color[HTML]{F1F1F1} 0.75 (0.74-0.77)} & \cellcolor[HTML]{1967AD}{\color[HTML]{F1F1F1} 0.81 (0.80-0.83)} & \cellcolor[HTML]{1B69AF}{\color[HTML]{F1F1F1} 0.81 (0.80-0.83)} & \cellcolor[HTML]{084D96}{\color[HTML]{F1F1F1} 0.87 (0.85-0.88)} & \cellcolor[HTML]{2373B6}{\color[HTML]{F1F1F1} 0.79 (0.78-0.81)} & \cellcolor[HTML]{0A549E}{\color[HTML]{F1F1F1} 0.85 (0.84-0.87)} \\ \midrule
     & \textbf{5} & \cellcolor[HTML]{BDD7EC}0.56 (0.53-0.58) & \cellcolor[HTML]{92C4DE}0.62 (0.60-0.64) & \cellcolor[HTML]{AACFE5}0.59 (0.56-0.61) & \cellcolor[HTML]{74B3D8}0.66 (0.64-0.68) & \cellcolor[HTML]{B9D6EA}0.56 (0.54-0.58) & \cellcolor[HTML]{89BEDC}0.63 (0.61-0.65) \\
     & \textbf{10} & \cellcolor[HTML]{81BADB}0.64 (0.62-0.66) & \cellcolor[HTML]{539ECD}{\color[HTML]{F1F1F1} 0.71 (0.69-0.72)} & \cellcolor[HTML]{61A7D2}{\color[HTML]{F1F1F1} 0.68 (0.67-0.70)} & \cellcolor[HTML]{3D8DC4}{\color[HTML]{F1F1F1} 0.74 (0.72-0.76)} & \cellcolor[HTML]{6CAED6}{\color[HTML]{F1F1F1} 0.67 (0.65-0.69)} & \cellcolor[HTML]{4292C6}{\color[HTML]{F1F1F1} 0.73 (0.71-0.75)} \\
     & \textbf{25} & \cellcolor[HTML]{4D99CA}{\color[HTML]{F1F1F1} 0.72 (0.70-0.73)} & \cellcolor[HTML]{2E7EBC}{\color[HTML]{F1F1F1} 0.77 (0.75-0.79)} & \cellcolor[HTML]{2474B7}{\color[HTML]{F1F1F1} 0.79 (0.77-0.80)} & \cellcolor[HTML]{125DA6}{\color[HTML]{F1F1F1} 0.83 (0.82-0.85)} & \cellcolor[HTML]{3383BE}{\color[HTML]{F1F1F1} 0.76 (0.75-0.78)} & \cellcolor[HTML]{1663AA}{\color[HTML]{F1F1F1} 0.82 (0.81-0.84)} \\
    \multirow{-4}{*}{\textbf{Hybrid}} & \textbf{50} & \cellcolor[HTML]{3181BD}{\color[HTML]{F1F1F1} 0.76 (0.75-0.78)} & \cellcolor[HTML]{1A68AE}{\color[HTML]{F1F1F1} 0.81 (0.80-0.83)} & \cellcolor[HTML]{1663AA}{\color[HTML]{F1F1F1} 0.82 (0.81-0.84)} & \cellcolor[HTML]{08488E}{\color[HTML]{FFFF00} \textbf{0.88 (0.87-0.89)}} & \cellcolor[HTML]{2575B7}{\color[HTML]{F1F1F1} 0.79 (0.77-0.80)} & \cellcolor[HTML]{0A539E}{\color[HTML]{F1F1F1} 0.86 (0.84-0.87)} \\ \midrule
     & \textbf{5} & \cellcolor[HTML]{A6CEE4}0.59 (0.57-0.61) & \cellcolor[HTML]{77B5D9}0.65 (0.64-0.67) & \cellcolor[HTML]{A9CFE5}0.59 (0.57-0.61) & \cellcolor[HTML]{72B2D8}{\color[HTML]{F1F1F1} 0.66 (0.64-0.68)} & \cellcolor[HTML]{B8D5EA}0.56 (0.54-0.59) & \cellcolor[HTML]{89BEDC}0.63 (0.61-0.65) \\
     & \textbf{10} & \cellcolor[HTML]{6AAED6}{\color[HTML]{F1F1F1} 0.67 (0.65-0.69)} & \cellcolor[HTML]{4493C7}{\color[HTML]{F1F1F1} 0.73 (0.71-0.75)} & \cellcolor[HTML]{60A7D2}{\color[HTML]{F1F1F1} 0.68 (0.67-0.70)} & \cellcolor[HTML]{3C8CC3}{\color[HTML]{F1F1F1} 0.74 (0.73-0.76)} & \cellcolor[HTML]{6FB0D7}{\color[HTML]{F1F1F1} 0.66 (0.64-0.68)} & \cellcolor[HTML]{4191C6}{\color[HTML]{F1F1F1} 0.73 (0.72-0.75)} \\
     & \textbf{25} & \cellcolor[HTML]{3C8CC3}{\color[HTML]{F1F1F1} 0.74 (0.72-0.76)} & \cellcolor[HTML]{1E6DB2}{\color[HTML]{F1F1F1} 0.80 (0.79-0.82)} & \cellcolor[HTML]{2474B7}{\color[HTML]{F1F1F1} 0.79 (0.77-0.80)} & \cellcolor[HTML]{105BA4}{\color[HTML]{F1F1F1} 0.84 (0.83-0.85)} & \cellcolor[HTML]{3383BE}{\color[HTML]{F1F1F1} 0.76 (0.75-0.78)} & \cellcolor[HTML]{1764AB}{\color[HTML]{F1F1F1} 0.82 (0.81-0.84)} \\
    \multirow{-4}{*}{\textbf{Rerank}} & \textbf{50} & \cellcolor[HTML]{2171B5}{\color[HTML]{F1F1F1} 0.79 (0.78-0.81)} & \cellcolor[HTML]{0D57A1}{\color[HTML]{F1F1F1} 0.85 (0.83-0.86)} & \cellcolor[HTML]{1865AC}{\color[HTML]{F1F1F1} 0.82 (0.80-0.83)} & \cellcolor[HTML]{08478D}{\color[HTML]{FFFF00} \textbf{0.88 (0.87-0.89)}} & \cellcolor[HTML]{2474B7}{\color[HTML]{F1F1F1} 0.79 (0.77-0.80)} & \cellcolor[HTML]{09529D}{\color[HTML]{F1F1F1} 0.86 (0.84-0.87)} \\ \bottomrule
    \end{tabular}
    \caption{Gwet's AC1 for \textit{VeriFact} vs. Ground Truth computed across all propositions extracted from LLM-written Brief Hospital Course (BHC) from 100 patients. This table shows results when decompose the BHC as Atomic Claim Propositions and using \textit{VeriFact} to make a prediction in the Original 3-Label Prediction Task. The table displays the evaluation results for each combination of the following experimental variables: Top N (number of facts retrieved), Retrieval Method, Reference Context Format, and whether to retrieve facts only from current admission versus the entire EHR. Background cell color intensity is darker blue with higher agreement. The highest agreement is highlighted with yellow text. 95\% confidence intervals shown in parenthesis.}
    \label{tab:suptable4_sub1}
\end{table}

% Sentence Propositions | Original Label Space
\begin{table}[h]
    \footnotesize
    \begin{tabular}{@{}ccllllll@{}}
    \toprule
    \multicolumn{8}{c}{\textbf{\shortstack{Table B11: Gwet's AC1 for LLM-written Brief Hospital Course \\ with Sentence Propositions \& Original 3-Label Prediction Task}}} \\ \midrule
    \multicolumn{1}{l}{} & \textbf{\begin{tabular}[c]{@{}c@{}}Reference Context \\ Format\end{tabular}} & \multicolumn{2}{c}{\textbf{Relevance Score}} & \multicolumn{2}{c}{\textbf{Absolute Time}} & \multicolumn{2}{c}{\textbf{Relative Time}} \\ \cmidrule{2-8}
    \multicolumn{1}{l}{} & \textbf{\begin{tabular}[c]{@{}c@{}}Retrieve Facts \\ Only From \\ Current Admission\end{tabular}} & \multicolumn{1}{c}{\textbf{Yes}} & \multicolumn{1}{c}{\textbf{No}} & \multicolumn{1}{c}{\textbf{Yes}} & \multicolumn{1}{c}{\textbf{No}} & \multicolumn{1}{c}{\textbf{Yes}} & \multicolumn{1}{c}{\textbf{No}} \\ \midrule
    \textbf{\begin{tabular}[c]{@{}c@{}}Retrieval \\ Method\end{tabular}} & \textbf{Top   N} &  &  &  &  &  &  \\ \midrule
     & \textbf{5} & \cellcolor[HTML]{B7D4EA}0.57   (0.52-0.61) & \cellcolor[HTML]{B0D2E7}0.58 (0.53-0.62) & \cellcolor[HTML]{C2D9EE}0.55 (0.50-0.59) & \cellcolor[HTML]{BDD7EC}0.56 (0.51-0.60) & \cellcolor[HTML]{C7DCEF}0.54 (0.49-0.58) & \cellcolor[HTML]{B4D3E9}0.57 (0.53-0.61) \\
     & \textbf{10} & \cellcolor[HTML]{529DCC}{\color[HTML]{F1F1F1} 0.71 (0.67-0.74)} & \cellcolor[HTML]{3F8FC5}{\color[HTML]{F1F1F1} 0.74 (0.70-0.77)} & \cellcolor[HTML]{4997C9}{\color[HTML]{F1F1F1} 0.72 (0.69-0.76)} & \cellcolor[HTML]{3484BF}{\color[HTML]{F1F1F1} 0.76 (0.73-0.79)} & \cellcolor[HTML]{5DA5D1}{\color[HTML]{F1F1F1} 0.69 (0.65-0.72)} & \cellcolor[HTML]{3686C0}{\color[HTML]{F1F1F1} 0.76 (0.72-0.79)} \\
     & \textbf{25} & \cellcolor[HTML]{135FA7}{\color[HTML]{F1F1F1} 0.83 (0.81-0.86)} & \cellcolor[HTML]{0A539E}{\color[HTML]{F1F1F1} 0.86 (0.83-0.88)} & \cellcolor[HTML]{0E58A2}{\color[HTML]{F1F1F1} 0.85 (0.82-0.87)} & \cellcolor[HTML]{08488E}{\color[HTML]{F1F1F1} 0.88 (0.85-0.90)} & \cellcolor[HTML]{1562A9}{\color[HTML]{F1F1F1} 0.82 (0.80-0.85)} & \cellcolor[HTML]{08488E}{\color[HTML]{F1F1F1} 0.88 (0.85-0.90)} \\
    \multirow{-4}{*}{\textbf{Dense}} & \textbf{50} & \cellcolor[HTML]{084285}{\color[HTML]{F1F1F1} 0.89 (0.87-0.91)} & \cellcolor[HTML]{083877}{\color[HTML]{F1F1F1} 0.91 (0.89-0.92)} & \cellcolor[HTML]{08468B}{\color[HTML]{F1F1F1} 0.88 (0.86-0.90)} & \cellcolor[HTML]{083573}{\color[HTML]{F1F1F1} 0.91 (0.89-0.93)} & \cellcolor[HTML]{08488E}{\color[HTML]{F1F1F1} 0.88 (0.86-0.90)} & \cellcolor[HTML]{083776}{\color[HTML]{F1F1F1} 0.91 (0.89-0.93)} \\ \midrule
     & \textbf{5} & \cellcolor[HTML]{77B5D9}0.65 (0.62-0.69) & \cellcolor[HTML]{6AAED6}{\color[HTML]{F1F1F1} 0.67 (0.64-0.71)} & \cellcolor[HTML]{85BCDC}0.64 (0.60-0.67) & \cellcolor[HTML]{6DAFD7}{\color[HTML]{F1F1F1} 0.66 (0.63-0.70)} & \cellcolor[HTML]{8CC0DD}0.63 (0.59-0.67) & \cellcolor[HTML]{69ADD5}{\color[HTML]{F1F1F1} 0.67 (0.64-0.71)} \\
     & \textbf{10} & \cellcolor[HTML]{3282BE}{\color[HTML]{F1F1F1} 0.76 (0.73-0.79)} & \cellcolor[HTML]{1F6EB3}{\color[HTML]{F1F1F1} 0.80 (0.77-0.83)} & \cellcolor[HTML]{2F7FBC}{\color[HTML]{F1F1F1} 0.77 (0.74-0.80)} & \cellcolor[HTML]{1966AD}{\color[HTML]{F1F1F1} 0.82 (0.79-0.84)} & \cellcolor[HTML]{3282BE}{\color[HTML]{F1F1F1} 0.76 (0.73-0.79)} & \cellcolor[HTML]{1967AD}{\color[HTML]{F1F1F1} 0.81 (0.79-0.84)} \\
     & \textbf{25} & \cellcolor[HTML]{084F99}{\color[HTML]{F1F1F1} 0.86 (0.84-0.89)} & \cellcolor[HTML]{084184}{\color[HTML]{F1F1F1} 0.89 (0.87-0.91)} & \cellcolor[HTML]{084F99}{\color[HTML]{F1F1F1} 0.86 (0.84-0.88)} & \cellcolor[HTML]{083776}{\color[HTML]{F1F1F1} 0.91 (0.89-0.93)} & \cellcolor[HTML]{0A539E}{\color[HTML]{F1F1F1} 0.86 (0.83-0.88)} & \cellcolor[HTML]{083D7F}{\color[HTML]{F1F1F1} 0.90 (0.88-0.92)} \\
    \multirow{-4}{*}{\textbf{Hybrid}} & \textbf{50} & \cellcolor[HTML]{083E81}{\color[HTML]{F1F1F1} 0.90 (0.88-0.91)} & \cellcolor[HTML]{083471}{\color[HTML]{FFFF00} \textbf{0.92 (0.90-0.93)}} & \cellcolor[HTML]{083D7F}{\color[HTML]{F1F1F1} 0.90 (0.88-0.92)} & \cellcolor[HTML]{08316D}{\color[HTML]{FFFF00} \textbf{0.92 (0.90-0.94)}} & \cellcolor[HTML]{084990}{\color[HTML]{F1F1F1} 0.87 (0.85-0.90)} & \cellcolor[HTML]{083573}{\color[HTML]{F1F1F1} 0.91 (0.89-0.93)} \\ \midrule
     & \textbf{5} & \cellcolor[HTML]{6DAFD7}{\color[HTML]{F1F1F1} 0.67 (0.63-0.70)} & \cellcolor[HTML]{5FA6D1}{\color[HTML]{F1F1F1} 0.69 (0.65-0.72)} & \cellcolor[HTML]{82BBDB}0.64 (0.60-0.67) & \cellcolor[HTML]{72B2D8}{\color[HTML]{F1F1F1} 0.66 (0.62-0.69)} & \cellcolor[HTML]{8CC0DD}0.63 (0.59-0.67) & \cellcolor[HTML]{66ABD4}{\color[HTML]{F1F1F1} 0.68 (0.64-0.71)} \\
     & \textbf{10} & \cellcolor[HTML]{2979B9}{\color[HTML]{F1F1F1} 0.78 (0.75-0.81)} & \cellcolor[HTML]{1C6AB0}{\color[HTML]{F1F1F1} 0.81 (0.78-0.84)} & \cellcolor[HTML]{2E7EBC}{\color[HTML]{F1F1F1} 0.77 (0.74-0.80)} & \cellcolor[HTML]{1865AC}{\color[HTML]{F1F1F1} 0.82 (0.79-0.85)} & \cellcolor[HTML]{3686C0}{\color[HTML]{F1F1F1} 0.76 (0.72-0.79)} & \cellcolor[HTML]{1967AD}{\color[HTML]{F1F1F1} 0.82 (0.79-0.84)} \\
     & \textbf{25} & \cellcolor[HTML]{08468B}{\color[HTML]{F1F1F1} 0.88 (0.86-0.90)} & \cellcolor[HTML]{083877}{\color[HTML]{F1F1F1} 0.91 (0.89-0.93)} & \cellcolor[HTML]{084E98}{\color[HTML]{F1F1F1} 0.86 (0.84-0.89)} & \cellcolor[HTML]{083674}{\color[HTML]{F1F1F1} 0.91 (0.89-0.93)} & \cellcolor[HTML]{0A539E}{\color[HTML]{F1F1F1} 0.86 (0.83-0.88)} & \cellcolor[HTML]{083A7A}{\color[HTML]{F1F1F1} 0.90 (0.88-0.92)} \\
    \multirow{-4}{*}{\textbf{Rerank}} & \textbf{50} & \cellcolor[HTML]{083979}{\color[HTML]{F1F1F1} 0.90 (0.88-0.92)} & \cellcolor[HTML]{08306B}{\color[HTML]{FFFF00} \textbf{0.92 (0.91-0.94)}} & \cellcolor[HTML]{083E81}{\color[HTML]{F1F1F1} 0.89 (0.87-0.92)} & \cellcolor[HTML]{083370}{\color[HTML]{FFFF00} \textbf{0.92 (0.90-0.93)}} & \cellcolor[HTML]{084C95}{\color[HTML]{F1F1F1} 0.87 (0.85-0.89)} & \cellcolor[HTML]{08326E}{\color[HTML]{FFFF00} \textbf{0.92 (0.90-0.94)}} \\ \bottomrule
    \end{tabular}
    \caption{Gwet's AC1 for \textit{VeriFact} vs. Ground Truth computed across all propositions extracted from LLM-written Brief Hospital Course (BHC) from 100 patients. This table shows results when decompose the BHC as Sentence Propositions and using \textit{VeriFact} to make a prediction in the Original 3-Label Prediction Task. The table displays the evaluation results for each combination of the following experimental variables: Top N (number of facts retrieved), Retrieval Method, Reference Context Format, and whether to retrieve facts only from current admission versus the entire EHR. Background cell color intensity is darker blue with higher agreement. The highest agreement is highlighted with yellow text. 95\% confidence intervals shown in parenthesis.}
    \label{tab:suptable4_sub2}
\end{table}

% Atomic Claim Propositions | Binarized Label Space
\begin{table}[h]
    \footnotesize
    \begin{tabular}{@{}ccllllll@{}}
    \toprule
    \multicolumn{8}{c}{\textbf{\shortstack{Table B12: Gwet's AC1 for LLM-written Brief Hospital Course \\ with Atomic Claim Propositions \& Binarized 2-Label Prediction Task}}} \\ \midrule
    \multicolumn{1}{l}{} & \textbf{\begin{tabular}[c]{@{}c@{}}Reference Context \\ Format\end{tabular}} & \multicolumn{2}{c}{\textbf{Relevance Score}} & \multicolumn{2}{c}{\textbf{Absolute Time}} & \multicolumn{2}{c}{\textbf{Relative Time}} \\ \cmidrule{2-8}
    \multicolumn{1}{l}{} & \textbf{\begin{tabular}[c]{@{}c@{}}Retrieve Facts \\ Only From \\ Current Admission\end{tabular}} & \multicolumn{1}{c}{\textbf{Yes}} & \multicolumn{1}{c}{\textbf{No}} & \multicolumn{1}{c}{\textbf{Yes}} & \multicolumn{1}{c}{\textbf{No}} & \multicolumn{1}{c}{\textbf{Yes}} & \multicolumn{1}{c}{\textbf{No}} \\ \midrule
    \textbf{\begin{tabular}[c]{@{}c@{}}Retrieval \\ Method\end{tabular}} & \textbf{Top   N} &  &  &  &  &  &  \\ \midrule
     & \textbf{5} & \cellcolor[HTML]{F7FBFF}0.41   (0.38-0.45) & \cellcolor[HTML]{D3E4F3}0.51 (0.48-0.54) & \cellcolor[HTML]{E8F1FA}0.45 (0.42-0.48) & \cellcolor[HTML]{B9D6EA}0.56 (0.53-0.59) & \cellcolor[HTML]{F0F6FD}0.43 (0.40-0.47) & \cellcolor[HTML]{CDE0F1}0.52 (0.49-0.55) \\
     & \textbf{10} & \cellcolor[HTML]{CBDEF1}0.53 (0.50-0.56) & \cellcolor[HTML]{9AC8E0}0.61 (0.58-0.64) & \cellcolor[HTML]{A0CBE2}0.60 (0.58-0.63) & \cellcolor[HTML]{64A9D3}{\color[HTML]{F1F1F1} 0.68 (0.65-0.70)} & \cellcolor[HTML]{B4D3E9}0.57 (0.54-0.60) & \cellcolor[HTML]{79B5D9}0.65 (0.63-0.67) \\
     & \textbf{25} & \cellcolor[HTML]{72B2D8}{\color[HTML]{F1F1F1} 0.66 (0.64-0.68)} & \cellcolor[HTML]{4191C6}{\color[HTML]{F1F1F1} 0.73 (0.71-0.75)} & \cellcolor[HTML]{3888C1}{\color[HTML]{F1F1F1} 0.75 (0.73-0.77)} & \cellcolor[HTML]{1A68AE}{\color[HTML]{F1F1F1} 0.81 (0.80-0.83)} & \cellcolor[HTML]{549FCD}{\color[HTML]{F1F1F1} 0.70 (0.68-0.72)} & \cellcolor[HTML]{2979B9}{\color[HTML]{F1F1F1} 0.78 (0.76-0.80)} \\
    \multirow{-4}{*}{\textbf{Dense}} & \textbf{50} & \cellcolor[HTML]{4292C6}{\color[HTML]{F1F1F1} 0.73 (0.71-0.75)} & \cellcolor[HTML]{1F6EB3}{\color[HTML]{F1F1F1} 0.80 (0.78-0.82)} & \cellcolor[HTML]{206FB4}{\color[HTML]{F1F1F1} 0.80 (0.78-0.82)} & \cellcolor[HTML]{08509B}{\color[HTML]{F1F1F1} 0.86 (0.85-0.87)} & \cellcolor[HTML]{2C7CBA}{\color[HTML]{F1F1F1} 0.77 (0.76-0.79)} & \cellcolor[HTML]{0D57A1}{\color[HTML]{F1F1F1} 0.85 (0.83-0.86)} \\ \midrule
     & \textbf{5} & \cellcolor[HTML]{E2EDF8}0.47 (0.44-0.50) & \cellcolor[HTML]{BDD7EC}0.56 (0.53-0.58) & \cellcolor[HTML]{D2E3F3}0.51 (0.48-0.54) & \cellcolor[HTML]{9FCAE1}0.60 (0.58-0.63) & \cellcolor[HTML]{DFECF7}0.47 (0.44-0.50) & \cellcolor[HTML]{B4D3E9}0.57 (0.54-0.60) \\
     & \textbf{10} & \cellcolor[HTML]{AACFE5}0.59 (0.56-0.61) & \cellcolor[HTML]{6AAED6}{\color[HTML]{F1F1F1} 0.67 (0.65-0.69)} & \cellcolor[HTML]{82BBDB}0.64 (0.62-0.66) & \cellcolor[HTML]{4D99CA}{\color[HTML]{F1F1F1} 0.71 (0.69-0.74)} & \cellcolor[HTML]{94C4DF}0.62 (0.59-0.64) & \cellcolor[HTML]{56A0CE}{\color[HTML]{F1F1F1} 0.70 (0.68-0.72)} \\
     & \textbf{25} & \cellcolor[HTML]{63A8D3}{\color[HTML]{F1F1F1} 0.68 (0.66-0.71)} & \cellcolor[HTML]{3989C1}{\color[HTML]{F1F1F1} 0.75 (0.73-0.77)} & \cellcolor[HTML]{2F7FBC}{\color[HTML]{F1F1F1} 0.77 (0.75-0.79)} & \cellcolor[HTML]{1561A9}{\color[HTML]{F1F1F1} 0.83 (0.81-0.84)} & \cellcolor[HTML]{3F8FC5}{\color[HTML]{F1F1F1} 0.74 (0.72-0.76)} & \cellcolor[HTML]{1B69AF}{\color[HTML]{F1F1F1} 0.81 (0.79-0.83)} \\
    \multirow{-4}{*}{\textbf{Hybrid}} & \textbf{50} & \cellcolor[HTML]{3D8DC4}{\color[HTML]{F1F1F1} 0.74 (0.72-0.76)} & \cellcolor[HTML]{1F6EB3}{\color[HTML]{F1F1F1} 0.80 (0.78-0.82)} & \cellcolor[HTML]{1C6AB0}{\color[HTML]{F1F1F1} 0.81 (0.79-0.83)} & \cellcolor[HTML]{084A91}{\color[HTML]{F1F1F1} 0.87 (0.86-0.89)} & \cellcolor[HTML]{2E7EBC}{\color[HTML]{F1F1F1} 0.77 (0.75-0.79)} & \cellcolor[HTML]{0C56A0}{\color[HTML]{F1F1F1} 0.85 (0.83-0.86)} \\ \midrule
     & \textbf{5} & \cellcolor[HTML]{CFE1F2}0.52 (0.49-0.55) & \cellcolor[HTML]{A0CBE2}0.60 (0.58-0.63) & \cellcolor[HTML]{D1E2F3}0.51 (0.48-0.54) & \cellcolor[HTML]{9AC8E0}0.61 (0.58-0.64) & \cellcolor[HTML]{DEEBF7}0.48 (0.45-0.51) & \cellcolor[HTML]{B3D3E8}0.57 (0.54-0.60) \\
     & \textbf{10} & \cellcolor[HTML]{91C3DE}0.62 (0.60-0.65) & \cellcolor[HTML]{56A0CE}{\color[HTML]{F1F1F1} 0.70 (0.68-0.72)} & \cellcolor[HTML]{82BBDB}0.64 (0.62-0.66) & \cellcolor[HTML]{4D99CA}{\color[HTML]{F1F1F1} 0.71 (0.69-0.74)} & \cellcolor[HTML]{99C7E0}0.61 (0.59-0.64) & \cellcolor[HTML]{549FCD}{\color[HTML]{F1F1F1} 0.70 (0.68-0.73)} \\
     & \textbf{25} & \cellcolor[HTML]{4D99CA}{\color[HTML]{F1F1F1} 0.71 (0.69-0.74)} & \cellcolor[HTML]{2373B6}{\color[HTML]{F1F1F1} 0.79 (0.77-0.81)} & \cellcolor[HTML]{2E7EBC}{\color[HTML]{F1F1F1} 0.77 (0.75-0.79)} & \cellcolor[HTML]{135FA7}{\color[HTML]{F1F1F1} 0.83 (0.82-0.85)} & \cellcolor[HTML]{3F8FC5}{\color[HTML]{F1F1F1} 0.74 (0.72-0.76)} & \cellcolor[HTML]{1B69AF}{\color[HTML]{F1F1F1} 0.81 (0.79-0.83)} \\
    \multirow{-4}{*}{\textbf{Rerank}} & \textbf{50} & \cellcolor[HTML]{2A7AB9}{\color[HTML]{F1F1F1} 0.78 (0.76-0.80)} & \cellcolor[HTML]{105BA4}{\color[HTML]{F1F1F1} 0.84 (0.82-0.85)} & \cellcolor[HTML]{1D6CB1}{\color[HTML]{F1F1F1} 0.81 (0.79-0.82)} & \cellcolor[HTML]{084990}{\color[HTML]{FFFF00} \textbf{0.88 (0.86-0.89)}} & \cellcolor[HTML]{2E7EBC}{\color[HTML]{F1F1F1} 0.77 (0.75-0.79)} & \cellcolor[HTML]{0B559F}{\color[HTML]{F1F1F1} 0.85 (0.83-0.87)} \\ \bottomrule
    \end{tabular}
    \caption{Gwet's AC1 for \textit{VeriFact} vs. Ground Truth computed across all propositions extracted from LLM-written Brief Hospital Course (BHC) from 100 patients. This table shows results when decompose the BHC as Atomic Claim Propositions and using \textit{VeriFact} to make a prediction in the Binarized 2-Label Prediction Task. The table displays the evaluation results for each combination of the following experimental variables: Top N (number of facts retrieved), Retrieval Method, Reference Context Format, and whether to retrieve facts only from current admission versus the entire EHR. Background cell color intensity is darker blue with higher agreement. The highest agreement is highlighted with yellow text. 95\% confidence intervals shown in parenthesis.}
    \label{tab:suptable4_sub3}
\end{table}

% Sentence Propositions | Binarized Label Space
\begin{table}[h]
    \footnotesize
    \begin{tabular}{@{}ccllllll@{}}
    \toprule
    \multicolumn{8}{c}{\textbf{\shortstack{Table B13: Gwet's AC1 for LLM-written Brief Hospital Course \\ with Sentence Propositions \& Binarized 2-Label Prediction Task}}} \\ \midrule
    \multicolumn{1}{l}{} & \textbf{\begin{tabular}[c]{@{}c@{}}Reference Context \\ Format\end{tabular}} & \multicolumn{2}{c}{\textbf{Relevance Score}} & \multicolumn{2}{c}{\textbf{Absolute Time}} & \multicolumn{2}{c}{\textbf{Relative Time}} \\ \cmidrule{2-8}
    \multicolumn{1}{l}{} & \textbf{\begin{tabular}[c]{@{}c@{}}Retrieve Facts \\ Only From \\ Current Admission\end{tabular}} & \multicolumn{1}{c}{\textbf{Yes}} & \multicolumn{1}{c}{\textbf{No}} & \multicolumn{1}{c}{\textbf{Yes}} & \multicolumn{1}{c}{\textbf{No}} & \multicolumn{1}{c}{\textbf{Yes}} & \multicolumn{1}{c}{\textbf{No}} \\ \midrule
    \textbf{\begin{tabular}[c]{@{}c@{}}Retrieval \\ Method\end{tabular}} & \textbf{Top   N} &  &  &  &  &  &  \\ \midrule
     & \textbf{5} & \cellcolor[HTML]{DDEAF7}0.48   (0.42-0.54) & \cellcolor[HTML]{D6E6F4}0.50 (0.44-0.55) & \cellcolor[HTML]{E8F1FA}0.45 (0.39-0.51) & \cellcolor[HTML]{E3EEF9}0.46 (0.39-0.52) & \cellcolor[HTML]{EDF4FC}0.44 (0.37-0.50) & \cellcolor[HTML]{DCE9F6}0.48 (0.42-0.54) \\
     & \textbf{10} & \cellcolor[HTML]{69ADD5}{\color[HTML]{F1F1F1} 0.67 (0.63-0.72)} & \cellcolor[HTML]{519CCC}{\color[HTML]{F1F1F1} 0.71 (0.67-0.75)} & \cellcolor[HTML]{5DA5D1}{\color[HTML]{F1F1F1} 0.69 (0.65-0.73)} & \cellcolor[HTML]{4090C5}{\color[HTML]{F1F1F1} 0.73 (0.70-0.77)} & \cellcolor[HTML]{7CB7DA}0.65 (0.60-0.70) & \cellcolor[HTML]{4292C6}{\color[HTML]{F1F1F1} 0.73 (0.69-0.77)} \\
     & \textbf{25} & \cellcolor[HTML]{1865AC}{\color[HTML]{F1F1F1} 0.82 (0.79-0.85)} & \cellcolor[HTML]{0D57A1}{\color[HTML]{F1F1F1} 0.85 (0.82-0.87)} & \cellcolor[HTML]{125DA6}{\color[HTML]{F1F1F1} 0.83 (0.80-0.86)} & \cellcolor[HTML]{084B93}{\color[HTML]{F1F1F1} 0.87 (0.84-0.90)} & \cellcolor[HTML]{1C6AB0}{\color[HTML]{F1F1F1} 0.81 (0.78-0.84)} & \cellcolor[HTML]{084B93}{\color[HTML]{F1F1F1} 0.87 (0.85-0.90)} \\
    \multirow{-4}{*}{\textbf{Dense}} & \textbf{50} & \cellcolor[HTML]{084488}{\color[HTML]{F1F1F1} 0.88 (0.86-0.91)} & \cellcolor[HTML]{083A7A}{\color[HTML]{F1F1F1} 0.90 (0.88-0.92)} & \cellcolor[HTML]{084990}{\color[HTML]{F1F1F1} 0.88 (0.85-0.90)} & \cellcolor[HTML]{083674}{\color[HTML]{F1F1F1} 0.91 (0.89-0.93)} & \cellcolor[HTML]{084B93}{\color[HTML]{F1F1F1} 0.87 (0.84-0.89)} & \cellcolor[HTML]{083979}{\color[HTML]{F1F1F1} 0.91 (0.88-0.93)} \\ \midrule
     & \textbf{5} & \cellcolor[HTML]{A3CCE3}0.60 (0.54-0.65) & \cellcolor[HTML]{92C4DE}0.62 (0.57-0.67) & \cellcolor[HTML]{B4D3E9}0.57 (0.52-0.62) & \cellcolor[HTML]{99C7E0}0.61 (0.56-0.66) & \cellcolor[HTML]{B5D4E9}0.57 (0.51-0.62) & \cellcolor[HTML]{8FC2DE}0.62 (0.57-0.67) \\
     & \textbf{10} & \cellcolor[HTML]{3F8FC5}{\color[HTML]{F1F1F1} 0.74 (0.70-0.78)} & \cellcolor[HTML]{2777B8}{\color[HTML]{F1F1F1} 0.78 (0.75-0.82)} & \cellcolor[HTML]{3B8BC2}{\color[HTML]{F1F1F1} 0.74 (0.71-0.78)} & \cellcolor[HTML]{1F6EB3}{\color[HTML]{F1F1F1} 0.80 (0.77-0.83)} & \cellcolor[HTML]{3E8EC4}{\color[HTML]{F1F1F1} 0.74 (0.70-0.78)} & \cellcolor[HTML]{1F6EB3}{\color[HTML]{F1F1F1} 0.80 (0.77-0.83)} \\
     & \textbf{25} & \cellcolor[HTML]{09529D}{\color[HTML]{F1F1F1} 0.86 (0.83-0.88)} & \cellcolor[HTML]{084387}{\color[HTML]{F1F1F1} 0.89 (0.86-0.91)} & \cellcolor[HTML]{0A549E}{\color[HTML]{F1F1F1} 0.85 (0.83-0.88)} & \cellcolor[HTML]{083979}{\color[HTML]{F1F1F1} 0.91 (0.88-0.93)} & \cellcolor[HTML]{0E58A2}{\color[HTML]{F1F1F1} 0.85 (0.82-0.87)} & \cellcolor[HTML]{083E81}{\color[HTML]{F1F1F1} 0.89 (0.87-0.92)} \\
    \multirow{-4}{*}{\textbf{Hybrid}} & \textbf{50} & \cellcolor[HTML]{084285}{\color[HTML]{F1F1F1} 0.89 (0.87-0.91)} & \cellcolor[HTML]{083674}{\color[HTML]{F1F1F1} 0.91 (0.89-0.93)} & \cellcolor[HTML]{084184}{\color[HTML]{F1F1F1} 0.89 (0.87-0.91)} & \cellcolor[HTML]{083370}{\color[HTML]{FFFF00} \textbf{0.92 (0.90-0.93)}} & \cellcolor[HTML]{084D96}{\color[HTML]{F1F1F1} 0.87 (0.84-0.89)} & \cellcolor[HTML]{083674}{\color[HTML]{F1F1F1} 0.91 (0.89-0.93)} \\ \midrule
     & \textbf{5} & \cellcolor[HTML]{97C6DF}0.61 (0.56-0.66) & \cellcolor[HTML]{7FB9DA}0.64 (0.60-0.69) & \cellcolor[HTML]{B0D2E7}0.57 (0.52-0.62) & \cellcolor[HTML]{9DCAE1}0.60 (0.55-0.65) & \cellcolor[HTML]{B8D5EA}0.56 (0.51-0.62) & \cellcolor[HTML]{8ABFDD}0.63 (0.58-0.68) \\
     & \textbf{10} & \cellcolor[HTML]{3484BF}{\color[HTML]{F1F1F1} 0.76 (0.72-0.80)} & \cellcolor[HTML]{2171B5}{\color[HTML]{F1F1F1} 0.79 (0.76-0.83)} & \cellcolor[HTML]{3A8AC2}{\color[HTML]{F1F1F1} 0.75 (0.71-0.78)} & \cellcolor[HTML]{1D6CB1}{\color[HTML]{F1F1F1} 0.80 (0.77-0.84)} & \cellcolor[HTML]{4292C6}{\color[HTML]{F1F1F1} 0.73 (0.69-0.77)} & \cellcolor[HTML]{1F6EB3}{\color[HTML]{F1F1F1} 0.80 (0.77-0.83)} \\
     & \textbf{25} & \cellcolor[HTML]{08488E}{\color[HTML]{F1F1F1} 0.88 (0.85-0.90)} & \cellcolor[HTML]{083979}{\color[HTML]{F1F1F1} 0.91 (0.88-0.93)} & \cellcolor[HTML]{0A539E}{\color[HTML]{F1F1F1} 0.86 (0.83-0.88)} & \cellcolor[HTML]{083776}{\color[HTML]{F1F1F1} 0.91 (0.89-0.93)} & \cellcolor[HTML]{0D57A1}{\color[HTML]{F1F1F1} 0.85 (0.82-0.87)} & \cellcolor[HTML]{083B7C}{\color[HTML]{F1F1F1} 0.90 (0.88-0.92)} \\
    \multirow{-4}{*}{\textbf{Rerank}} & \textbf{50} & \cellcolor[HTML]{083B7C}{\color[HTML]{F1F1F1} 0.90 (0.88-0.92)} & \cellcolor[HTML]{08316D}{\color[HTML]{FFFF00} \textbf{0.92 (0.90-0.94)}} & \cellcolor[HTML]{084285}{\color[HTML]{F1F1F1} 0.89 (0.87-0.91)} & \cellcolor[HTML]{083573}{\color[HTML]{F1F1F1} 0.91 (0.89-0.93)} & \cellcolor[HTML]{084F99}{\color[HTML]{F1F1F1} 0.86 (0.84-0.89)} & \cellcolor[HTML]{083370}{\color[HTML]{FFFF00} \textbf{0.92 (0.90-0.94)}} \\ \bottomrule
    \end{tabular}
    \caption{Gwet's AC1 for \textit{VeriFact} vs. Ground Truth computed across all propositions extracted from LLM-written Brief Hospital Course (BHC) from 100 patients. This table shows results when decompose the BHC as Sentence Propositions and using \textit{VeriFact} to make a prediction in the Binarized 2-Label Prediction Task. The table displays the evaluation results for each combination of the following experimental variables: Top N (number of facts retrieved), Retrieval Method, Reference Context Format, and whether to retrieve facts only from current admission versus the entire EHR. Background cell color intensity is darker blue with higher agreement. The highest agreement is highlighted with yellow text. 95\% confidence intervals shown in parenthesis.}
    \label{tab:suptable4_sub4}
\end{table}

\clearpage
\subsection{Gwet's AC1 for \textit{VeriFact} vs. Ground Truth with Human-written Brief Hospital Course}\label{suptables_sub5}
% Atomic Claim Propositions | Original Label Space
\begin{table}[h]
    \footnotesize
    \begin{tabular}{@{}ccllllll@{}}
    \toprule
    \multicolumn{8}{c}{\textbf{\shortstack{Table B14: Gwet's AC1 for Human-written Brief Hospital Course \\ with Atomic Claim Propositions \& Original 3-Label Prediction Task}}} \\ \midrule
    \multicolumn{1}{l}{} & \textbf{\begin{tabular}[c]{@{}c@{}}Reference Context\\ Format\end{tabular}} & \multicolumn{2}{c}{\textbf{Relevance Score}} & \multicolumn{2}{c}{\textbf{Absolute Time}} & \multicolumn{2}{c}{\textbf{Relative Time}} \\ \cmidrule{2-8}
    \multicolumn{1}{l}{} & \textbf{\begin{tabular}[c]{@{}c@{}}Retrieve Facts \\ Only From \\ Current Admission\end{tabular}} & \multicolumn{1}{c}{\textbf{Yes}} & \multicolumn{1}{c}{\textbf{No}} & \multicolumn{1}{c}{\textbf{Yes}} & \multicolumn{1}{c}{\textbf{No}} & \multicolumn{1}{c}{\textbf{Yes}} & \multicolumn{1}{c}{\textbf{No}} \\ \midrule
    \textbf{\begin{tabular}[c]{@{}c@{}}Retrieval \\ Method\end{tabular}} & \textbf{Top   N} &  &  &  &  &  &  \\ \midrule
     & \textbf{5} & \cellcolor[HTML]{EEF5FC}0.32 (0.30-0.34) & \cellcolor[HTML]{DEEBF7}0.34 (0.32-0.36) & \cellcolor[HTML]{F5F9FE}0.31 (0.29-0.33) & \cellcolor[HTML]{E3EEF8}0.34 (0.32-0.36) & \cellcolor[HTML]{F4F9FE}0.31 (0.29-0.33) & \cellcolor[HTML]{E3EEF8}0.34 (0.32-0.36) \\
     & \textbf{10} & \cellcolor[HTML]{D2E3F3}0.36 (0.34-0.38) & \cellcolor[HTML]{BDD7EC}0.39 (0.37-0.41) & \cellcolor[HTML]{CDDFF1}0.37 (0.35-0.39) & \cellcolor[HTML]{AFD1E7}0.40 (0.38-0.42) & \cellcolor[HTML]{CDE0F1}0.37 (0.35-0.39) & \cellcolor[HTML]{B0D2E7}0.40 (0.38-0.42) \\
     & \textbf{25} & \cellcolor[HTML]{A0CBE2}0.42 (0.40-0.44) & \cellcolor[HTML]{6DAFD7}{\color[HTML]{F1F1F1} 0.46 (0.44-0.48)} & \cellcolor[HTML]{81BADB}0.44 (0.42-0.46) & \cellcolor[HTML]{56A0CE}{\color[HTML]{F1F1F1} 0.48 (0.46-0.50)} & \cellcolor[HTML]{87BDDC}0.44 (0.42-0.46) & \cellcolor[HTML]{57A0CE}{\color[HTML]{F1F1F1} 0.48 (0.46-0.50)} \\
    \multirow{-4}{*}{\textbf{Dense}} & \textbf{50} & \cellcolor[HTML]{61A7D2}{\color[HTML]{F1F1F1} 0.47 (0.45-0.49)} & \cellcolor[HTML]{4191C6}{\color[HTML]{F1F1F1} 0.50 (0.48-0.52)} & \cellcolor[HTML]{5CA4D0}{\color[HTML]{F1F1F1} 0.47 (0.45-0.49)} & \cellcolor[HTML]{3484BF}{\color[HTML]{F1F1F1} 0.51 (0.49-0.53)} & \cellcolor[HTML]{65AAD4}{\color[HTML]{F1F1F1} 0.46 (0.44-0.48)} & \cellcolor[HTML]{3A8AC2}{\color[HTML]{F1F1F1} 0.51 (0.49-0.53)} \\ \midrule
     & \textbf{5} & \cellcolor[HTML]{ECF4FB}0.32 (0.30-0.34) & \cellcolor[HTML]{DAE8F6}0.35 (0.33-0.37) & \cellcolor[HTML]{F3F8FE}0.31 (0.29-0.33) & \cellcolor[HTML]{D9E8F5}0.35 (0.33-0.37) & \cellcolor[HTML]{F6FAFF}0.31 (0.29-0.33) & \cellcolor[HTML]{DEEBF7}0.34 (0.32-0.36) \\
     & \textbf{10} & \cellcolor[HTML]{CFE1F2}0.37 (0.35-0.39) & \cellcolor[HTML]{AFD1E7}0.40 (0.39-0.42) & \cellcolor[HTML]{C4DAEE}0.38 (0.37-0.40) & \cellcolor[HTML]{9AC8E0}0.42 (0.40-0.44) & \cellcolor[HTML]{CCDFF1}0.37 (0.35-0.39) & \cellcolor[HTML]{A1CBE2}0.42 (0.40-0.44) \\
     & \textbf{25} & \cellcolor[HTML]{8ABFDD}0.44 (0.42-0.45) & \cellcolor[HTML]{64A9D3}{\color[HTML]{F1F1F1} 0.47 (0.45-0.48)} & \cellcolor[HTML]{68ACD5}{\color[HTML]{F1F1F1} 0.46 (0.44-0.48)} & \cellcolor[HTML]{3F8FC5}{\color[HTML]{F1F1F1} 0.50 (0.48-0.52)} & \cellcolor[HTML]{6DAFD7}{\color[HTML]{F1F1F1} 0.46 (0.44-0.48)} & \cellcolor[HTML]{4191C6}{\color[HTML]{F1F1F1} 0.50 (0.48-0.52)} \\
    \multirow{-4}{*}{\textbf{Hybrid}} & \textbf{50} & \cellcolor[HTML]{71B1D7}{\color[HTML]{F1F1F1} 0.45 (0.43-0.47)} & \cellcolor[HTML]{4B98CA}{\color[HTML]{F1F1F1} 0.49 (0.47-0.51)} & \cellcolor[HTML]{4997C9}{\color[HTML]{F1F1F1} 0.49 (0.47-0.51)} & \cellcolor[HTML]{2979B9}{\color[HTML]{FFFF00} \textbf{0.53 (0.51-0.54)}} & \cellcolor[HTML]{58A1CF}{\color[HTML]{F1F1F1} 0.48 (0.46-0.50)} & \cellcolor[HTML]{3080BD}{\color[HTML]{F1F1F1} 0.52 (0.50-0.54)} \\ \midrule
     & \textbf{5} & \cellcolor[HTML]{E6F0F9}0.33 (0.31-0.35) & \cellcolor[HTML]{D1E2F3}0.36 (0.34-0.38) & \cellcolor[HTML]{F2F7FD}0.31 (0.29-0.33) & \cellcolor[HTML]{DAE8F6}0.35 (0.33-0.37) & \cellcolor[HTML]{F7FBFF}0.31 (0.29-0.33) & \cellcolor[HTML]{DFEBF7}0.34 (0.32-0.36) \\
     & \textbf{10} & \cellcolor[HTML]{BED8EC}0.39 (0.37-0.41) & \cellcolor[HTML]{97C6DF}0.43 (0.41-0.44) & \cellcolor[HTML]{C4DAEE}0.38 (0.36-0.40) & \cellcolor[HTML]{99C7E0}0.42 (0.40-0.44) & \cellcolor[HTML]{CADDF0}0.38 (0.36-0.40) & \cellcolor[HTML]{A0CBE2}0.42 (0.40-0.44) \\
     & \textbf{25} & \cellcolor[HTML]{72B2D8}{\color[HTML]{F1F1F1} 0.45 (0.43-0.47)} & \cellcolor[HTML]{4D99CA}{\color[HTML]{F1F1F1} 0.49 (0.47-0.50)} & \cellcolor[HTML]{68ACD5}{\color[HTML]{F1F1F1} 0.46 (0.44-0.48)} & \cellcolor[HTML]{4191C6}{\color[HTML]{F1F1F1} 0.50 (0.48-0.52)} & \cellcolor[HTML]{6FB0D7}{\color[HTML]{F1F1F1} 0.46 (0.44-0.47)} & \cellcolor[HTML]{4896C8}{\color[HTML]{F1F1F1} 0.49 (0.47-0.51)} \\
    \multirow{-4}{*}{\textbf{Rerank}} & \textbf{50} & \cellcolor[HTML]{549FCD}{\color[HTML]{F1F1F1} 0.48 (0.46-0.50)} & \cellcolor[HTML]{3686C0}{\color[HTML]{F1F1F1} 0.51 (0.49-0.53)} & \cellcolor[HTML]{4A98C9}{\color[HTML]{F1F1F1} 0.49 (0.47-0.51)} & \cellcolor[HTML]{2A7AB9}{\color[HTML]{FFFF00} \textbf{0.53 (0.51-0.54)}} & \cellcolor[HTML]{549FCD}{\color[HTML]{F1F1F1} 0.48 (0.46-0.50)} & \cellcolor[HTML]{3181BD}{\color[HTML]{F1F1F1} 0.52 (0.50-0.54)} \\ \bottomrule
    \end{tabular}
    \caption{Gwet's AC1 for \textit{VeriFact} vs. Ground Truth computed across all propositions extracted from Human-written Brief Hospital Course (BHC) from 100 patients. This table shows results when decompose the BHC as Atomic Claim Propositions and using \textit{VeriFact} to make a prediction in the Original 3-Label Prediction Task. The table displays the evaluation results for each combination of the following experimental variables: Top N (number of facts retrieved), Retrieval Method, Reference Context Format, and whether to retrieve facts only from current admission versus the entire EHR. Background cell color intensity is darker blue with higher agreement. The highest agreement is highlighted with yellow text. 95\% confidence intervals shown in parenthesis.}
    \label{tab:suptable5_sub1}
\end{table}

% Sentence Propositions | Original Label Space
\begin{table}[h]
    \footnotesize
    \begin{tabular}{@{}ccllllll@{}}
    \toprule
    \multicolumn{8}{c}{\textbf{\shortstack{Table B15: Gwet's AC1 for Human-written Brief Hospital Course \\ with Sentence Propositions \& Original 3-Label Prediction Task}}} \\ \midrule
    \multicolumn{1}{l}{} & \textbf{\begin{tabular}[c]{@{}c@{}}Reference Context\\ Format\end{tabular}} & \multicolumn{2}{c}{\textbf{Relevance Score}} & \multicolumn{2}{c}{\textbf{Absolute Time}} & \multicolumn{2}{c}{\textbf{Relative Time}} \\ \cmidrule{2-8}
    \multicolumn{1}{l}{} & \textbf{\begin{tabular}[c]{@{}c@{}}Retrieve Facts \\ Only From \\ Current Admission\end{tabular}} & \multicolumn{1}{c}{\textbf{Yes}} & \multicolumn{1}{c}{\textbf{No}} & \multicolumn{1}{c}{\textbf{Yes}} & \multicolumn{1}{c}{\textbf{No}} & \multicolumn{1}{c}{\textbf{Yes}} & \multicolumn{1}{c}{\textbf{No}} \\ \midrule
    \textbf{\begin{tabular}[c]{@{}c@{}}Retrieval \\ Method\end{tabular}} & \textbf{Top   N} &  &  &  &  &  &  \\ \midrule
     & \textbf{5} & \cellcolor[HTML]{D0E1F2}0.37 (0.34-0.39) & \cellcolor[HTML]{CEE0F2}0.37 (0.34-0.40) & \cellcolor[HTML]{E0ECF8}0.34 (0.32-0.37) & \cellcolor[HTML]{DCEAF6}0.35 (0.32-0.37) & \cellcolor[HTML]{E3EEF9}0.34 (0.31-0.36) & \cellcolor[HTML]{DAE8F6}0.35 (0.32-0.38) \\
     & \textbf{10} & \cellcolor[HTML]{AACFE5}0.41 (0.38-0.44) & \cellcolor[HTML]{94C4DF}0.43 (0.40-0.45) & \cellcolor[HTML]{B5D4E9}0.40 (0.37-0.42) & \cellcolor[HTML]{A8CEE4}0.41 (0.38-0.44) & \cellcolor[HTML]{BAD6EB}0.39 (0.37-0.42) & \cellcolor[HTML]{A6CEE4}0.41 (0.39-0.44) \\
     & \textbf{25} & \cellcolor[HTML]{56A0CE}{\color[HTML]{F1F1F1} 0.48 (0.45-0.50)} & \cellcolor[HTML]{4594C7}{\color[HTML]{F1F1F1} 0.49 (0.47-0.52)} & \cellcolor[HTML]{6AAED6}{\color[HTML]{F1F1F1} 0.46 (0.43-0.49)} & \cellcolor[HTML]{4D99CA}{\color[HTML]{F1F1F1} 0.49 (0.46-0.51)} & \cellcolor[HTML]{6AAED6}{\color[HTML]{F1F1F1} 0.46 (0.43-0.48)} & \cellcolor[HTML]{539ECD}{\color[HTML]{F1F1F1} 0.48 (0.45-0.51)} \\
    \multirow{-4}{*}{\textbf{Dense}} & \textbf{50} & \cellcolor[HTML]{3686C0}{\color[HTML]{F1F1F1} 0.51 (0.49-0.53)} & \cellcolor[HTML]{2F7FBC}{\color[HTML]{F1F1F1} 0.52 (0.49-0.54)} & \cellcolor[HTML]{4A98C9}{\color[HTML]{F1F1F1} 0.49 (0.46-0.51)} & \cellcolor[HTML]{4191C6}{\color[HTML]{F1F1F1} 0.50 (0.47-0.52)} & \cellcolor[HTML]{4B98CA}{\color[HTML]{F1F1F1} 0.49 (0.46-0.51)} & \cellcolor[HTML]{3383BE}{\color[HTML]{F1F1F1} 0.51 (0.49-0.54)} \\ \midrule
     & \textbf{5} & \cellcolor[HTML]{99C7E0}0.42 (0.40-0.45) & \cellcolor[HTML]{9DCAE1}0.42 (0.40-0.45) & \cellcolor[HTML]{B0D2E7}0.40 (0.38-0.43) & \cellcolor[HTML]{AFD1E7}0.40 (0.38-0.43) & \cellcolor[HTML]{AFD1E7}0.40 (0.38-0.43) & \cellcolor[HTML]{A9CFE5}0.41 (0.38-0.44) \\
     & \textbf{10} & \cellcolor[HTML]{5CA4D0}{\color[HTML]{F1F1F1} 0.47 (0.45-0.50)} & \cellcolor[HTML]{4E9ACB}{\color[HTML]{F1F1F1} 0.49 (0.46-0.51)} & \cellcolor[HTML]{79B5D9}0.45 (0.42-0.47) & \cellcolor[HTML]{65AAD4}{\color[HTML]{F1F1F1} 0.46 (0.44-0.49)} & \cellcolor[HTML]{74B3D8}0.45 (0.43-0.48) & \cellcolor[HTML]{66ABD4}{\color[HTML]{F1F1F1} 0.46 (0.44-0.49)} \\
     & \textbf{25} & \cellcolor[HTML]{3A8AC2}{\color[HTML]{F1F1F1} 0.51 (0.48-0.53)} & \cellcolor[HTML]{2B7BBA}{\color[HTML]{F1F1F1} 0.52 (0.50-0.55)} & \cellcolor[HTML]{529DCC}{\color[HTML]{F1F1F1} 0.48 (0.45-0.51)} & \cellcolor[HTML]{3C8CC3}{\color[HTML]{F1F1F1} 0.50 (0.48-0.53)} & \cellcolor[HTML]{519CCC}{\color[HTML]{F1F1F1} 0.48 (0.46-0.51)} & \cellcolor[HTML]{3B8BC2}{\color[HTML]{F1F1F1} 0.50 (0.48-0.53)} \\
    \multirow{-4}{*}{\textbf{Hybrid}} & \textbf{50} & \cellcolor[HTML]{3686C0}{\color[HTML]{F1F1F1} 0.51 (0.49-0.54)} & \cellcolor[HTML]{2979B9}{\color[HTML]{FFFF00} \textbf{0.53 (0.50-0.55)}} & \cellcolor[HTML]{519CCC}{\color[HTML]{F1F1F1} 0.48 (0.46-0.51)} & \cellcolor[HTML]{3D8DC4}{\color[HTML]{F1F1F1} 0.50 (0.48-0.53)} & \cellcolor[HTML]{4695C8}{\color[HTML]{F1F1F1} 0.49 (0.47-0.52)} & \cellcolor[HTML]{3686C0}{\color[HTML]{F1F1F1} 0.51 (0.48-0.53)} \\ \midrule
     & \textbf{5} & \cellcolor[HTML]{8DC1DD}0.43 (0.41-0.46) & \cellcolor[HTML]{92C4DE}0.43 (0.41-0.46) & \cellcolor[HTML]{B2D2E8}0.40 (0.38-0.43) & \cellcolor[HTML]{AFD1E7}0.40 (0.38-0.43) & \cellcolor[HTML]{B0D2E7}0.40 (0.38-0.43) & \cellcolor[HTML]{AFD1E7}0.40 (0.38-0.43) \\
     & \textbf{10} & \cellcolor[HTML]{539ECD}{\color[HTML]{F1F1F1} 0.48 (0.46-0.51)} & \cellcolor[HTML]{549FCD}{\color[HTML]{F1F1F1} 0.48 (0.45-0.51)} & \cellcolor[HTML]{75B4D8}0.45 (0.42-0.48) & \cellcolor[HTML]{68ACD5}{\color[HTML]{F1F1F1} 0.46 (0.44-0.49)} & \cellcolor[HTML]{71B1D7}{\color[HTML]{F1F1F1} 0.45 (0.43-0.48)} & \cellcolor[HTML]{68ACD5}{\color[HTML]{F1F1F1} 0.46 (0.44-0.49)} \\
     & \textbf{25} & \cellcolor[HTML]{3989C1}{\color[HTML]{F1F1F1} 0.51 (0.48-0.53)} & \cellcolor[HTML]{2777B8}{\color[HTML]{FFFF00} \textbf{0.53 (0.50-0.55)}} & \cellcolor[HTML]{56A0CE}{\color[HTML]{F1F1F1} 0.48 (0.45-0.50)} & \cellcolor[HTML]{3C8CC3}{\color[HTML]{F1F1F1} 0.50 (0.48-0.53)} & \cellcolor[HTML]{529DCC}{\color[HTML]{F1F1F1} 0.48 (0.46-0.51)} & \cellcolor[HTML]{3F8FC5}{\color[HTML]{F1F1F1} 0.50 (0.47-0.53)} \\
    \multirow{-4}{*}{\textbf{Rerank}} & \textbf{50} & \cellcolor[HTML]{3383BE}{\color[HTML]{F1F1F1} 0.51 (0.49-0.54)} & \cellcolor[HTML]{2171B5}{\color[HTML]{FFFF00} \textbf{0.53 (0.51-0.56)}} & \cellcolor[HTML]{4B98CA}{\color[HTML]{F1F1F1} 0.49 (0.46-0.51)} & \cellcolor[HTML]{3E8EC4}{\color[HTML]{F1F1F1} 0.50 (0.47-0.53)} & \cellcolor[HTML]{4695C8}{\color[HTML]{F1F1F1} 0.49 (0.47-0.52)} & \cellcolor[HTML]{3888C1}{\color[HTML]{F1F1F1} 0.51 (0.48-0.53)} \\ \bottomrule
    \end{tabular}
    \caption{Gwet's AC1 for \textit{VeriFact} vs. Ground Truth computed across all propositions extracted from Human-written Brief Hospital Course (BHC) from 100 patients. This table shows results when decompose the BHC as Sentence Propositions and using \textit{VeriFact} to make a prediction in the Original 3-Label Prediction Task. The table displays the evaluation results for each combination of the following experimental variables: Top N (number of facts retrieved), Retrieval Method, Reference Context Format, and whether to retrieve facts only from current admission versus the entire EHR. Background cell color intensity is darker blue with higher agreement. The highest agreement is highlighted with yellow text. 95\% confidence intervals shown in parenthesis.}
    \label{tab:suptable5_sub2}
\end{table}

% Atomic Claim Propositions | Binarized Label Space
\begin{table}[h]
    \footnotesize
    \begin{tabular}{@{}ccllllll@{}}
    \toprule
    \multicolumn{8}{c}{\textbf{\shortstack{Table B16: Gwet's AC1 for Human-written Brief Hospital Course \\ with Atomic Claim Propositions \& Binarized 2-Label Prediction Task}}} \\ \midrule
    \multicolumn{1}{l}{} & \textbf{\begin{tabular}[c]{@{}c@{}}Reference Context\\ Format\end{tabular}} & \multicolumn{2}{c}{\textbf{Relevance Score}} & \multicolumn{2}{c}{\textbf{Absolute Time}} & \multicolumn{2}{c}{\textbf{Relative Time}} \\ \cmidrule{2-8}
    \multicolumn{1}{l}{} & \textbf{\begin{tabular}[c]{@{}c@{}}Retrieve Facts \\ Only From \\ Current Admission\end{tabular}} & \multicolumn{1}{c}{\textbf{Yes}} & \multicolumn{1}{c}{\textbf{No}} & \multicolumn{1}{c}{\textbf{Yes}} & \multicolumn{1}{c}{\textbf{No}} & \multicolumn{1}{c}{\textbf{Yes}} & \multicolumn{1}{c}{\textbf{No}} \\ \midrule
    \textbf{\begin{tabular}[c]{@{}c@{}}Retrieval \\ Method\end{tabular}} & \textbf{Top   N} &  &  &  &  &  &  \\ \midrule
     & \textbf{5} & \cellcolor[HTML]{D0E1F2}0.37   (0.34-0.39) & \cellcolor[HTML]{B3D3E8}0.40 (0.38-0.42) & \cellcolor[HTML]{D3E3F3}0.36 (0.34-0.39) & \cellcolor[HTML]{B4D3E9}0.40 (0.38-0.42) & \cellcolor[HTML]{D3E4F3}0.36 (0.34-0.38) & \cellcolor[HTML]{BCD7EB}0.39 (0.37-0.42) \\
     & \textbf{10} & \cellcolor[HTML]{A8CEE4}0.41 (0.39-0.44) & \cellcolor[HTML]{72B2D8}{\color[HTML]{F1F1F1} 0.45 (0.43-0.48)} & \cellcolor[HTML]{8CC0DD}0.43 (0.41-0.46) & \cellcolor[HTML]{56A0CE}{\color[HTML]{F1F1F1} 0.48 (0.45-0.50)} & \cellcolor[HTML]{94C4DF}0.43 (0.40-0.45) & \cellcolor[HTML]{57A0CE}{\color[HTML]{F1F1F1} 0.48 (0.45-0.50)} \\
     & \textbf{25} & \cellcolor[HTML]{5DA5D1}{\color[HTML]{F1F1F1} 0.47 (0.45-0.49)} & \cellcolor[HTML]{2F7FBC}{\color[HTML]{F1F1F1} 0.52 (0.50-0.54)} & \cellcolor[HTML]{3181BD}{\color[HTML]{F1F1F1} 0.52 (0.49-0.54)} & \cellcolor[HTML]{0F5AA3}{\color[HTML]{F1F1F1} 0.56 (0.54-0.58)} & \cellcolor[HTML]{3888C1}{\color[HTML]{F1F1F1} 0.51 (0.49-0.53)} & \cellcolor[HTML]{125DA6}{\color[HTML]{F1F1F1} 0.56 (0.54-0.58)} \\
    \multirow{-4}{*}{\textbf{Dense}} & \textbf{50} & \cellcolor[HTML]{206FB4}{\color[HTML]{F1F1F1} 0.54 (0.51-0.56)} & \cellcolor[HTML]{08519C}{\color[HTML]{F1F1F1} 0.57 (0.55-0.59)} & \cellcolor[HTML]{1562A9}{\color[HTML]{F1F1F1} 0.55 (0.53-0.58)} & \cellcolor[HTML]{08306B}{\color[HTML]{FFFF00} \textbf{0.61 (0.59-0.63)}} & \cellcolor[HTML]{206FB4}{\color[HTML]{F1F1F1} 0.54 (0.52-0.56)} & \cellcolor[HTML]{084082}{\color[HTML]{F1F1F1} 0.59 (0.57-0.61)} \\ \midrule
     & \textbf{5} & \cellcolor[HTML]{C6DBEF}0.38 (0.36-0.41) & \cellcolor[HTML]{A4CCE3}0.41 (0.39-0.44) & \cellcolor[HTML]{BFD8ED}0.39 (0.36-0.41) & \cellcolor[HTML]{8CC0DD}0.43 (0.41-0.46) & \cellcolor[HTML]{CCDFF1}0.37 (0.35-0.40) & \cellcolor[HTML]{9CC9E1}0.42 (0.40-0.44) \\
     & \textbf{10} & \cellcolor[HTML]{8CC0DD}0.43 (0.41-0.46) & \cellcolor[HTML]{529DCC}{\color[HTML]{F1F1F1} 0.48 (0.46-0.50)} & \cellcolor[HTML]{69ADD5}{\color[HTML]{F1F1F1} 0.46 (0.44-0.48)} & \cellcolor[HTML]{3484BF}{\color[HTML]{F1F1F1} 0.51 (0.49-0.54)} & \cellcolor[HTML]{7AB6D9}0.45 (0.42-0.47) & \cellcolor[HTML]{3F8FC5}{\color[HTML]{F1F1F1} 0.50 (0.48-0.52)} \\
     & \textbf{25} & \cellcolor[HTML]{3C8CC3}{\color[HTML]{F1F1F1} 0.50 (0.48-0.53)} & \cellcolor[HTML]{1C6BB0}{\color[HTML]{F1F1F1} 0.54 (0.52-0.56)} & \cellcolor[HTML]{1966AD}{\color[HTML]{F1F1F1} 0.55 (0.53-0.57)} & \cellcolor[HTML]{083A7A}{\color[HTML]{F1F1F1} 0.60 (0.58-0.62)} & \cellcolor[HTML]{1D6CB1}{\color[HTML]{F1F1F1} 0.54 (0.52-0.56)} & \cellcolor[HTML]{084488}{\color[HTML]{F1F1F1} 0.59 (0.57-0.61)} \\
    \multirow{-4}{*}{\textbf{Hybrid}} & \textbf{50} & \cellcolor[HTML]{2474B7}{\color[HTML]{F1F1F1} 0.53 (0.51-0.55)} & \cellcolor[HTML]{084B93}{\color[HTML]{F1F1F1} 0.58 (0.56-0.60)} & \cellcolor[HTML]{0F5AA3}{\color[HTML]{F1F1F1} 0.56 (0.54-0.58)} & \cellcolor[HTML]{08306B}{\color[HTML]{FFFF00} \textbf{0.61 (0.59-0.63)}} & \cellcolor[HTML]{125DA6}{\color[HTML]{F1F1F1} 0.56 (0.54-0.58)} & \cellcolor[HTML]{08306B}{\color[HTML]{FFFF00} \textbf{0.61 (0.59-0.63)}} \\ \midrule
     & \textbf{5} & \cellcolor[HTML]{B4D3E9}0.40 (0.38-0.42) & \cellcolor[HTML]{87BDDC}0.44 (0.42-0.46) & \cellcolor[HTML]{BFD8ED}0.39 (0.36-0.41) & \cellcolor[HTML]{8CC0DD}0.43 (0.41-0.46) & \cellcolor[HTML]{CBDEF1}0.37 (0.35-0.40) & \cellcolor[HTML]{A0CBE2}0.42 (0.39-0.44) \\
     & \textbf{10} & \cellcolor[HTML]{60A7D2}{\color[HTML]{F1F1F1} 0.47 (0.45-0.49)} & \cellcolor[HTML]{3484BF}{\color[HTML]{F1F1F1} 0.51 (0.49-0.54)} & \cellcolor[HTML]{69ADD5}{\color[HTML]{F1F1F1} 0.46 (0.44-0.48)} & \cellcolor[HTML]{3080BD}{\color[HTML]{F1F1F1} 0.52 (0.50-0.54)} & \cellcolor[HTML]{71B1D7}{\color[HTML]{F1F1F1} 0.45 (0.43-0.48)} & \cellcolor[HTML]{3D8DC4}{\color[HTML]{F1F1F1} 0.50 (0.48-0.53)} \\
     & \textbf{25} & \cellcolor[HTML]{2979B9}{\color[HTML]{F1F1F1} 0.53 (0.50-0.55)} & \cellcolor[HTML]{08509B}{\color[HTML]{F1F1F1} 0.57 (0.55-0.60)} & \cellcolor[HTML]{1562A9}{\color[HTML]{F1F1F1} 0.55 (0.53-0.57)} & \cellcolor[HTML]{083A7A}{\color[HTML]{F1F1F1} 0.60 (0.58-0.62)} & \cellcolor[HTML]{206FB4}{\color[HTML]{F1F1F1} 0.54 (0.51-0.56)} & \cellcolor[HTML]{08468B}{\color[HTML]{F1F1F1} 0.59 (0.56-0.61)} \\
    \multirow{-4}{*}{\textbf{Rerank}} & \textbf{50} & \cellcolor[HTML]{0A549E}{\color[HTML]{F1F1F1} 0.57 (0.55-0.59)} & \cellcolor[HTML]{083877}{\color[HTML]{F1F1F1} 0.60 (0.58-0.62)} & \cellcolor[HTML]{0C56A0}{\color[HTML]{F1F1F1} 0.57 (0.55-0.59)} & \cellcolor[HTML]{08306B}{\color[HTML]{FFFF00} \textbf{0.61 (0.59-0.63)}} & \cellcolor[HTML]{0E59A2}{\color[HTML]{F1F1F1} 0.56 (0.54-0.58)} & \cellcolor[HTML]{083471}{\color[HTML]{FFFF00} \textbf{0.61 (0.59-0.63)}} \\ \bottomrule
    \end{tabular}
    \caption{Gwet's AC1 for \textit{VeriFact} vs. Ground Truth computed across all propositions extracted from Human-written Brief Hospital Course (BHC) from 100 patients. This table shows results when decompose the BHC as Atomic Claim Propositions and using \textit{VeriFact} to make a prediction in the Binarized 2-Label Prediction Task. The table displays the evaluation results for each combination of the following experimental variables: Top N (number of facts retrieved), Retrieval Method, Reference Context Format, and whether to retrieve facts only from current admission versus the entire EHR. Background cell color intensity is darker blue with higher agreement. The highest agreement is highlighted with yellow text. 95\% confidence intervals shown in parenthesis.}
    \label{tab:suptable5_sub3}
\end{table}

% Sentence Propositions | Binarized Label Space
\begin{table}[h]
    \footnotesize
    \begin{tabular}{@{}ccllllll@{}}
    \toprule
    \multicolumn{8}{c}{\textbf{\shortstack{Table B17: Gwet's AC1 for Human-written Brief Hospital Course \\ with Sentence Propositions \& Binarized 2-Label Prediction Task}}} \\ \midrule
    \multicolumn{1}{l}{} & \textbf{\begin{tabular}[c]{@{}c@{}}Reference Context\\ Format\end{tabular}} & \multicolumn{2}{c}{\textbf{Relevance Score}} & \multicolumn{2}{c}{\textbf{Absolute Time}} & \multicolumn{2}{c}{\textbf{Relative Time}} \\ \cmidrule{2-8}
    \multicolumn{1}{l}{} & \textbf{\begin{tabular}[c]{@{}c@{}}Retrieve Facts \\ Only From \\ Current Admission\end{tabular}} & \multicolumn{1}{c}{\textbf{Yes}} & \multicolumn{1}{c}{\textbf{No}} & \multicolumn{1}{c}{\textbf{Yes}} & \multicolumn{1}{c}{\textbf{No}} & \multicolumn{1}{c}{\textbf{Yes}} & \multicolumn{1}{c}{\textbf{No}} \\ \midrule
    \textbf{\begin{tabular}[c]{@{}c@{}}Retrieval \\ Method\end{tabular}} & \textbf{Top   N} &  &  &  &  &  &  \\ \midrule
     & \textbf{5} & \cellcolor[HTML]{58A1CF}{\color[HTML]{F1F1F1} 0.48 (0.45-0.51)} & \cellcolor[HTML]{4B98CA}{\color[HTML]{F1F1F1} 0.49 (0.46-0.52)} & \cellcolor[HTML]{5AA2CF}{\color[HTML]{F1F1F1} 0.47 (0.45-0.50)} & \cellcolor[HTML]{58A1CF}{\color[HTML]{F1F1F1} 0.48 (0.45-0.51)} & \cellcolor[HTML]{5CA4D0}{\color[HTML]{F1F1F1} 0.47 (0.44-0.50)} & \cellcolor[HTML]{57A0CE}{\color[HTML]{F1F1F1} 0.48 (0.45-0.51)} \\
     & \textbf{10} & \cellcolor[HTML]{3E8EC4}{\color[HTML]{F1F1F1} 0.50 (0.47-0.53)} & \cellcolor[HTML]{2C7CBA}{\color[HTML]{F1F1F1} 0.52 (0.49-0.55)} & \cellcolor[HTML]{3888C1}{\color[HTML]{F1F1F1} 0.51 (0.48-0.54)} & \cellcolor[HTML]{2D7DBB}{\color[HTML]{F1F1F1} 0.52 (0.49-0.55)} & \cellcolor[HTML]{4292C6}{\color[HTML]{F1F1F1} 0.50 (0.47-0.53)} & \cellcolor[HTML]{3181BD}{\color[HTML]{F1F1F1} 0.52 (0.49-0.55)} \\
     & \textbf{25} & \cellcolor[HTML]{2070B4}{\color[HTML]{F1F1F1} 0.54 (0.51-0.57)} & \cellcolor[HTML]{125DA6}{\color[HTML]{F1F1F1} 0.56 (0.53-0.59)} & \cellcolor[HTML]{2070B4}{\color[HTML]{F1F1F1} 0.54 (0.51-0.57)} & \cellcolor[HTML]{1460A8}{\color[HTML]{F1F1F1} 0.55 (0.53-0.59)} & \cellcolor[HTML]{2979B9}{\color[HTML]{F1F1F1} 0.53 (0.50-0.56)} & \cellcolor[HTML]{2474B7}{\color[HTML]{F1F1F1} 0.53 (0.50-0.56)} \\
    \multirow{-4}{*}{\textbf{Dense}} & \textbf{50} & \cellcolor[HTML]{1764AB}{\color[HTML]{F1F1F1} 0.55 (0.52-0.58)} & \cellcolor[HTML]{0E58A2}{\color[HTML]{F1F1F1} 0.56 (0.54-0.59)} & \cellcolor[HTML]{1A68AE}{\color[HTML]{F1F1F1} 0.54 (0.51-0.57)} & \cellcolor[HTML]{2070B4}{\color[HTML]{F1F1F1} 0.54 (0.50-0.57)} & \cellcolor[HTML]{2373B6}{\color[HTML]{F1F1F1} 0.53 (0.50-0.56)} & \cellcolor[HTML]{2171B5}{\color[HTML]{F1F1F1} 0.54 (0.50-0.56)} \\ \midrule
     & \textbf{5} & \cellcolor[HTML]{2171B5}{\color[HTML]{F1F1F1} 0.53 (0.51-0.56)} & \cellcolor[HTML]{1E6DB2}{\color[HTML]{F1F1F1} 0.54 (0.51-0.57)} & \cellcolor[HTML]{2474B7}{\color[HTML]{F1F1F1} 0.53 (0.50-0.56)} & \cellcolor[HTML]{2474B7}{\color[HTML]{F1F1F1} 0.53 (0.50-0.56)} & \cellcolor[HTML]{2474B7}{\color[HTML]{F1F1F1} 0.53 (0.50-0.56)} & \cellcolor[HTML]{206FB4}{\color[HTML]{F1F1F1} 0.54 (0.51-0.57)} \\
     & \textbf{10} & \cellcolor[HTML]{1562A9}{\color[HTML]{F1F1F1} 0.55 (0.53-0.58)} & \cellcolor[HTML]{0A539E}{\color[HTML]{F1F1F1} 0.57 (0.54-0.60)} & \cellcolor[HTML]{1460A8}{\color[HTML]{F1F1F1} 0.56 (0.53-0.58)} & \cellcolor[HTML]{0E58A2}{\color[HTML]{F1F1F1} 0.56 (0.54-0.59)} & \cellcolor[HTML]{135FA7}{\color[HTML]{F1F1F1} 0.56 (0.53-0.58)} & \cellcolor[HTML]{0F5AA3}{\color[HTML]{F1F1F1} 0.56 (0.53-0.59)} \\
     & \textbf{25} & \cellcolor[HTML]{105BA4}{\color[HTML]{F1F1F1} 0.56 (0.53-0.59)} & \cellcolor[HTML]{084F99}{\color[HTML]{F1F1F1} 0.57 (0.55-0.60)} & \cellcolor[HTML]{1460A8}{\color[HTML]{F1F1F1} 0.55 (0.53-0.59)} & \cellcolor[HTML]{084E98}{\color[HTML]{FFFF00} \textbf{0.58 (0.55-0.61)}} & \cellcolor[HTML]{115CA5}{\color[HTML]{F1F1F1} 0.56 (0.53-0.59)} & \cellcolor[HTML]{115CA5}{\color[HTML]{F1F1F1} 0.56 (0.53-0.59)} \\
    \multirow{-4}{*}{\textbf{Hybrid}} & \textbf{50} & \cellcolor[HTML]{1967AD}{\color[HTML]{F1F1F1} 0.55 (0.52-0.57)} & \cellcolor[HTML]{0A549E}{\color[HTML]{F1F1F1} 0.57 (0.54-0.60)} & \cellcolor[HTML]{1E6DB2}{\color[HTML]{F1F1F1} 0.54 (0.51-0.57)} & \cellcolor[HTML]{1967AD}{\color[HTML]{F1F1F1} 0.55 (0.52-0.58)} & \cellcolor[HTML]{1F6EB3}{\color[HTML]{F1F1F1} 0.54 (0.51-0.57)} & \cellcolor[HTML]{1F6EB3}{\color[HTML]{F1F1F1} 0.54 (0.51-0.57)} \\
     & \textbf{5} & \cellcolor[HTML]{1B69AF}{\color[HTML]{F1F1F1} 0.54 (0.52-0.57)} & \cellcolor[HTML]{1865AC}{\color[HTML]{F1F1F1} 0.55 (0.52-0.58)} & \cellcolor[HTML]{2575B7}{\color[HTML]{F1F1F1} 0.53 (0.50-0.56)} & \cellcolor[HTML]{2474B7}{\color[HTML]{F1F1F1} 0.53 (0.50-0.56)} & \cellcolor[HTML]{2777B8}{\color[HTML]{F1F1F1} 0.53 (0.50-0.56)} & \cellcolor[HTML]{2171B5}{\color[HTML]{F1F1F1} 0.53 (0.50-0.57)} \\
     & \textbf{10} & \cellcolor[HTML]{135FA7}{\color[HTML]{F1F1F1} 0.56 (0.53-0.59)} & \cellcolor[HTML]{115CA5}{\color[HTML]{F1F1F1} 0.56 (0.53-0.59)} & \cellcolor[HTML]{1562A9}{\color[HTML]{F1F1F1} 0.55 (0.53-0.58)} & \cellcolor[HTML]{105BA4}{\color[HTML]{F1F1F1} 0.56 (0.53-0.59)} & \cellcolor[HTML]{1460A8}{\color[HTML]{F1F1F1} 0.56 (0.53-0.58)} & \cellcolor[HTML]{125DA6}{\color[HTML]{F1F1F1} 0.56 (0.53-0.59)} \\
     & \textbf{25} & \cellcolor[HTML]{1663AA}{\color[HTML]{F1F1F1} 0.55 (0.52-0.58)} & \cellcolor[HTML]{09529D}{\color[HTML]{F1F1F1} 0.57 (0.54-0.60)} & \cellcolor[HTML]{1562A9}{\color[HTML]{F1F1F1} 0.55 (0.52-0.58)} & \cellcolor[HTML]{09529D}{\color[HTML]{F1F1F1} 0.57 (0.54-0.60)} & \cellcolor[HTML]{1561A9}{\color[HTML]{F1F1F1} 0.55 (0.53-0.58)} & \cellcolor[HTML]{0C56A0}{\color[HTML]{F1F1F1} 0.57 (0.54-0.60)} \\
    \multirow{-4}{*}{\textbf{Rerank}} & \textbf{50} & \cellcolor[HTML]{1C6BB0}{\color[HTML]{F1F1F1} 0.54 (0.51-0.57)} & \cellcolor[HTML]{0C56A0}{\color[HTML]{F1F1F1} 0.57 (0.54-0.60)} & \cellcolor[HTML]{1C6AB0}{\color[HTML]{F1F1F1} 0.54 (0.51-0.57)} & \cellcolor[HTML]{1C6AB0}{\color[HTML]{F1F1F1} 0.54 (0.51-0.57)} & \cellcolor[HTML]{1C6AB0}{\color[HTML]{F1F1F1} 0.54 (0.51-0.57)} & \cellcolor[HTML]{1F6EB3}{\color[HTML]{F1F1F1} 0.54 (0.51-0.57)} \\ \bottomrule
    \end{tabular}
    \caption{Gwet's AC1 for \textit{VeriFact} vs. Ground Truth computed across all propositions extracted from Human-written Brief Hospital Course (BHC) from 100 patients. This table shows results when decompose the BHC as Sentence Propositions and using \textit{VeriFact} to make a prediction in the Binarized 2-Label Prediction Task. The table displays the evaluation results for each combination of the following experimental variables: Top N (number of facts retrieved), Retrieval Method, Reference Context Format, and whether to retrieve facts only from current admission versus the entire EHR. Background cell color intensity is darker blue with higher agreement. The highest agreement is highlighted with yellow text. 95\% confidence intervals shown in parenthesis.}
    \label{tab:suptable5_sub4}
\end{table}

\end{landscape}

\end{appendices}

%%============%%
%% References %%
%%============%%

\bibliography{paperpile}% common bib file

\begin{thebibliography}{10}
\providecommand{\doi}[1]{\url{https://doi.org/#1}}
\bibcommenthead

\bibitem[\protect\citeauthoryear{Van~Veen et~al.}{2024}]{Van_Veen2024-tm}
Van~Veen D, Van~Uden C, Blankemeier L, Delbrouck JB, Aali A, Bluethgen C, et~al.
\newblock Adapted large language models can outperform medical experts in clinical text summarization.
\newblock Nat Med. 2024 Feb;.

\bibitem[\protect\citeauthoryear{Tang et~al.}{2023}]{Tang2023-jn}
Tang L, Sun Z, Idnay B, Nestor JG, Soroush A, Elias PA, et~al.
\newblock Evaluating large language models on medical evidence summarization.
\newblock NPJ Digit Med. 2023 Aug;6(1):158.

\bibitem[\protect\citeauthoryear{Chung et~al.}{2024}]{Chung2024-dk}
Chung P, Fong CT, Walters AM, Aghaeepour N, Yetisgen M, O'Reilly-Shah VN.
\newblock Large language model capabilities in perioperative risk prediction and prognostication.
\newblock JAMA Surg. 2024 Aug;159(8):928--937.

\bibitem[\protect\citeauthoryear{Savage et~al.}{2024}]{Savage2024-yh}
Savage T, Nayak A, Gallo R, Rangan E, Chen JH.
\newblock Diagnostic reasoning prompts reveal the potential for large language model interpretability in medicine.
\newblock NPJ Digit Med. 2024 Jan;7(1):20.

\bibitem[\protect\citeauthoryear{Goh et~al.}{2024}]{Goh2024-cc}
Goh E, Gallo R, Hom J, Strong E, Weng Y, Kerman H, et~al.
\newblock Large language model influence on diagnostic reasoning: A randomized clinical trial: A randomized clinical trial.
\newblock JAMA Netw Open. 2024 Oct;7(10):e2440969.

\bibitem[\protect\citeauthoryear{Zaretsky et~al.}{2024}]{Zaretsky2024-oj}
Zaretsky J, Kim JM, Baskharoun S, Zhao Y, Austrian J, Aphinyanaphongs Y, et~al.
\newblock Generative artificial intelligence to transform inpatient discharge summaries to patient-friendly language and format.
\newblock JAMA Netw Open. 2024 Mar;7(3):e240357.

\bibitem[\protect\citeauthoryear{Hong et~al.}{2024}]{Hong2024-cd}
Hong J, Lee N, Thorne J.
\newblock Reference-free monolithic preference optimization with odds ratio.
\newblock arXiv [csCL]. 2024 Mar;{[cs.CL]}.

\bibitem[\protect\citeauthoryear{Bedi et~al.}{2024}]{Bedi2024-rb}
Bedi S, Liu Y, Orr-Ewing L, Dash D, Koyejo S, Callahan A, et~al.
\newblock Testing and evaluation of health care applications of large language models: A systematic review: A systematic review.
\newblock JAMA. 2024 Oct;.

\bibitem[\protect\citeauthoryear{Bannur et~al.}{2024}]{Bannur2024-qn}
Bannur S, Bouzid K, Castro DC, Schwaighofer A, Thieme A, Bond-Taylor S, et~al.
\newblock {MAIRA}-2: Grounded Radiology Report Generation.
\newblock arXiv [csCL]. 2024 Jun;{[cs.CL]}.

\bibitem[\protect\citeauthoryear{Shah et~al.}{2023}]{Shah2023-ai}
Shah NH, Entwistle D, Pfeffer MA.
\newblock Creation and adoption of large language models in medicine.
\newblock JAMA. 2023 Sep;330(9):866--869.

\bibitem[\protect\citeauthoryear{Xie et~al.}{2024}]{Xie2024-lo}
Xie Y, Zhang S, Cheng H, Liu P, Gero Z, Wong C, et~al.
\newblock {DocLens}: Multi-aspect fine-grained medical text evaluation.
\newblock In: Proceedings of the 62nd Annual Meeting of the Association for Computational Linguistics (Volume 1: Long Papers). Stroudsburg, PA, USA: Association for Computational Linguistics; 2024. p. 649--679.

\bibitem[\protect\citeauthoryear{Munnangi et~al.}{2024}]{Munnangi2024-wl}
Munnangi M, Swaminathan A, Alan FJ, Jindal J, Narayanan S, Lopez I, et~al.
\newblock Assessing the limitations of large language models in clinical fact decomposition.
\newblock arXiv [csCL]. 2024 Dec;{[cs.CL]}.

\bibitem[\protect\citeauthoryear{Adler-Milstein et~al.}{2024}]{Adler-Milstein2024-qf}
Adler-Milstein J, Redelmeier DA, Wachter RM.
\newblock The limits of clinician vigilance as an {AI} safety bulwark.
\newblock JAMA. 2024 Apr;331(14):1173--1174.

\bibitem[\protect\citeauthoryear{Augenstein et~al.}{2024}]{Augenstein2024-tv}
Augenstein I, Baldwin T, Cha M, Chakraborty T, Ciampaglia GL, Corney D, et~al.
\newblock Factuality challenges in the era of large language models and opportunities for fact-checking.
\newblock Nat Mach Intell. 2024 Aug;6(8):852--863.

\bibitem[\protect\citeauthoryear{Lewis et~al.}{2020}]{Lewis2020-fk}
Lewis P, Perez E, Piktus A, Petroni F, Karpukhin V, Goyal N, et~al.
\newblock Retrieval-augmented generation for knowledge-intensive {NLP} tasks.
\newblock Neural Inf Process Syst. 2020 May;abs/2005.11401:9459--9474.

\bibitem[\protect\citeauthoryear{Zheng et~al.}{2023}]{Zheng2023-kj}
Zheng L, Chiang WL, Sheng Y, Zhuang S, Wu Z, Zhuang Y, et~al.
\newblock Judging {LLM}-as-a-judge with {MT}-bench and Chatbot Arena.
\newblock Neural Inf Process Syst. 2023 Jun;abs/2306.05685.

\bibitem[\protect\citeauthoryear{Guan et~al.}{2024}]{Guan2024-oq}
Guan J, Dodge J, Wadden D, Huang M, Peng H.
\newblock Language models hallucinate, but may excel at fact verification.
\newblock In: Proceedings of the 2024 Conference of the North American Chapter of the Association for Computational Linguistics: Human Language Technologies (Volume 1: Long Papers). Stroudsburg, PA, USA: Association for Computational Linguistics; 2024. p. 1090--1111.

\bibitem[\protect\citeauthoryear{Wei et~al.}{2024}]{Wei2024-ur}
Wei J, Yang C, Song X, Lu Y, Hu N, Tran D, et~al.
\newblock Long-form factuality in large language models.
\newblock arXiv [csCL]. 2024 Mar;{[cs.CL]}.

\bibitem[\protect\citeauthoryear{Min et~al.}{2023}]{Min2023-zk}
Min S, Krishna K, Lyu X, Lewis M, Yih WT, Koh P, et~al.
\newblock {FActScore}: Fine-grained atomic evaluation of factual precision in long form text generation.
\newblock In: Proceedings of the 2023 Conference on Empirical Methods in Natural Language Processing. Stroudsburg, PA, USA: Association for Computational Linguistics; 2023. p. 12076--12100.

\bibitem[\protect\citeauthoryear{Kamoi et~al.}{2023}]{Kamoi2023-tm}
Kamoi R, Goyal T, Rodriguez J, Durrett G.
\newblock {WiCE}: Real-world entailment for claims in Wikipedia.
\newblock In: Proceedings of the 2023 Conference on Empirical Methods in Natural Language Processing. Stroudsburg, PA, USA: Association for Computational Linguistics; 2023. p. 7561--7583.

\bibitem[\protect\citeauthoryear{Chen et~al.}{2024}]{Chen2024-rd}
Chen J, Kim G, Sriram A, Durrett G, Choi E.
\newblock Complex claim verification with evidence retrieved in the wild.
\newblock In: Proceedings of the 2024 Conference of the North American Chapter of the Association for Computational Linguistics: Human Language Technologies (Volume 1: Long Papers). Stroudsburg, PA, USA: Association for Computational Linguistics; 2024. p. 3569--3587.

\bibitem[\protect\citeauthoryear{Es et~al.}{2024}]{Es2024-qc}
Es S, James J, Anke LE, Schockaert S.
\newblock {RAGAs}: Automated Evaluation of Retrieval Augmented Generation.
\newblock In: Proceedings of the 18th Conference of the European Chapter of the Association for Computational Linguistics: System Demonstrations; 2024. p. 150--158.

\bibitem[\protect\citeauthoryear{Johnson et~al.}{2016}]{Johnson2016-bt}
Johnson AEW, Pollard TJ, Shen L, Lehman LWH, Feng M, Ghassemi M, et~al.
\newblock {MIMIC}-{III}, a freely accessible critical care database.
\newblock Sci Data. 2016 May;3:160035.

\bibitem[\protect\citeauthoryear{Johnson et~al.}{2016}]{Johnson2016-ra}
Johnson A, Pollard T, Mark R.: {MIMIC}-{III} Clinical Database (version 1.4).
\newblock PhysioNet.

\bibitem[\protect\citeauthoryear{Russell}{2009}]{Russell2009-tm}
Russell B.
\newblock The philosophy of logical atomism.
\newblock 1st ed. Routledge Classics. London, England: Routledge; 2009.

\bibitem[\protect\citeauthoryear{Wanner et~al.}{2024}]{Wanner2024-ts}
Wanner M, Ebner S, Jiang Z, Dredze M, Van~Durme B.
\newblock A closer look at claim decomposition.
\newblock arXiv [csCL]. 2024 Mar;{[cs.CL]}.

\bibitem[\protect\citeauthoryear{Smith}{2012}]{Smith2012-us}
Smith NJJ.
\newblock Logic: The laws of truth.
\newblock Princeton, NJ: Princeton University Press; 2012.

\bibitem[\protect\citeauthoryear{Wang et~al.}{2024}]{Wang2024-gb}
Wang Y, Gangi~Reddy R, Mujahid ZM, Arora A, Rubashevskii A, Geng J, et~al.
\newblock Factcheck-bench: Fine-grained evaluation benchmark for automatic fact-checkers.
\newblock In: Findings of the Association for Computational Linguistics: EMNLP 2024. Stroudsburg, PA, USA: Association for Computational Linguistics; 2024. p. 14199--14230.

\bibitem[\protect\citeauthoryear{Chen et~al.}{2024}]{Chen2024-lm}
Chen T, Wang H, Chen S, Yu W, Ma K, Zhao X, et~al.
\newblock Dense {X} retrieval: What retrieval granularity should we use?
\newblock In: Proceedings of the 2024 Conference on Empirical Methods in Natural Language Processing. Stroudsburg, PA, USA: Association for Computational Linguistics; 2024. p. 15159--15177.

\bibitem[\protect\citeauthoryear{Miller}{1998}]{Miller1998-ea}
Miller E.
\newblock An introduction to the resource description framework.
\newblock Bull Am Soc Inf Sci. 1998 Oct;25(1):15--19.

\bibitem[\protect\citeauthoryear{Bhattacharyya}{2015}]{Bhattacharyya2015-zx}
Bhattacharyya SB.
\newblock Introduction to {SNOMED} {CT}.
\newblock 1st ed. Singapore, Singapore: Springer; 2015.

\bibitem[\protect\citeauthoryear{Lassance and Clinchant}{2022}]{Lassance2022-qs}
Lassance C, Clinchant S.
\newblock An Efficiency Study for {SPLADE} Models.
\newblock In: Proceedings of the 45th International ACM SIGIR Conference on Research and Development in Information Retrieval. SIGIR '22. New York, NY, USA: Association for Computing Machinery; 2022. p. 2220--2226.

\bibitem[\protect\citeauthoryear{Chen et~al.}{2024}]{Chen2024-mi}
Chen J, Wang N, Li C, Wang B, Xiao S, Xiao H, et~al.
\newblock {AIR}-Bench: Automated Heterogeneous Information Retrieval Benchmark.
\newblock arXiv [csIR]. 2024 Dec;{[cs.IR]}.

\bibitem[\protect\citeauthoryear{Nogueira and Cho}{2019}]{Nogueira2019-ql}
Nogueira R, Cho K.
\newblock Passage re-ranking with {BERT}.
\newblock arXiv [csIR]. 2019 Jan;{[cs.IR]}.

\bibitem[\protect\citeauthoryear{Déjean et~al.}{2024}]{Dejean2024-hd}
Déjean H, Clinchant S, Formal T.
\newblock A thorough comparison of cross-encoders and {LLMs} for reranking {SPLADE}.
\newblock arXiv [csIR]. 2024 Mar;{[cs.IR]}.

\bibitem[\protect\citeauthoryear{Chen et~al.}{2024}]{Chen2024-vn}
Chen J, Xiao S, Zhang P, Luo K, Lian D, Liu Z.
\newblock {M3}-Embedding: Multi-Linguality, Multi-Functionality, Multi-Granularity Text Embeddings Through Self-Knowledge Distillation.
\newblock In: Findings of the Association for Computational Linguistics ACL 2024; 2024. p. 2318--2335.

\bibitem[\protect\citeauthoryear{Thakur et~al.}{2021}]{Thakur2021-qd}
Thakur N, Reimers N, Rücklé A, Srivastava A, Gurevych I.
\newblock {BEIR}: A Heterogeneous Benchmark for Zero-shot Evaluation of Information Retrieval Models.
\newblock Proceedings of the Neural Information Processing Systems Track on Datasets and Benchmarks. 2021 Dec;1.

\bibitem[\protect\citeauthoryear{Formal et~al.}{2021}]{Formal2021-qk}
Formal T, Piwowarski B, Clinchant S.
\newblock {SPLADE}: Sparse lexical and expansion model for first stage ranking.
\newblock In: Proceedings of the 44th International ACM SIGIR Conference on Research and Development in Information Retrieval. New York, NY, USA: ACM; 2021. .

\bibitem[\protect\citeauthoryear{Kong et~al.}{2023}]{Kong2023-vs}
Kong W, Dudek JM, Li C, Zhang M, Bendersky M.
\newblock {SparseEmbed}: Learning Sparse Lexical Representations with Contextual Embeddings for Retrieval.
\newblock In: Proceedings of the 46th International ACM SIGIR Conference on Research and Development in Information Retrieval. SIGIR '23. New York, NY, USA: Association for Computing Machinery; 2023. p. 2399--2403.

\bibitem[\protect\citeauthoryear{Wang et~al.}{2024}]{Wang2024-ao}
Wang B, Yue X, Su Y, Sun H.
\newblock Grokked transformers are implicit reasoners: A mechanistic journey to the edge of generalization.
\newblock arXiv [csCL]. 2024 May;{[cs.CL]}.

\bibitem[\protect\citeauthoryear{Song et~al.}{2024}]{Song2024-la}
Song J, Xu Z, Zhong Y.
\newblock Out-of-distribution generalization via composition: a lens through induction heads in Transformers.
\newblock arXiv [csCL]. 2024 Aug;{[cs.CL]}.

\bibitem[\protect\citeauthoryear{Parmar et~al.}{2024}]{Parmar2024-ep}
Parmar M, Patel N, Varshney N, Nakamura M, Luo M, Mashetty S, et~al.
\newblock {LogicBench}: Towards Systematic Evaluation of Logical Reasoning Ability of Large Language Models.
\newblock In: Proceedings of the 62nd Annual Meeting of the Association for Computational Linguistics (Volume 1: Long Papers); 2024. p. 13679--13707.

\bibitem[\protect\citeauthoryear{Pan et~al.}{2023}]{Pan2023-kj}
Pan L, Albalak A, Wang X, Wang W.
\newblock Logic-{LM}: Empowering large language models with symbolic solvers for faithful logical reasoning.
\newblock In: Findings of the Association for Computational Linguistics: EMNLP 2023. Stroudsburg, PA, USA: Association for Computational Linguistics; 2023. p. 3806--3824.

\bibitem[\protect\citeauthoryear{Wan et~al.}{2024}]{Wan2024-op}
Wan Y, Wang W, Yang Y, Yuan Y, Huang JT, He P, et~al.
\newblock {LogicAsker}: Evaluating and improving the logical reasoning ability of large language models.
\newblock In: Proceedings of the 2024 Conference on Empirical Methods in Natural Language Processing. Stroudsburg, PA, USA: Association for Computational Linguistics; 2024. p. 2124--2155.

\bibitem[\protect\citeauthoryear{Patel et~al.}{2024}]{Patel2024-bc}
Patel N, Kulkarni M, Parmar M, Budhiraja A, Nakamura M, Varshney N, et~al.
\newblock Multi-{LogiEval}: Towards evaluating multi-step logical reasoning ability of large language models.
\newblock In: Proceedings of the 2024 Conference on Empirical Methods in Natural Language Processing. Stroudsburg, PA, USA: Association for Computational Linguistics; 2024. p. 20856--20879.

\bibitem[\protect\citeauthoryear{Gunjal and Durrett}{2024}]{Gunjal2024-mi}
Gunjal A, Durrett G.
\newblock Molecular Facts: Desiderata for Decontextualization in {LLM} Fact Verification.
\newblock In: Findings of the Association for Computational Linguistics: EMNLP 2024; 2024. p. 3751--3768.

\bibitem[\protect\citeauthoryear{Choi et~al.}{2021}]{Choi2021-zu}
Choi E, Palomaki J, Lamm M, Kwiatkowski T, Das D, Collins M.
\newblock Decontextualization: Making sentences stand-alone.
\newblock Trans Assoc Comput Linguist. 2021 Apr;9:447--461.

\bibitem[\protect\citeauthoryear{Wang et~al.}{2024}]{Wang2024-hv}
Wang T, Kulikov I, Golovneva O, Yu P, Yuan W, Dwivedi-Yu J, et~al.
\newblock Self-Taught Evaluators.
\newblock arXiv [csCL]. 2024 Aug;{[cs.CL]}.

\bibitem[\protect\citeauthoryear{Bavaresco et~al.}{2024}]{Bavaresco2024-gt}
Bavaresco A, Bernardi R, Bertolazzi L, Elliott D, Fernández R, Gatt A, et~al.
\newblock {LLMs} instead of Human Judges? A Large Scale Empirical Study across 20 {NLP} Evaluation Tasks.
\newblock arXiv [csCL]. 2024 Jun;{[cs.CL]}.

\bibitem[\protect\citeauthoryear{Bell et~al.}{2020}]{Bell2020-md}
Bell SK, Delbanco T, Elmore JG, Fitzgerald PS, Fossa A, Harcourt K, et~al.
\newblock Frequency and types of patient-reported errors in electronic health record ambulatory care notes.
\newblock JAMA Netw Open. 2020 Jun;3(6):e205867.

\bibitem[\protect\citeauthoryear{Collins et~al.}{2024}]{Collins2024-ey}
Collins GS, Moons KGM, Dhiman P, Riley RD, Beam AL, Van~Calster B, et~al.
\newblock {TRIPOD+AI} statement: updated guidance for reporting clinical prediction models that use regression or machine learning methods.
\newblock BMJ. 2024 Apr;385:e078378.

\bibitem[\protect\citeauthoryear{Kwon et~al.}{2023}]{Kwon2023-gr}
Kwon W, Li Z, Zhuang S, Sheng Y, Zheng L, Yu CH, et~al.
\newblock Efficient Memory Management for Large Language Model Serving with {PagedAttention}.
\newblock In: Proceedings of the 29th Symposium on Operating Systems Principles. New York, NY, USA: ACM; 2023. p. 611--626.

\bibitem[\protect\citeauthoryear{Dubey et~al.}{2024}]{Dubey2024-gu}
Dubey A, Jauhri A, Pandey A, Kadian A, Al-Dahle A, Letman A, et~al.
\newblock The Llama 3 herd of models.
\newblock arXiv [csAI]. 2024 Jul;{[cs.AI]}.

\bibitem[\protect\citeauthoryear{Zhang et~al.}{2024}]{Zhang2024-tz}
Zhang E, Zhu V, Saphra N, Kleiman A, Edelman BL, Tambe M, et~al.
\newblock Transcendence: Generative Models Can Outperform The Experts That Train Them.
\newblock arXiv [csLG]. 2024 Jun;{[cs.LG]}.

\bibitem[\protect\citeauthoryear{Willard and Louf}{2023}]{Willard2023-wp}
Willard BT, Louf R.
\newblock Efficient Guided Generation for Large Language Models.
\newblock arXiv [csCL]. 2023 Jul;{[cs.CL]}.

\bibitem[\protect\citeauthoryear{Brown et~al.}{2020}]{Brown2020-sb}
Brown TB, Mann B, Ryder N, Subbiah M, Kaplan J, Dhariwal P, et~al.
\newblock Language models are few-shot learners.
\newblock In: Proceedings of the 34th International Conference on Neural Information Processing Systems. No. Article 159 in NIPS'20. Red Hook, NY, USA: Curran Associates Inc.; 2020. p. 1877--1901.

\bibitem[\protect\citeauthoryear{Olsson et~al.}{2022}]{Olsson2022-oc}
Olsson C, Elhage N, Nanda N, Joseph N, DasSarma N, Henighan T, et~al.
\newblock In-context Learning and Induction Heads.
\newblock arXiv [csLG]. 2022 Sep;{[cs.LG]}.

\bibitem[\protect\citeauthoryear{Feil}{}]{FeilUnknown-ni}
Feil M.: infinity: Infinity is a high-throughput, low-latency {REST} {API} for serving text-embeddings, reranking models and clip.

\bibitem[\protect\citeauthoryear{Li et~al.}{2023}]{Li2023-ow}
Li C, Liu Z, Xiao S, Shao Y.
\newblock Making large language models A better foundation for dense retrieval.
\newblock arXiv [csCL]. 2023 Dec;{[cs.CL]}.

\bibitem[\protect\citeauthoryear{Bird et~al.}{2009}]{Bird2009-nr}
Bird S, Klein E, Loper E.
\newblock Natural language processing with python.
\newblock Sebastopol, CA: O'Reilly Media; 2009.

\bibitem[\protect\citeauthoryear{}{}]{UnknownUnknown-nb}
: Label-Studio: Label Studio is a multi-type data labeling and annotation tool with standardized output format.

\bibitem[\protect\citeauthoryear{Malkov and Yashunin}{2016}]{Malkov2016-tb}
Malkov YA, Yashunin DA.
\newblock Efficient and robust approximate nearest neighbor search using Hierarchical Navigable Small World graphs.
\newblock arXiv [csDS]. 2016 Mar;{[cs.DS]}.

\bibitem[\protect\citeauthoryear{Mazzeschi}{2023}]{Mazzeschi2023-ov}
Mazzeschi M.: Distribution-Based Score Fusion ({DBSF}), a new approach to Vector Search Ranking.
\newblock Accessed: 2024-9-18.
\newblock \url{https://medium.com/plain-simple-software/distribution-based-score-fusion-dbsf-a-new-approach-to-vector-search-ranking-f87c37488b18}.

\bibitem[\protect\citeauthoryear{}{}]{UnknownUnknown-ku}
: {DeepEval}: The {LLM} Evaluation Framework.

\bibitem[\protect\citeauthoryear{Gwet}{2008}]{Gwet2008-ga}
Gwet KL.
\newblock Computing inter-rater reliability and its variance in the presence of high agreement.
\newblock Br J Math Stat Psychol. 2008 May;61(Pt 1):29--48.

\end{thebibliography}
%% if required, the content of .bbl file can be included here once bbl is generated
%%\input sn-article.bbl

\end{document}